\documentclass[sigconf, nonacm]{acmart}
\renewcommand\footnotetextcopyrightpermission[1]{} 
\renewcommand{\footnotetextcopyrightpermission}[1]{}
\AtBeginDocument{%
  }

\copyrightyear{2025}
\acmYear{2025}
\setcopyright{cc}
\setcctype{by}

\acmConference[CUHK]{ }{2025}{MESH}

\usepackage{microtype}
\usepackage{graphicx}
\usepackage{subcaption}
\usepackage{booktabs} 
\usepackage{xcolor}
\usepackage{multirow}
\usepackage{longtable}
\usepackage{algorithm} 
\usepackage{algpseudocode}
\usepackage{hyperref}
\newcommand{\rewr}[1]{\textcolor{black}{#1}}
\newcommand{\add}[1]{\textcolor{black}{#1}}
\newcommand{\acm}[1]{\textcolor{black}{#1}}
\newcommand{\acmo}[1]{\textcolor{black}{#1}}
\newcommand{\uselink}[1]{\textcolor{blue}{#1}}

\newcommand{\benchf}{\textbf{M}ise-\textbf{E}n-\textbf{S}cène-\textbf{H}allucinator}
\newcommand{\miset}{\textit{Mise-En-Scène}}

\usepackage{amsmath}
\usepackage{mathtools}
\usepackage{amsthm}
\usepackage[capitalize,noabbrev]{cleveref}

\theoremstyle{plain}
\newtheorem{theorem}{Theorem}[section]

\theoremstyle{definition}
\newtheorem{definition}[theorem]{Definition}

\theoremstyle{remark}

\newenvironment{myquotation}{\setlength{\leftmargini}{0em}\quotation}{\endquotation}

\usepackage[textsize=tiny]{todonotes}

\begin{document}

\title{\textit{MESH} - Understanding Videos Like Human: Measuring Hallucinations in Large Video Models}

\author{Garry Yang}
\authornote{Both authors contributed equally to this research.}
\affiliation{%
  \institution{The Chinese University of Hong Kong}
  \city{Hong Kong}
  \country{Hong Kong SAR}}
\email{hcyang@cse.cuhk.edu.hk}
\orcid{0009-0000-9534-6765}

\author{Zizhe Chen}
\authornotemark[1]
\affiliation{%
  \institution{The Chinese University of Hong Kong}
  \city{Hong Kong}
  \country{Hong Kong SAR}}
\email{zzchen2@cse.cuhk.edu.hk}
\orcid{0009-0005-7032-1525}

\author{Man Hon Wong}
\affiliation{%
  \institution{The Chinese University of Hong Kong}
  \city{Hong Kong}
  \country{Hong Kong SAR}}
\email{mhwharry@gmail.com}
\orcid{0009-0000-8553-6004}

\author{Haoyu Lei}
\affiliation{%
  \institution{The Chinese University of Hong Kong}
  \city{Hong Kong}
  \country{Hong Kong SAR}}
  \email{hylei22@cse.cuhk.edu.hk}
\orcid{0009-0009-1393-2273}

\author{Yongqiang Chen}
\affiliation{%
 \institution{The Chinese University of Hong Kong}
 \city{Hong Kong}
  \country{Hong Kong SAR}}
  \email{yqchen@cse.cuhk.edu.hk}
\orcid{0000-0003-2485-3529}

\author{Zhenguo Li}
\affiliation{%
  \institution{Huawei Noah's Ark Lab}
 \city{Hong Kong}
  \country{Hong Kong SAR}}
  \email{zhenguol@gmail.com}
\orcid{0000-0002-8492-3069}

\author{Kaiwen Zhou}
\affiliation{%
  \institution{Huawei Noah's Ark Lab}
 \city{Shenzhen}
  \country{China}}
  \email{zhoukaiwen2@huawei.com}
\orcid{0000-0002-3088-8085}

\author{James Cheng}
\affiliation{%
  \institution{The Chinese University of Hong Kong}
 \city{Hong Kong}
  \country{Hong Kong SAR}}
  \email{jcheng@cse.cuhk.edu.hk}
\orcid{0000-0001-6313-6288}

\renewcommand{\shortauthors}{Garry Yang et al.}

\begin{abstract}
Large Video Models (LVMs) build on the semantic capabilities of Large Language Models (LLMs) and vision modules by integrating temporal information to better understand dynamic video content.
Despite their progress, LVMs are prone to hallucinations—producing inaccurate or irrelevant descriptions.
\acm{Current benchmarks for video hallucination depend heavily on manual categorization of video content, neglecting the perception-based processes through which humans naturally interpret videos.}
We introduce \textbf{MESH}, a benchmark designed to evaluate hallucinations in LVMs systematically. 
\rewr{MESH uses a Question-Answering framework with binary and multi-choice formats incorporating target and trap instances.}
It follows a bottom-up approach, evaluating basic objects, coarse-to-fine subject features, and subject-action pairs, aligning with human video understanding.
\add{We demonstrate that MESH offers an effective and comprehensive approach for identifying hallucinations in videos. Our evaluations show that while LVMs excel at recognizing basic objects and features, their susceptibility to hallucinations increases markedly when handling fine details or aligning multiple actions involving various subjects in longer videos.}
\acmo{The benchmark is available at \href{https://github.com/HCYANG2000/MESH-Benchmark}{\uselink{MESH-Benchmark}}.}
\end{abstract}




\maketitle

\section{Introduction}
Large language models (LLMs) \cite{llm-start} excel in handling real-world tasks by responding to user queries and generalizing across multi-modal information, including images, videos, and audio \cite{llm-multi-1}. Open-source models like LLaMA \cite{llama-1} can be pre-trained on modality-specific datasets to align these modalities with language tokens \cite{bind-1}, while closed-source models like GPT-4 \cite{gpt4-1} refine and summarize information from modality-specific encoders \cite{mm-vid}. 
\acm{
For vision-language tasks, pre-trained vision encoders such as CLIP \cite{clip} convert visual inputs into dense visual tokens that capture semantic and spatial layout. 
A language model encodes the text prompt into textual tokens. These visual and textual tokens are then aligned and fused to enable joint reasoning across modalities.
This integration enables a wide range of downstream tasks, including image captioning, visual question answering (VQA), and image-based storytelling \cite{lvlm-task-1,lvlm-task-2}. This architecture, which combines vision and language understanding, constitutes a large vision-language model (LVLM).
}

\acm{
Video understanding involves integrating visual, auditory, and temporal cues to interpret dynamic content~\cite{lvlm-survey-1, re-v1-2}. 
Humans naturally track actions, dialogue, and subtle temporal changes across frames. Basic grounding information, such as object locations and identities, serves as the foundation for higher-level comprehension, including emotion recognition \cite{human-under-4, human-under-5}.
To perform similar tasks, such as video captioning, large vision-language models (LVLMs) first apply sampling techniques to convert videos into sequences of frames, then encode and summarize each frame’s visual content to extract semantic meaning and capture temporal dynamics \cite{lvlm-survey-2}. 
Through fine-tuning on video datasets, LVLMs can function as large video models (LVMs), enabling more accurate extraction, alignment, and summarization of multimodal information from video content~\cite{lvlm-finetune-1}.
}
\begin{figure}[t]
    \centering
    \includegraphics[width=\columnwidth]{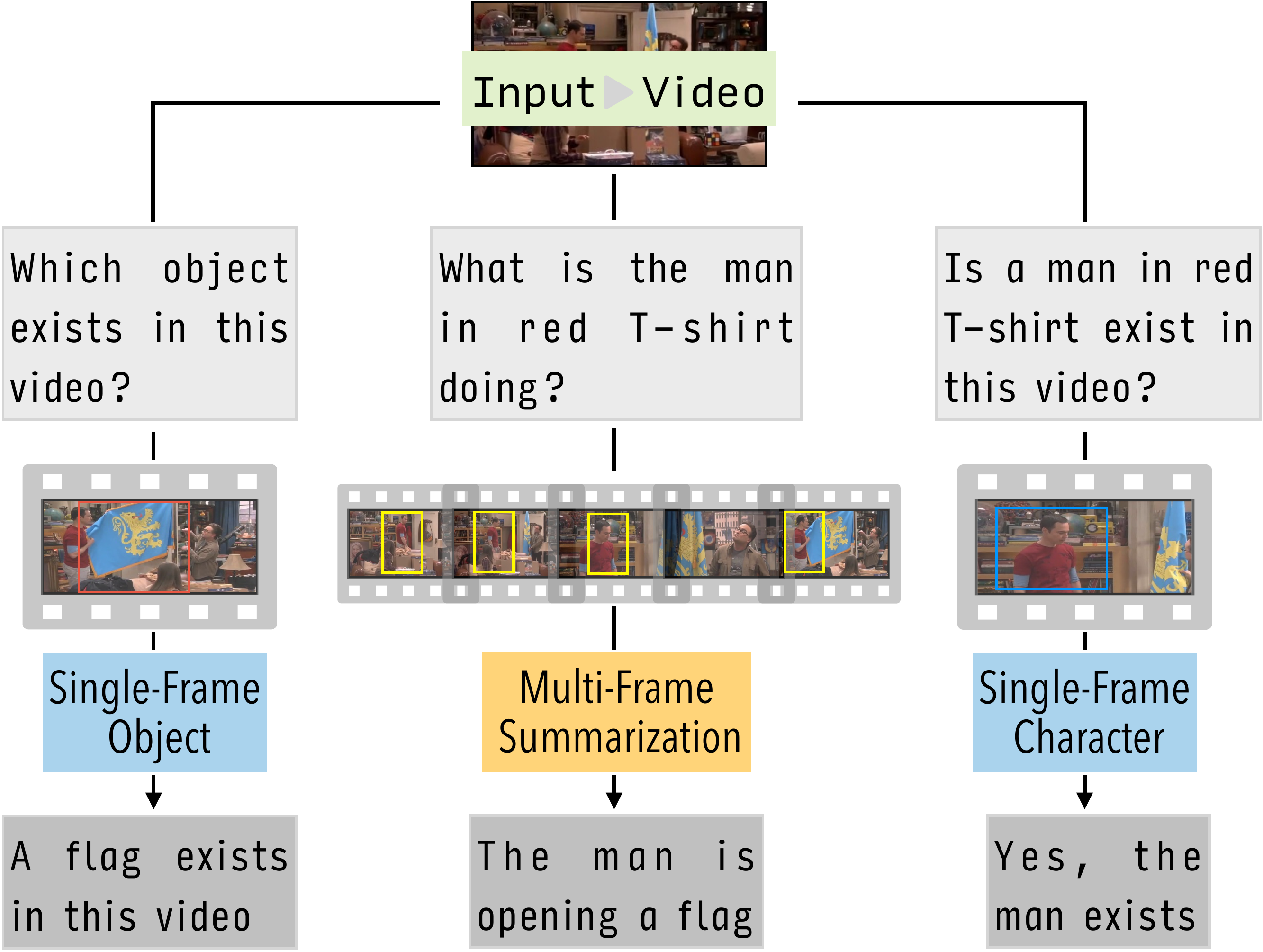} 
    \vspace{-0.6cm}
    \caption{LVMs understand videos.}
    \label{fig:intro-under}
    \Description{intro}
    \vspace{-0.6cm}
\end{figure}


A key challenge when applying LVMs to video-specific tasks is the exacerbation of hallucination issues \cite{hallu-2}. 
LLMs, which are prone to generating irrelevant or incorrect content in response to prompts \cite{hallu-1}, face a heightened risk of hallucination due to complex interactions between multiple components \cite{pope}. Vision encoders, which are pre-trained to map both visual and textual data into a shared latent space \cite{clip, bind-1}, often prioritize correlations learned from training data over accurate visual interpretation \cite{vision-corr-1}. 
Processing videos generates a large volume of visual tokens from sequential frames, requiring aggressive sampling strategies that risk discarding important information~\cite{token-many-1}.
As illustrated in Figure~\ref{fig:intro-under}, distinguishing objects or characters and their features requires localizing the relevant frame, while actions spanning multiple frames—such as opening a flag—demand precise summarization of temporally distributed visual tokens~\cite{tem-adapt, vript}.
To investigate into the hallucination of video understanding, we pose the following challenge:
\begin{myquotation}
\emph{``Is there a comprehensive benchmark for measuring hallucination in video understanding?”}
\end{myquotation}


\acm{Most existing benchmarks either adopt caption-based comparisons or adapt the binary question format from POPE \cite{pope}. They evaluate LVMs on various 
aspects, including subject recognition or temporal details \cite{v-hallucer, vript, VidHalluc, VidHal, EventHallusion, vqa-3, TVBench, VELOCITI}.} 
\acm{
However, most existing approaches depend on manual categorization of video content, leading to questions with varying levels of granularity—for instance, identifying scene transitions (coarse) versus counting specific actions (fine).
Currently, there is no perception-based framework for categorizing hallucination types that aligns with how humans intuitively understand video.
Humans typically process video in a bottom-up manner: first recognizing the environment and objects (\textit{Setting}), then identifying key individuals (\textit{Character}), and finally interpreting their actions and dialogues (\textit{Stage}) \cite{human-under-1, visual}.
To replicate this, we target measuring human understanding of the visual aspects of videos and emphasizes \textit{mise-en-scène}, which encompasses all visual elements in videos~\cite{human-under-1, human-under-2}. \textit{Mise-en-scène} is categorized into three domains: \textbf{setting}, \textbf{characters}, and \textbf{stage}. 
The \textbf{setting} includes the physical environment and objects, providing narrative context and relationships~\cite{human-under-1, human-under-3}. \textbf{Characters} involve physical traits for main subjects, including clothing, contributing to identity portrayal~\cite{human-under-2}. We focus on distinguishing human characters in this work.
\textbf{Stage} 
includes character movements, requiring models to accurately identify action/dialogue sequences and associate them with the relevant subjects~\cite{human-under-3}.
Detailed description is present in Appendix \ref{ap:human-under}.
}
\acm{
To demonstrate the practical use of this perception categorization, we use the TVQA+ dataset \cite{tvqa} for its richer context compared to daily footage. 
}
\acmo{Our method is easily transferable to other datasets, such as UCF101 \cite{action_recog_1}. The generalized annotation pipeline and experiment results are shown in Appendix \ref{ap:transfer-method}.}
We then create binary and multi-choice questions featuring target and trap instances based on TVQA+ annotations, forming \benchf~(\textbf{MESH})
This benchmark assesses LVMs’ capability to recognize basic elements, discern fine details, and detect subject-action pairs in videos with complex interactions.

\begin{table}[h!]
\vspace{-0.3cm}
\caption{Results with various difficulties of MESH}
\label{tb:intro-all}
\centering
\vspace{-0.4cm}
\begin{tabular}{lccc}
\toprule
\textbf{Model} & \textbf{Advanced} & \textbf{Basic} & \textbf{Average}  \\\midrule
InternVL2.5-78B  & \textbf{73.4} & \textbf{90.1}  & \textbf{85.6} \\ 
LLaVA-Video-72B  & 70.1 & 89.5 & 84.8   \\ 
LLaVA-OV-72B     & 62.4 & 81.9  & 77.3 \\ 
Aria-23B         & 57.0 & 81.9  & 76.6 \\ 
LLaVA-NV-32B     & 54.0 & 76.7  & 69.6 \\ 
\hline
GPT-4o           & 58.7 & 83.8  & 79.1\\
\bottomrule
\end{tabular}
\vspace{-0.3cm}
\end{table}
Using MESH, we conduct extensive experiments across LVMs with various configurations. Our findings are: \textbf{1)} LVMs \textbf{struggle with fine character details and aligning multiple actions across subjects in longer videos}, despite accurately capturing basic objects and coarse features, as demonstrated by basic and advanced tasks in Table \ref{tb:intro-all}. \textbf{2)} LVMs with weaker performance in Setting/Character aspects \textbf{exhibit increased hallucinations when processing spanning frames}, while stronger LVMs maintain accuracy and \textbf{leverage multi-frame tokens for finer detail prediction}. \textbf{3)} MESH results \textbf{align with other video understanding benchmarks}, suggesting that mitigating MESH hallucinations improves performance on general tasks.
\section{Related Works}

\subsection{Video Understanding Models}\label{sec:relate-lvlm}

\acmo{Video understanding tasks 
\cite{lvlm-survey-1, lvlm-survey-2, lvlm-task-1, lvlm-task-2} traditionally relied on feature extraction techniques 
to encode relevant video characteristics \cite{trad-1, trad-2, trad-3, trad-4}. }
\acmo{The emergence of deep learning techniques
has significantly advanced spatial and temporal feature capture \cite{lvlm-survey-1}, outperforming handcrafted methods.}
To address the lack of task-specific video labels, self-supervised learning \cite{clip, self-1, contrast-1} enabled the development of general pre-trained video models, fine-tuned for diverse downstream tasks \cite{lvlm-survey-2}. 
Despite advancements, video understanding models still struggle to comprehend the underlying semantic content of each frame and often fail to relate relevant contexts when answering questions involving both visual and chronological data \cite{lvlm-task-1, lvlm-survey-3}. With the improvement of large language models (LLMs), such as LLaMA \cite{llama-1} and GPT-4 \cite{gpt4-1}, these models have demonstrated a superior ability to understand user queries, including images and videos, and provide detailed answers or explanations without needing further fine-tuning. 
Techniques like these could facilitate video understanding models to communicate through user prompts, capturing, processing, and summarizing video content to address various tasks \cite{lvlm-task-2}. 

\subsection{Measuring Hallucination in Vision Models}\label{sec:relate-hallu}

In generative models, hallucination refers to outputs deviating from the input source (intrinsic) or established external knowledge (extrinsic)~\cite{hallu-3, hallu-4}. 
For LVLMs, hallucinations encompass object, multi-modal conflict, counter-knowledge, and attribute types~\cite{v-hallucer, hallu-7, hallu-6}. 
The POPE framework~\cite{pope} uses a polling-based question schema to assess object hallucinations with scalability and flexibility, while \citet{HallusionBench} extend this approach to binary questions for vision and language hallucinations.
Video benchmarks challenge LVMs to address hallucinations, such as identifying objects, people, and actions accurately~\cite{tem-adapt, vqa-1, vqa-2, vqa-3, vqa-4, vqa-5}. 
\acm{Many adopt POPE’s question structure to query spatial/temporal contents in videos~\cite{v-hallucer, vript, VidHalluc, VidHal, EventHallusion, TVBench, vqa-3, VideoCon}. }
\acm{
For example, \citet{vqa-3} categorize static/dynamic video content into sub-tasks of varying granularity, from objects/actions to scenes/states. \citet{TVBench} focus on multi-choice questions with temporally confusing options, while \citet{VideoCon} use contrastive methods to craft convincing negative captions. MESH also generates complex negative options across categories.
These benchmarks vary in difficulty due to a lack of bottom-up modeling of human video understanding and often miss key grounding details. 
A common issue is the neglect of subject-action alignment—humans comprehend videos by recognizing each subject’s specific action.
\citet{VELOCITI} constructs negative captioning using entity-action substitution. 
Compared with \citet{VELOCITI}, MESH uses longer videos, richer subject-action pairs, and precise speaker localization.
}
\section{\textit{MESH} Benchmark}
In this section, we provide a comprehensive explanation of our \textit{\benchf} benchmark. This includes both binary and multi-choice (MC) questions that examine various perspectives of a video.
Section \ref{sec:bench-data} introduces the annotations of the base dataset. Section \ref{sec:bench-meth} presents the polling pipeline to formulate hallucination questions alongside the evaluation metric. Section \ref{sec:bench-de} presents questions concerning hallucinations involving three parts: \textit{Setting with objects}, \textit{Characters with features}, and \textit{Stage with actions/dialogues}.

\subsection{Dataset Grounding}\label{sec:bench-data}
\begin{figure*}[t]
    \centering
    \includegraphics[width=17.1cm]{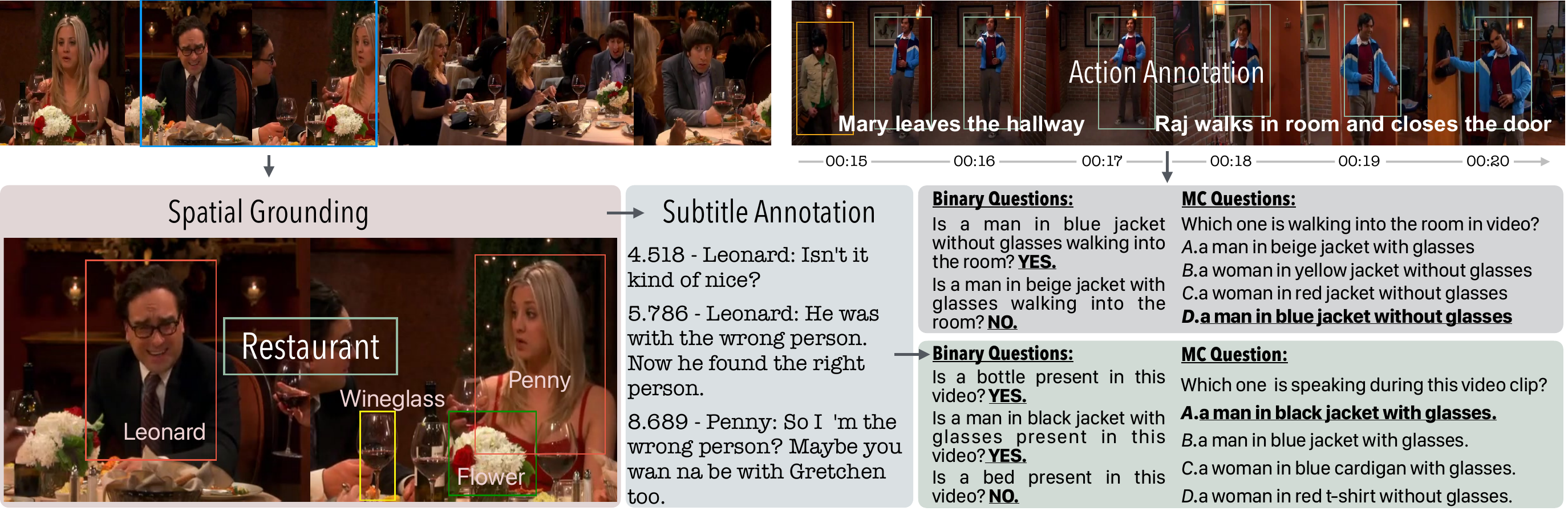} 
    \vspace{-0.4cm}
    \caption{Space Grounding and Subtitle/Action Annotations from TVQA+ and example questions}
    \label{fig:tvqa-struct}
    \Description{fig:tvqa-struct}
    \vspace{-0.4cm}
\end{figure*}
We begin by selecting the TVQA+ dataset \cite{tvqa} which provides spatial and temporal annotations along with subtitles as the basis dataset. 
Figure \ref{fig:tvqa-struct} illustrates the structure of TVQA+, which includes three key components. \textbf{Spatial grounding} provides frame-level bounding boxes for objects and characters mentioned in the original questions, such as identifying \textit{flowers}, \textit{bottle}, or characters like \textit{Penny} and \textit{Leonard}. We replace character names by their features, \add{detailed in Appendix \ref{ap:mesh-char}}. \textbf{Subtitle} annotates the start and end frames for character dialogues, marking speech timings precisely—for instance, the conversation between \textit{Penny} and \textit{Leonard}. We link speakers' name with their features, \add{detailed in Appendix \ref{ap:mesh-dia}}. \textbf{Action timestamp annotation} marks the start and end frames of actions from the original questions, such as identifying actions like “Raj walks into the room”. We align subjects with their features to construct subject-action pairs, \add{detailed in Appendix \ref{ap:mesh-act}}.

\subsection{Polling-based Methodology}\label{sec:bench-meth}
Empirical results in image hallucination measurements reveal that caption-based benchmarks are unstable, biased towards short captions, and reliant on manual, dataset-specific parsing rules, leading to frequent omissions and misclassifications \cite{pope}.
\rewr{To refine hallucination measurement beyond captioning, we extend POPE’s polling-based binary classification method \cite{pope} to include MC questions using multiple-probing techniques.} This format diversifies negative options, increasing evaluation difficulty and improving LVM performance differentiation \cite{mc-adv-1}. 
\add{Its effectiveness and robustness are demonstrated in Appendix \ref{ap:metric-mc}.}
We illustrate binary and MC question format in Figure \ref{fig:tvqa-struct}. 
\textbf{Binary evaluation} involves selecting a ground truth instance (\textit{target}) for positive questions (e.g., \textit{Is a bottle present in this video?}) and a non-existent or fake instance (\textit{trap}) for negative questions (e.g., \textit{Is a bed present in this video?}). Performance is evaluated using Accuracy, with the Positive Ratio metric accounting for the tendency of LVMs to overproduce ``yes" responses.
\textbf{MC evaluation} constructs MC questions by pairing one \textit{target} with three \textit{trap} instances. 
Overall accuracy measures the proportion of correct responses, while \textit{Option Balance} (OB) and \textit{Correct Option Balance} (COB) assess the distribution of option selections and correct answers, respectively, using standard deviation formulas, details given in \ref{ep:metric}. 
The target and trap instances span objects, character features, and actions, depending on the hallucination focus, as discussed further in the next section.

\subsection{Benchmark Detail}\label{sec:bench-de}
We present the definition of the question set shared across three dimensions. The target and trap instances in various hallucination aspects are detailed in the following section.
\begin{definition}
We constructs a 3-tuple list for each video clip, consisting of a clip, a \textit{yes-no} question set, and a multi-choice question set. A hallucinator tuple is:
\begin{equation}\label{eqn:whole}
   \langle v, \{q^b(m_i), a^b_i\}_{i=1}^{l_b}, \{q^c(n^{+}_j, n^{-}_{j_1}, \dots, n^{-}_{j_k}), a^c_j\}_{j=1}^{l_c}\rangle
\end{equation}
where $\{q^b(m_i), a^b_i\}_{i=1}^{l_b}$ represents the binary question set, with $m_i$ denoting a target (positive $m_i^+$ , $a^b_i = \text{yes}$) or trap instance (negative $m_i^-$ , $a^b_i = \text{no}$), generated by the template ``Is \_\_ present in this video?''. $\{q^c(n^{+}_j, n^{-}_{j_1}, \dots, n^{-}_{j_k}), a^c_j\}_{j=1}^{l_c}$ represents the multi-choice question set, where $n^{+}_j$ is a positive instance, $n^{-}_{j_1}, \dots, n^{-}_{j_k}$ are $k$ negative instances, and $a^c_j$ is the answer. Questions are generated using the template ``Which of the following is $present/speaking/action$ in this video?'' with $k = 3$ for a four-choice structure.
\end{definition}

\subsubsection{Setting Hallucination}\label{sec:bench-set}
\begin{figure}[h]
    \centering
    \includegraphics[width=\columnwidth]{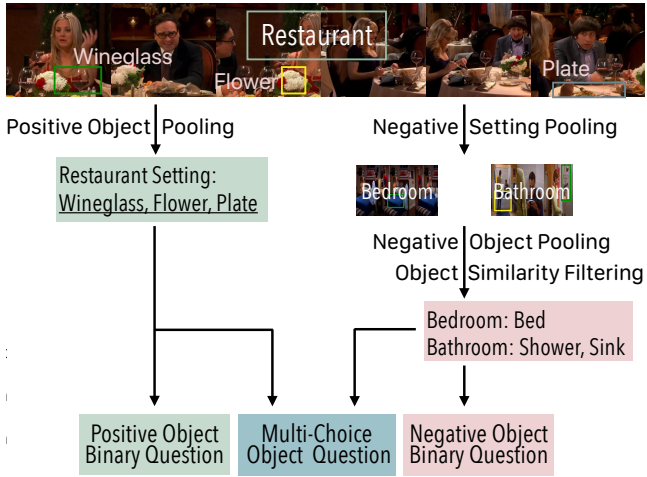} 
    \vspace{-0.6cm}
    \caption{Constructing setting hallucination questions}
    \label{fig:pipeline_setting}
    \vspace{-0.4cm}
    \Description{pipeline_setting}
\end{figure}
In this section, we describe how to construct hallucination questions for the \textbf{Setting} perspective.
The pipeline is outlined in Figure \ref{fig:pipeline_setting}. 
\add{Objects annotations in videos are more sparse, compared with images. We discuss this distinction in Appendix \ref{ap:mesh-hallu-set-uniq}.} 
We propose a \textbf{space-centered selection method}, where each video clip is assigned a \textbf{space label}. For example, in Figure \ref{fig:pipeline_setting}, the first video is labeled \textit{restaurant}, while the others are labeled \textit{bedroom} and \textit{bathroom}, respectively. 
The unique object set for each space label is defined as: 
\begin{equation}
        \mathcal{O}_s \leftarrow \text{Filter}\left(\cup_{\{v_y\vert s_y=s\}}\mathcal{O}_y\right)
\end{equation}
where \(\mathcal{O}_y\) represents the set of annotated objects in video \(v_y\) with label \(s_y = s\). \textit{Filter} removes common objects that appear across multiple space labels, ensuring that \(\mathcal{O}_s\) contains only distinct objects of label \(s\). \add{Detailed steps are provided in Appendix \ref{ap:mesh-hallu-set-anno}.}
For a video \( v_x \), the targets \( m_i^+ \) and \( n_i^+ \) in Equation \ref{eqn:whole} are directly selected from the annotated objects within its frames. Conversely, the traps for are selected from object sets associated with different space labels:  
\rewr{
\begin{equation}
        m_i^-, n_i^- \leftarrow \text{RandomSelect}\left(\cup_{\{s\neq s_x\}}\mathcal{O}_s\right).
\end{equation}
}
This ensures that trap objects belong to spaces other than \( v_x \), thereby enforcing the space-aware selection criterion.

\subsubsection{Character Hallucination}\label{sec:bench-char}
\begin{figure}[h]
    \centering
    \includegraphics[width=\columnwidth]{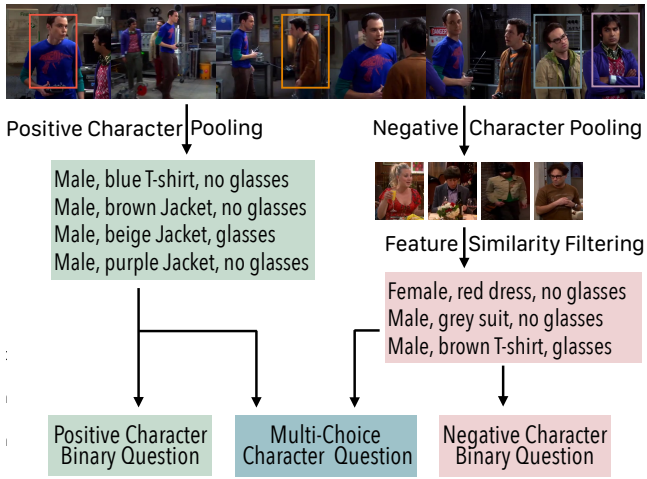} 
    \vspace{-0.4cm}
    \caption{Constructing character hallucination questions}
    \label{fig:pipeline_character}
    \vspace{-0.4cm}
    \Description{pipeline_character}
\end{figure}
In this section, we present how to construct hallucination questions for the \textbf{Characters with their Features}.
The detailed pipeline is illustrated in Figure~\ref{fig:pipeline_character}. \rewr{Human characters have been the subjects for most videos, despite their lower average density compared to objects. Therefore, precisely localizing and distinguishing different characters by their features are key to complex video understanding.} 
\add{We demonstrate this in detail in Appendix \ref{ap:mesh-hallu-char-uniq}.}

\textbf{Character Features in Videos:} 
In the TVQA+ dataset \cite{tvqa}, main characters appear without shielding or background blurring, allowing for clear identification of character features. As shown in Figure~\ref{fig:pipeline_character}, videos include multiple characters in various appearances, providing a basis for distinguishing each. 
For a video $v_x$, the $k$-th character is denoted as $c_{xk}$. We utilize GPT4o \cite{gpt4-2} to perform initial auto-detection of character features, with manual adjustments for errors. \add{Detailed annotation process is presented in Appendix \ref{ap:mesh-char}}. Features include gender $f_{ge}$, garment type $f_t$, garment color $f_c$, glasses $f_{gl}$, sleeve type $f_{sl}$, collar presence $f_{co}$, pocket presence $f_p$, and garment shade $f_{sh}$. Ambiguous cases are labeled as ``unknown." The $k$-th character in $v_x$ is represented by the feature vector:
\begin{equation}\label{eq:charf}
\begin{aligned}
    \mathbf{f}_{xk} = \{f_{ge}(c_{xk}), f_t(c_{xk}), f_c(c_{xk}), f_{gl}(c_{xk}), \\
    f_{sl}(c_{xk}), f_{co}(c_{xk}), f_{p}(c_{xk}), f_{sh}(c_{xk})\}.
\end{aligned}
\end{equation}

\textbf{Selection of Target and Trap Instances:} 
For video $v_x$, a subset of annotated features is selected to represent different hallucination levels (coarse, medium, fine), forming the target feature set $\{\mathbf{g}_{xk}\}_{k=1}^{l_{xc}}$, where $l_{xc}$ is the number of characters in the video.
To construct the trap candidate set $\mathcal{G}_\text{neg}$, features are randomly selected from videos other than $v_x$, and similarity metrics are applied to exclude indistinguishable instances:
\begin{align}\label{eq:gneg}
    \mathcal{G}_\text{neg} = \{\mathbf{g}_{yk} \mid v_y \neq v_x \text{ and } \mathbf{g}_{yk} \not\sim \mathbf{g}_{xk}, \forall y,k\}.
\end{align}
A suitable number of negative instances are then sampled from $\mathcal{G}_\text{neg}$ for binary and MC questions. \add{Detailed selection algorithms and similarity metrics are provided in Appendix \ref{ap:mesh-hallu-char-anno}.}

\subsubsection{Stage Hallucination}\label{sec:bench-stage}
\begin{figure}[h] 
    \centering
    \begin{subfigure}[b]{\columnwidth} 
        \includegraphics[width=\columnwidth]{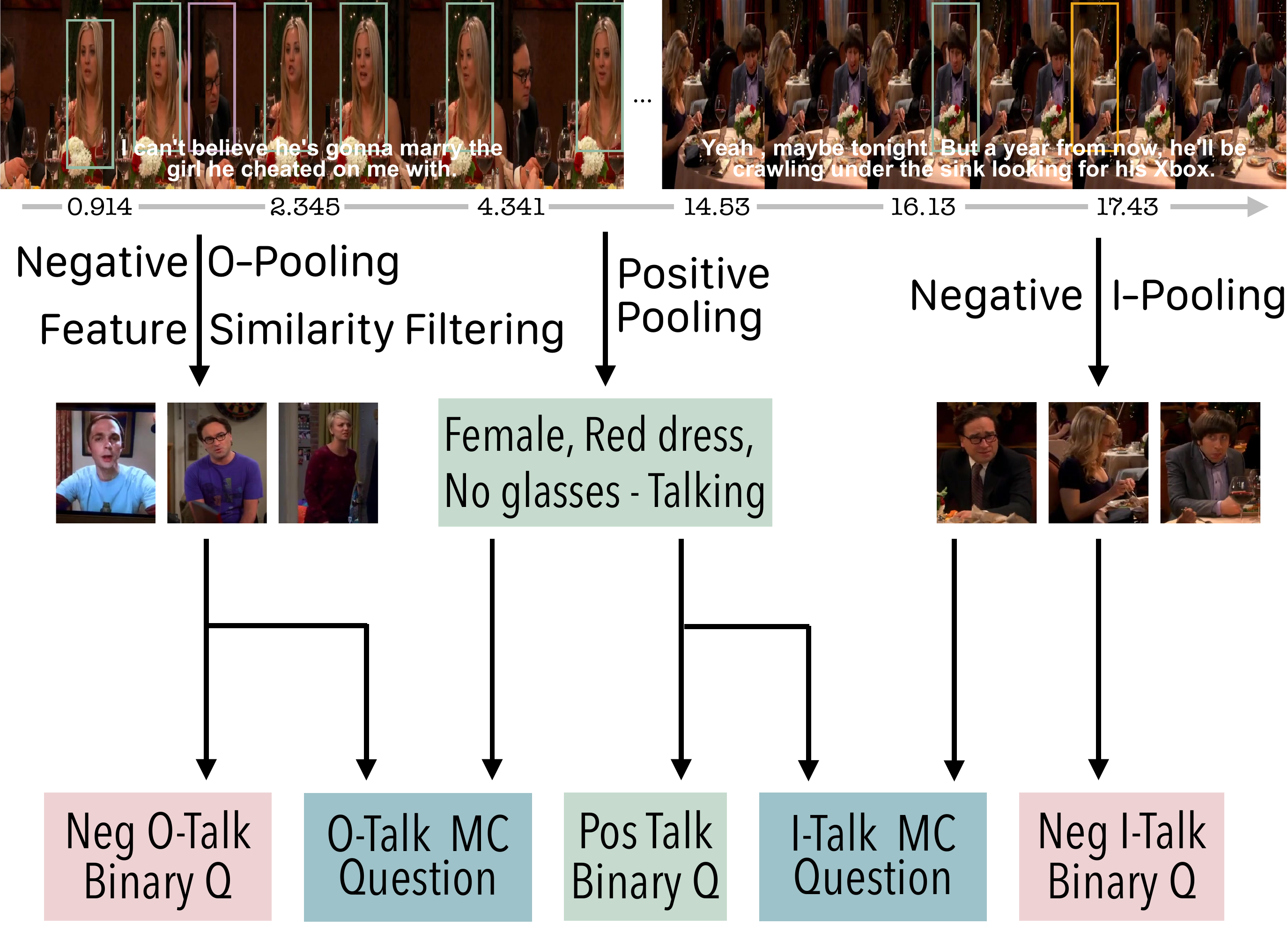}
        \subcaption{Hallucination questions involving dialogues}
        \label{fig:stage2}
    \end{subfigure}
    \begin{subfigure}[b]{\columnwidth} 
        \includegraphics[width=\columnwidth]{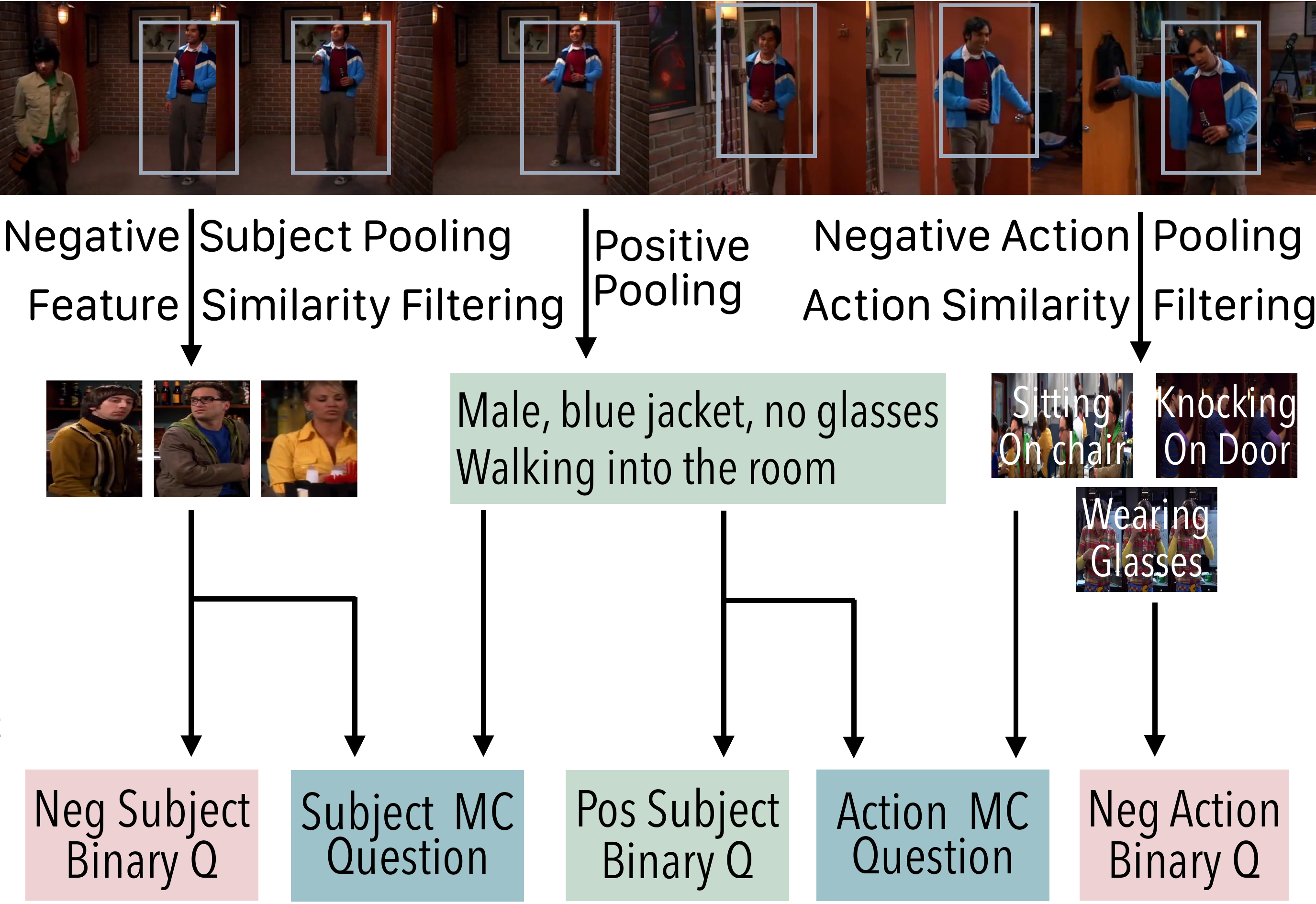}
        \subcaption{Hallucination questions involving actions}
        \label{fig:stage1}
    \end{subfigure}
    \vspace{-0.6cm}
    \caption{Constructing stage hallucination questions}
    \label{fig:pipeline_stage}
    \vspace{-0.4cm}
    \Description{Constructing stage hallucination questions}
\end{figure}
In this section, we describe the construction of hallucination questions for the \textbf{Stage with Actions/ Dialogues}. Actions conducted by a given character across multiple frames serve as unique key content in videos. 
\add{We illustrate its importance in video understanding in detail in Appendix \ref{ap:mesh-hallu-stage-uniq}.} 
\rewr{Based on the nature of actions, we devide them into \textbf{Dialogue-based} and \textbf{Action-based} ones.}
\textbf{Annotating Stage in Videos:}
For dialogues in Figure~\ref{fig:stage2}, we construct the \textit{subject-within-clip} set using subtitles with their start/end frames, along with character features:
\begin{equation}\label{eq:stage-dia}
    \{t_{xk}\}_{k=1}^{l_{xt}} \text{  where  } t_{xk} = \langle \mathbf{g}^{coarse}_{xk}, \mathrm{frame}_{\text{sta}}, \mathrm{frame}_{\text{end}} \rangle.
\end{equation}
For actions in Figure~\ref{fig:stage1}, we construct the \textit{subject-action} annotation set:
\begin{equation}
    \{a_{xk}\}_{k=1}^{l_{xa}} \text{  where  } a_{xk} = \langle \mathbf{g}^{coarse}_{xk}, \mathrm{act}_{xk} \rangle.
\end{equation}
$l_{xt}$ and $l_{xa}$ denotes the number of annotated talkings/actions.

\textbf{Selection of Target and Trap Instances:}
For video $v_x$, the target instances are derived from the dialogue-based annotated set $\{t_{xk}\}_{k=1}^{l_{xt}}$ (\textbf{ET}) and the action-based annotated set $\{a_{xk}\}_{k=1}^{l_{xa}}$ (\textbf{EA}). 
For each target instance, we select dissimilar character features $\mathbf{g}_{xk}^{{coarse}^-}$ either from the negative candidate set $\mathcal{G}_\text{neg}$ (Eq. \ref{eq:gneg}) (\textbf{CO}) or the other characters within the video (\textbf{CI}) and substitute $\mathbf{g}^{coarse}_{xk}$ with it to form a trap instance
\begin{equation}\label{eq:action-trap-dia}
t_{xk}^- = \langle \mathbf{g}_{xk}^{{coarse}^-}, \mathrm{frame}_{\text{sta}}, \mathrm{frame}_{\text{end}} \rangle
\end{equation}
for dialogue and for action pairs
\begin{equation}\label{eq:action-trap-act}
a_{xk}^- = \langle \mathbf{g}_{xk}^{{coarse}^-}, \mathrm{action}_{xk} \rangle.
\end{equation}
We also apply a substitution technique for action pairs. First, we construct the negative action candidate set: 
\begin{equation}\label{eq:action-neg}
    \mathcal{A}_\text{neg} =\{\mathrm{act}_{yk} \mid v_y \neq v_x   \And \mathrm{act}_{yk} \not\sim \mathrm{act}_{xk}, \, \forall y,k \}.
\end{equation}
Then, we either randomly select negative actions $\mathrm{act}_{xk}^-$ from $\mathcal{A}_\text{neg}$ (\textbf{AO}) or generate similar actions (\textbf{SA}) to substitute $\mathrm{act}_{ik}$, forming a trap instance:
$
a_{xk}^- = \langle \mathbf{f}_{xk}, \mathrm{act}_{xk}^- \rangle.
$
Besides, we also mix subject-action pairs by swapping their roles within the video (\textbf{MI}).
\add{The detailed selection algorithm and similarity metrics are provided in Appendix \ref{ap:mesh-hallu-stage-anno}.}

\section{Experiments with MESH}

We evaluate selected LVMs using MESH. For each video, we generate binary classification and multi-choice (MC) questions targeting setting, character, and stage perspectives to assess hallucination levels. 
\add{Detailed dataset statistics, LVM characteristics, and experimental technical details are provided in Appendix \ref{ap:ep-gd}.} In Section~\ref{section:results_analysis}, we present and analyze the results.
Through ablation studies in Section~\ref{section:ablation_studies}, 
we verify that MESH distinguishes LVMs based on multi-frame understanding abilities and examines the impact of hallucination levels on Video Question Answering (VQA) benchmarks.

\subsection{Experiments Details}\label{ep:metric}
\paragraph{Question format}
We use both binary and multi-choice (MC) formats. Binary question sets balance positive and negative instances, while MC questions pair one ground-truth with three negatives to assess multiple video contents efficiently, reducing LVM inference costs. MC also enhances clarity by reducing ambiguity in nuanced actions and improves model discrimination by minimizing subjective bias and random variability. Detailed analysis is in Appendix \ref{ap:metric-mc}.

\begin{table}[h!]
\vspace{-0.3cm}
\centering
\caption{Answer Distribution by Aspect and Category}
\label{tab:AnswerDistribution}
\vspace{-0.3cm}
\begin{tabular}{lll}
\toprule
\textbf{Aspect} & \textbf{Class} & \textbf{Category Counts} \\
\midrule
Setting   & Binary          & Yes:14663 \quad\quad\quad\quad No:13975 \\
          & MC & A:3646 \quad B:3651 \quad C:3598 \quad D:3713 \\
\midrule
Character & Binary          & Yes:10512 \quad\quad\quad\quad No:10512 \\
          & MC & A:5199 \quad B:5413 \quad C:5197 \quad D:5215 \\
\midrule
Action    &     Binary  &           Yes- EA:2491 \\
          &             &           No- AO:6518 \, CO:7131 \, SA:1200 \, MI:240 \\
          &      MC     &  AO- A:865 \:\:\: B:868 \:\,\: C:896 \:\,\: D:905 \\
          &             &  CO- A:1149 \, B:1177 \, C:1190 \, D:1121 \\
          &             &  SA- \,A:880 \:\:\: B:874 \:\:\: C:935 \:\,\: D:908 \\
          &             &  MI- \,A:59 \quad \, B:73 \;\;\;\;\:C:66 \;\;\;\;   D: 82 \\
\midrule
Dialogue          & Binary               & Yes-   ET: 2303 \\
                &                  & No-  AO: 5646 \: CO: 6756 \:  CI: 1450 \\
            & MC               & CO- A: 577 \quad B: 576 \quad C: 569 \quad D: 605 \\
              &                  &  CI- \,A: 563 \quad B: 566 \quad C: 588 \quad D: 610 \\
\bottomrule
\end{tabular}
\vspace{-0.6cm}
\end{table}

\paragraph{Dataset statistics}
\textbf{Setting} and \textbf{Character} datasets are structured as binary classification and multiple-choice tasks with balanced options.
\textbf{Stage} datasets contains dialogue/action questions with mixed negative sampling strategies as shown in \ref{sec:bench-stage}. Detailed distribution is shown in Table \ref{tab:AnswerDistribution}

\paragraph{Evaluation metric}
We use \textit{accuracy} to measure the proportion of correct answers in both formats. 
To assess if LVMs are biased toward specific options, we introduce additional balance metrics .
For binary questions, the \textit{positive rate (Pos)} reflects the model's tendency to answer ``yes''. In MC questions, \textit{Option Balance (OB)} and \textit{Correct Option Balance (COB)} assess bias across the four choices:
\begin{align}
    \text{OB} &= \sqrt{ \frac{1}{4} \sum_{i=1}^{4} \left( P_i - \overline{P} \right)^2 }, \quad
    \text{COB} = \sqrt{ \frac{1}{4} \sum_{i=1}^{4} \left( P_i^+ - \overline{P}^+ \right)^2 }
\end{align}
where $P_i = N_i/N$ and $P_i^+ = N_i^+/N^+$, with $N_i$ and $N_i^+$ denoting the number of total and correct predictions for option $i$, respectively. Lower OB and COB values indicate more balanced predictions.

\subsection{Experiments Results}\label{section:results_analysis}
\subsubsection{Setting Hallucination}
\begin{table}[h]
\vspace{-0.4cm}
\centering
\footnotesize
\caption{LVMs' performances on Setting for Binary/MC tasks.}
\label{tb:setting_results_compact}
\vspace{-0.3cm}
\begin{tabular}{lcccccc}
\toprule
\textbf{Model} & \multicolumn{2}{c}{\textbf{Binary (\%)}} & \multicolumn{3}{c}{\textbf{MC (\%)}} \\
\cmidrule(lr){2-3} \cmidrule(lr){4-6}
              & Acc & Pos & Acc & OB & COB \\\midrule
LLaVA-NV-32B\textsuperscript{\(\dagger\)} & 63.0 & 86.3 & 81.4 & 2.78 & 1.35 \\
Oryx-7B\textsuperscript{\(\ddagger\)} & 68.4 & 22.4 & 62.6 & 3.26 & 1.73 \\
VideoXL-7B\textsuperscript{\(\ddagger\)} & 77.1 & 50.2 & 73.0 & 0.93 & 0.67 \\
Oryx1.5-32B\textsuperscript{\(\mathsection\)} & 69.0 & 23.3 & 62.8 & 5.84 & 2.70 \\
Qwen2VL-72B\textsuperscript{\(\ddagger\)} & 78.4 & 31.4 & 86.3 & 1.52 & 0.90 \\
VideoLLaMA2.1-7B\textsuperscript{\(\ddagger\)} & 82.6 & 36.1 & 84.9 & 2.51 & 1.21 \\
LLaVA-OV-72B\textsuperscript{\(\ddagger\)} & 83.8 & 51.4 & 83.0 & 3.13 & 1.61 \\
LLaVA-Video-72B\textsuperscript{\(\ddagger\)} & \textbf{90.3} & 49.1 & 92.3 & 0.76 & 0.65 \\
InternVL2.5-78B\textsuperscript{\(\mathsection\)} & 86.9 & 40.1 & \textbf{94.9} & 0.71 & 0.52 \\
LongVILA-8B\textsuperscript{\(\sharp\)} & 80.7 & 83.5 & 76.2 & 13.7 & 8.13 \\
VILA1.5-8B\textsuperscript{\(\sharp\)} & 87.2 & 48.1 & 66.3 & 1.58 & 0.57 \\
LLaMA-VID-LV-7B\textsuperscript{\(\star\)} & 47.2 & 30.8 & 23.5 & 36.9 & 37.3 \\
LLaMA-VID-13B\textsuperscript{\(\star\)} & 80.1 & 65.6 & 54.2 & 19.7 & 9.76 \\
LLaVA-NV-7B\textsuperscript{\(\star\)} & 76.1 & 28.3 & 81.9 & 5.77 & 2.43 \\
Aria-23B\textsuperscript{\(\pounds\)} & 90.0 & 43.4 & 85.6 & 1.95 & 0.80 \\
VideoAgent\textsuperscript{\(\natural\)} & 72.4 & 51.0 & 66.4 & 3.10 & 2.08 \\
\hline
GPT-4o & \textbf{92.6} & 50.4 & 86.6 & 2.65 & 2.30\\
\bottomrule
\end{tabular}
\begin{minipage}{\columnwidth}
\scriptsize
\textbf{Backbones:} \(\dagger\) Qwen1.5, \(\ddagger\) Qwen2, \(\mathsection\) Qwen2.5, \(\sharp\) Llama3, \(\star\) Vicuna, \(\pounds\) Aria, \(\natural\) Agent.
\end{minipage}
\vspace{-0.6cm}
\end{table}
For each LVM, 32 frames are sampled, centered around the frame where the positive object appears.
Table \ref{tb:setting_results_compact} show that LLaVA-Video-72B and InternVL2.5-78B models outperform others in both binary classification and MC tasks. 
Performance variations among LVMs with different architectures, training strategies, and datasets underscore the influence of these factors on hallucination levels. For example, LLaVA-Video-72B exhibits superior performance compared to LLaVA-OV-72B, primarily due to the inclusion of additional video data during fine-tuning. 
Conversely, the improved LLM backbone in Oryx1.5-32B does not notably reduce hallucination compared to Oryx-7B, likely due to limitations in the training data or video compression module.
\rewr{In MC tasks, model accuracy generally aligns with binary task performance, except for LLaVA-NV-32B, whose bias toward positive answers amplifies hallucination issues.}
The OB and COB performances remain low, indicating minimal differences in option selection probabilities and confirming the robustness of our MC format in supplementary to the binary pipeline. 
\acm{
In all, our benchmark shows that state-of-art models (e.g., \textit{InternVL2.5}) exhibit reduced hallucination compared to earlier ones (e.g., \textit{Video-LLaVA}), suggesting that improvements in video understanding correlate with less hallucination in setting aspect. 
These gains stem from architectural changes, including stronger LLM backbones, and the cleaning of training data \cite{InternVL2.5}.
Besides, models with high downsampling ratios in their multi-modal adapters (e.g., \textit{LLaMA-VID}) perform worse than those with lower ratios (e.g., \textit{LLaVA-Video-7B}), likely due to loss of fine-grained visual details. Agent-based models like \textit{VideoAgent}, although equipped with strong LLMs, are limited by their external tools (e.g., CLIP \cite{clip}), leading to inferior performance.
Surprisingly, scaling LLM parameters shows diminishing returns. In multiple series (\textit{LLaVA-Video}, \textit{LLaVA-OV}, \textit{Qwen2VL}), 7B and 72B models perform similarly, with drastic performance drops only in very small models (e.g., 0.5B). A notable exception is \textit{InternVL2.5-78B}, which achieves best performance using a significantly larger vision encoder, indicating the vision encoder becomes the bottleneck at scale.
Long-video fine-tuning (e.g., \textit{LLaMA-VID-7B-Long-Video}) leads to degraded performance, suggesting it may hurt object-level perception. Models with sharing architecture and training data (e.g., \textit{LLaVA-Video} and \textit{LLaVA-OV}) 
witness significant performance difference due to video-specific finetuning, showing our benchmark captures video-specific hallucination challenges beyond image-based evaluation.}
\add{The supplementary experiment details and results are in Appendix \ref{ap:ep-set}}.

\subsubsection{Character Hallucination}

\begin{figure}[h]
    \centering
    \vspace{-0.6cm}
    \includegraphics[width=\columnwidth]{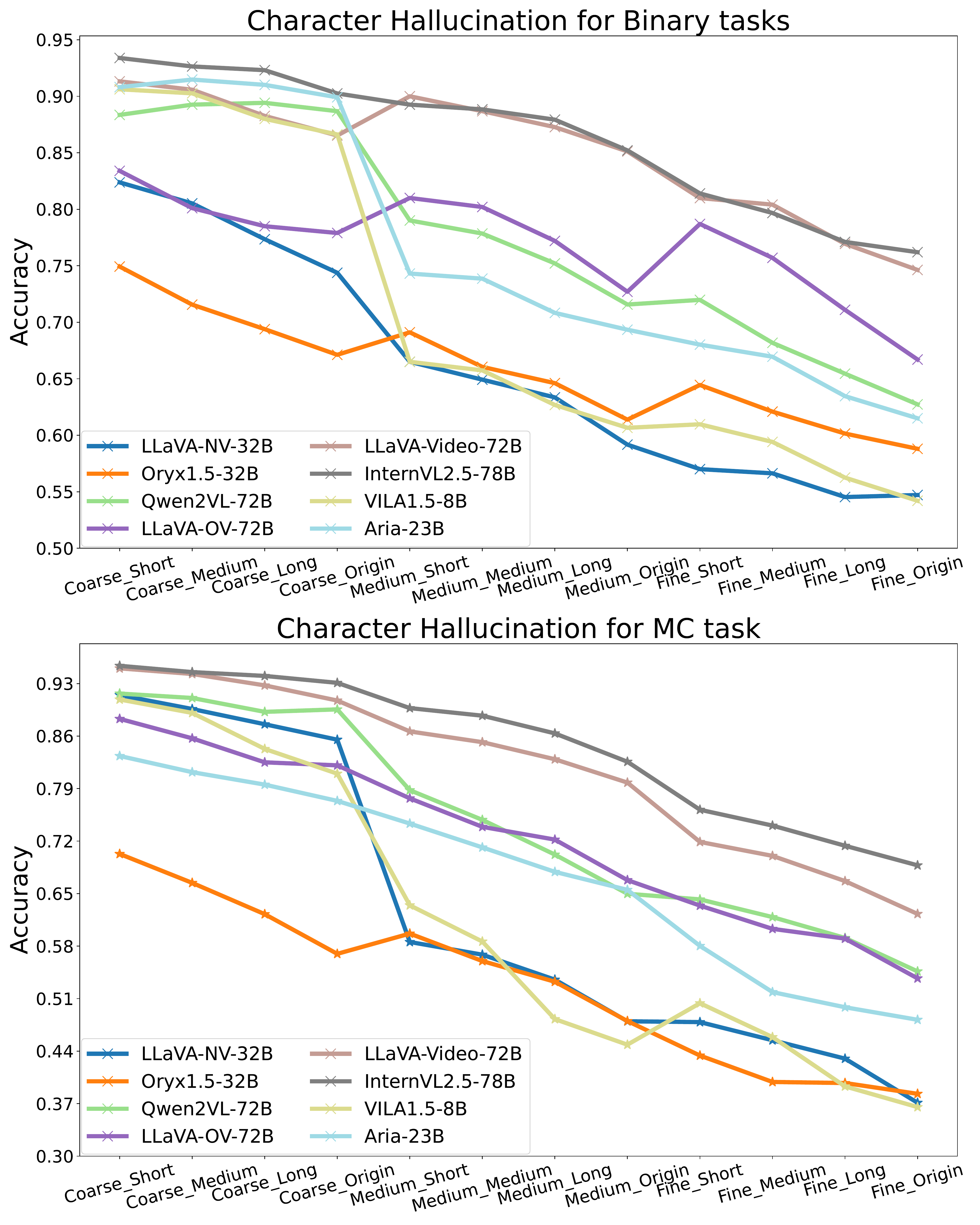} 
    \vspace{-0.7cm}
    \caption{Character Hallucination for Binary/MC tasks.}
    \label{fig:char_results_compact-under}
    \vspace{-0.3cm}
    \Description{char_results_compact}
\end{figure}

We design experiments at coarse, medium, and fine-grain levels to evaluate hallucination difficulties, ranging from general to detailed character features. We also measure hallucination levels using video clips of varying lengths: \textit{short} (8 frames), \textit{medium} (16 frames), \textit{long} (32 frames), and \textit{origin} (64 frames), centered on the frame with the largest bounding box for the character. 
Selected results for grain levels and clip lengths are summarized in Figure \ref{fig:char_results_compact-under}.
Overall, performance on \textit{character} aligns with \textit{setting}. LLaVA-Video-72B and InternVL2.5-78B consistently outperform others across grain levels and clip lengths. 
\rewr{LVMs show a steeper performance drop in MC tasks than binary tasks, mainly due to challenges in distinguishing multiple characters with fine-grained features across longer videos.}
Besides, performance declines significantly from coarse to fine-grain levels, demonstrating the increased difficulty of identifying detailed character features. 
\acm{
This challenge stems from: (1) Aggressive token reduction via pooling (e.g., 2×2 or 4×4 layers) \cite{oryx, lvlm-survey-1}, which sacrifices detail for efficiency
and (2) training data often lacks detailed person descriptions, relying mostly on coarse attributes like gender or clothing color \cite{llavavideo}.
(3) Fine character features may not be fully captured within a single frame, necessitating LVMs to aggregate features across multiple frames, thereby increasing complexity.}
Performance also deteriorates with longer clips, as additional frames often introduce noise rather than useful context \cite{reasoner}. 
Detailed analysis is provided in \ref{section:ablation_frame}.
\acm{
In all, LVLMs trade off long-context understanding for detail perception when handling fine-grained questions in long videos. Token reduction aids efficiency but reduces granularity, limiting top models like InternVL2.5 to below 0.7 accuracy on the hardest MC setting.
}
\acm{
Model size has little effect on simple tasks like \textit{Coarse-Short} but greatly impacts advanced tasks like \textit{Fine-Origin}, where larger models significantly outperform smaller ones. This suggests increasing parameter size could effective capture and retain fine-grained details in complex video scenarios.
}
\add{Additional results and analysis are in Appendix \ref{ap:ep-char}.}

\subsubsection{Stage Hallucination}
\begin{figure}[h]
    \centering
    \vspace{-0.5cm}
    \includegraphics[width=\columnwidth]{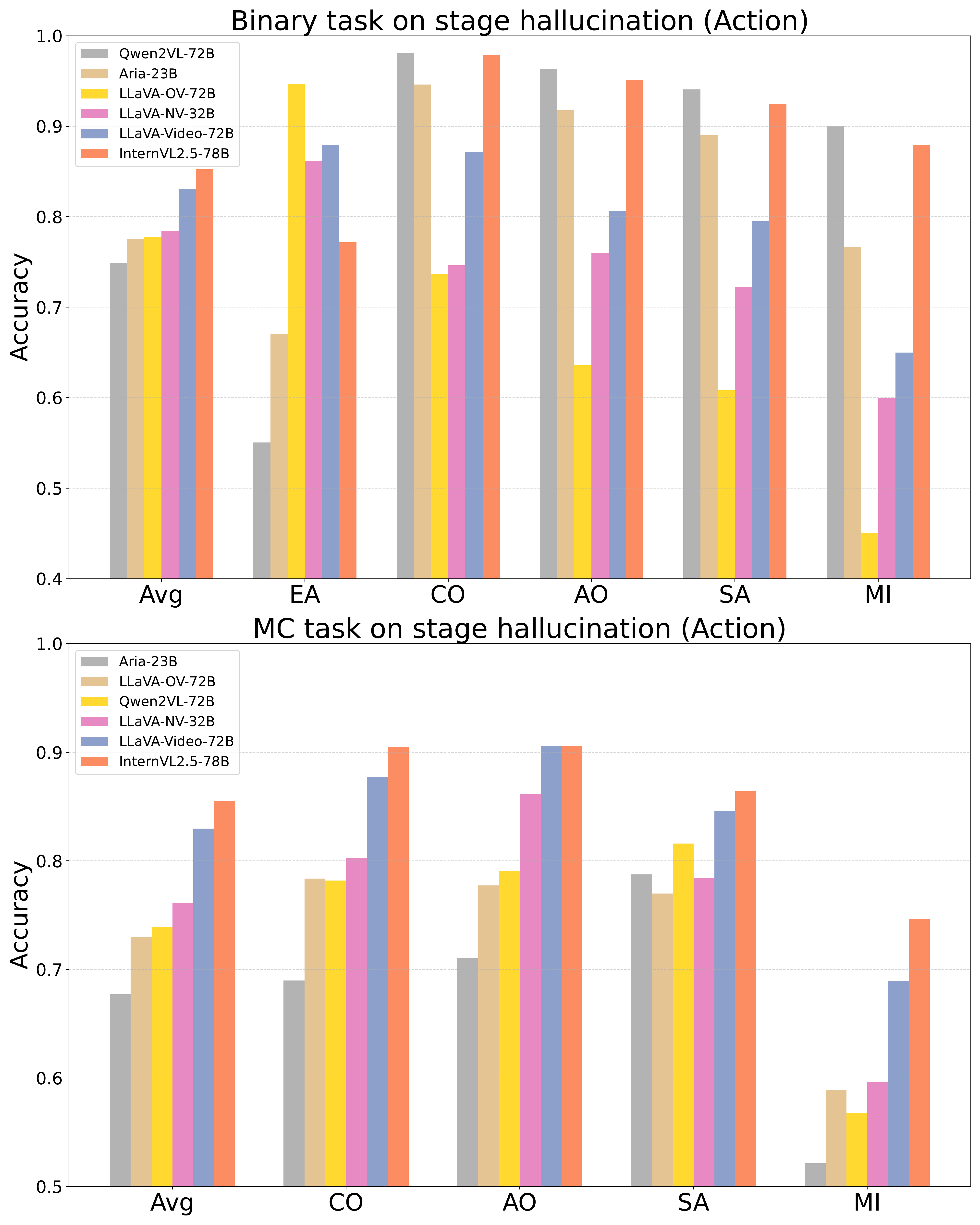}
    \vspace{-0.7cm}
    \Description{action_result}
    \caption{Action Hallucination for Binary/MC tasks}
    \label{fig:ordered_event_hallucination}
    \vspace{-0.2cm}
\end{figure}
\begin{figure}[h]
    \centering
    \vspace{-0.2cm}
    \includegraphics[width=\columnwidth]{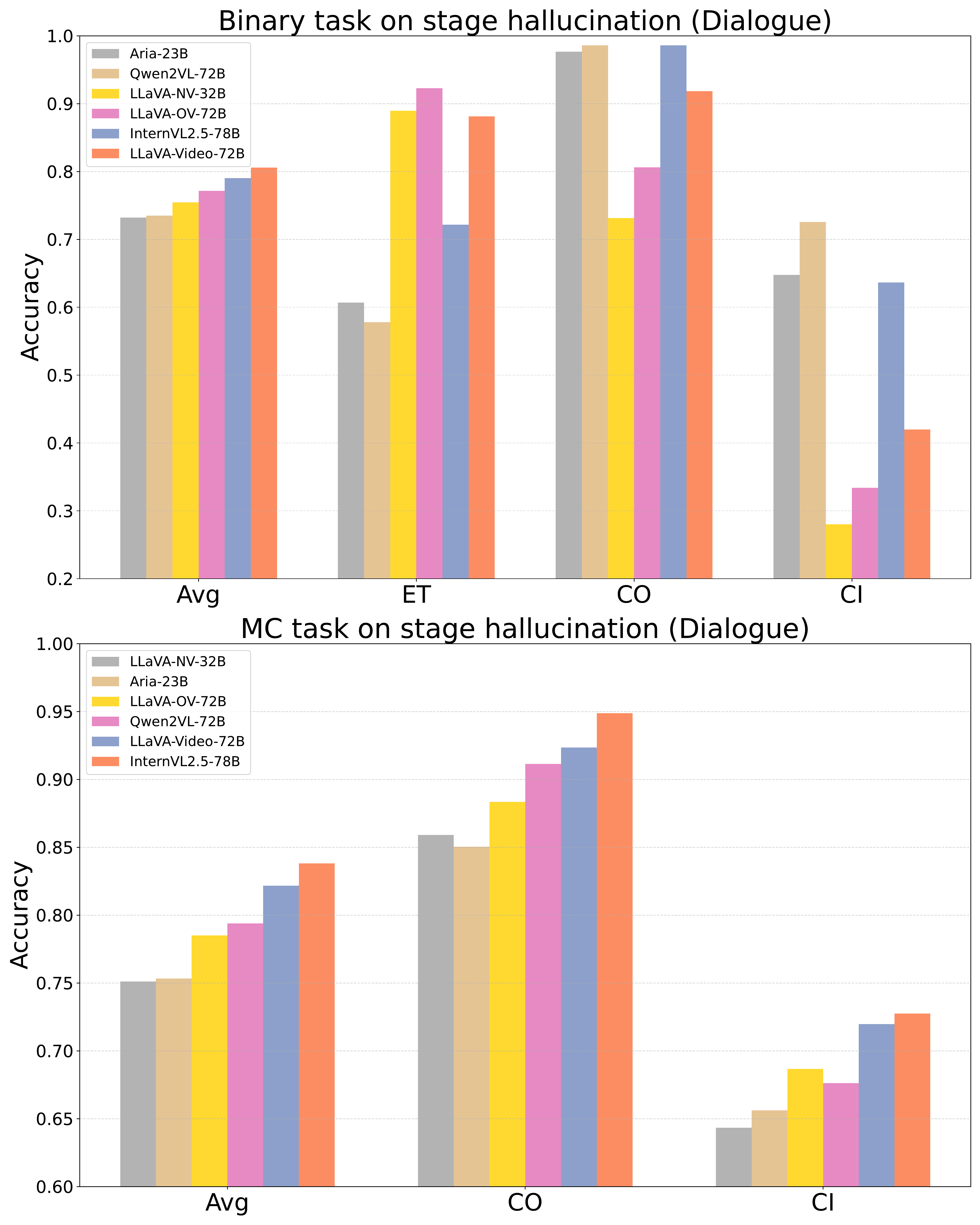} 
    \vspace{-0.7cm}
    \Description{talk_result}
    \caption{Dialogue Hallucination for Binary/MC tasks}
    \vspace{-0.2cm}
    \label{fig:reordered_talk_event}
\end{figure}
The stage hallucination problems relate subjects with coarse-level character features to their actions. We categorize the questions into \textit{Action} and \textit{Dialogue} types.
For \textbf{Action} tasks, video annotations for this category follow the format $S$ (subject with character features) + $A$ (action), e.g., ``a man wearing a red jacket ($S$) is walking on the street ($A$).'' 
\acm{
We identify and merge semantically equivalent actions in AO and CO via vector-based similarity, and use a generation pipeline to produce similar but distinct actions for SA.
}
Using positive action annotations, we extract the center action frame and sample 32 frames around it to construct clip instances.
\rewr{
Figure \ref{fig:ordered_event_hallucination} shows that average accuracy aligns with previous performance trends. 
}
\acm{
LVLMs perform worse on the CO category than on object or character alone, suggesting that combining objects and character into events adds complexity rather than clarity.
}
\rewr{
SA and MI tasks are more challenging than EA, CO, and AO, suggesting that LVMs struggle to differentiate similar actions and align multiple subject-action pairs within the same video. }
\acm{
One reason is the high variability of human actions. Unlike static objects with consistent forms, human could move and interact with objects in highly flexible ways. This complexity requires more video-language data for LVLMs to learn effectively, but current datasets remain insufficient \cite{llavavideo}.
}
Besides, models like InternVL2.5-78B exhibit biases in answering binary questions, performing well on negative questions but showing less confidence in positive EA predictions.
For \textbf{Dialogue} tasks, annotations for dialogue-based videos follow the format $S$ (subject: character features) + “is talking,” e.g., “a man wearing a red jacket ($S$) is talking.” 
Conversation clips are extracted at 1fps based on subtitle annotations, with variable frame lengths. 
\rewr{
Figure \ref{fig:reordered_talk_event} shows that LVMs perform well on ET and CO tasks in both binary and multiple-choice (MC) formats. 
However, CI tasks, which require accurately identifying the speaking character within a group, remain significantly more challenging. 
The sharp drop in CI performance highlights the limitations of current LVMs in recognizing speakers based solely on visual clues. 
} 
In both stage aspects, models in MC tasks remain stable across sections despite additional false options, aligning with their average performance and confirming the MC metric’s reliability.
\acmo{
By incorporating multiple actions' timestamps and contexts, we create extended questions that require LVMs to determine whether an action occurs before or after another action. Results are given in Appendix \ref{ap:tem-spa}. This allows us to further evaluate the models’ hallucinate on stage and their understanding of spatio-temporal relations.
}
\add{Additional results and analysis are in Appendix \ref{ap:ep-sta}.}

\subsection{Ablation study}\label{section:ablation_studies}
\subsubsection{Hallucination in images vs. videos}\label{section:ablation_frame}
We present the comparison with utilizing a single frame containing the related objects/ characters/motions to illustrate the uniqueness of MESH categorization.

\paragraph{\textbf{Setting: \textit{longer videos distract LVMs}}}
\begin{table}[h!]
\vspace{-0.2cm}
\caption{Setting for Image vs. Video}
\label{tb:ana-set}
\centering
\vspace{-0.4cm}
\begin{tabular}{lcccc}
\toprule
\textbf{Model} & \multicolumn{2}{c}{\textbf{Binary \%}} & \multicolumn{2}{c}{\textbf{MC \%}} \\
\cmidrule(lr){2-3} \cmidrule(lr){4-5}
              & Img. & Vid. & Img. & Vid. \\\midrule
LLaVA-Video-7B & 83.7  & \textcolor{red}{$\uparrow$}90.6  & 92.7  & \textcolor{blue}{$\downarrow$}92.1  \\
Aria-23B & 84.2  & \textcolor{red}{$\uparrow$}90.0  & 87.2  & \textcolor{blue}{$\downarrow$}85.6  \\
InternVL-8B & 79.5  & \textcolor{blue}{$\downarrow$}78.4  & 90.4  & \textcolor{blue}{$\downarrow$}82.4  \\
Qwen2VL-7B & 88.0  & \textcolor{blue}{$\downarrow$}85.9  & 86.2  & \textcolor{blue}{$\downarrow$}73.2  \\
Oryx-7B & 83.6  & \textcolor{blue}{$\downarrow$}68.4  & 89.6  & \textcolor{blue}{$\downarrow$}62.6  \\
LLaVA-NV-32B & 85.7  & \textcolor{blue}{$\downarrow$}63.0  & 85.6  & \textcolor{blue}{$\downarrow$}81.4  \\
\bottomrule
\end{tabular}
\vspace{-0.4cm}
\end{table}

For setting aspect, the image frame is sampled where the target object occupies its largest area. 
The result is in Table \ref{tb:ana-set}. 
In the binary task, LVMs with strong performance exhibit increased accuracy when prompted with more frames (Vid.), demonstrating their ability to leverage multi-frame information to reduce hallucination. Conversely, LVMs with poor performance experience a significant drop in accuracy, with declines reaching up to 20\%. In the MC task, all LVMs show a performance decrease, with the largest drops observed in underperforming models. This further confirms that our MC structure provides a more challenging and effective metric for measuring hallucination in localizing relevant frames.

\paragraph{\textbf{Character: \textit{frame continuity matters}}}
\begin{table}[h!]
\vspace{-0.4cm}
\caption{Character in Continu/Discrete frames (Binary)}
\label{tb:ana-char}
\centering
\vspace{-0.4cm}
\begin{tabular}{lcccc}
\toprule
\textbf{Model} & \multicolumn{2}{c}{\textbf{Continuous \%}} & \multicolumn{2}{c}{\textbf{Discrete \%}} \\
\cmidrule(lr){2-3} \cmidrule(lr){4-5}
              & 1 Fra. & 4 Fra. & 1 Fra. & 4 Fra. \\\midrule
LLaVA-Video-7B  & 77.8  & \textcolor{red}{$\uparrow$}78.8  & 77.8  & \textcolor{blue}{$\downarrow$}70.7  \\ 
Aria-23B            & 64.6  & \textcolor{red}{$\uparrow$}69.7  & 64.6  & \textcolor{blue}{$\downarrow$}62.6  \\ 
InternVL-8B     & 67.0  & \textcolor{blue}{$\downarrow$}62.9  & 67.0  & \textcolor{blue}{$\downarrow$}59.6  \\ 
Qwen2VL-7B      & 82.3  & \textcolor{blue}{$\downarrow$}76.3  & 82.3  & \textcolor{blue}{$\downarrow$}67.7  \\ 
Oryx-7B         & 68.7  & \textcolor{blue}{$\downarrow$}59.6  & 59.0  & \textcolor{blue}{$\downarrow$}46.6  \\ 
LLaVA-NV-32B & 68.2  & \textcolor{blue}{$\downarrow$}61.6  & 68.2  & \textcolor{blue}{$\downarrow$}61.1  \\ 
\bottomrule
\end{tabular}
\vspace{-0.4cm}
\end{table}
For the character aspect, we use fine-grained features to evaluate LVMs’ ability to associate relevant frames for prediction. 
The result is in Table \ref{tb:ana-char}. 
Continuous frames are sampled around the center frame, while discrete frames are uniformly sampled across the video. With more frames, LLaVA-Video and Aria demonstrate higher accuracy, leveraging multi-frame tokens to better identify characters, while other LVMs are distracted by spanning frames. In contrast, with discrete sampling, all LVMs show reduced accuracy, suggesting irrelevant frames introduce noise when propagated through frame tokens.

\paragraph{\textbf{Stage: \textit{seeing actions needs multiple frames}}}
\begin{table}[h!]
\vspace{-0.4cm}
\caption{Stage (Action) for Image vs. Video}
\label{tb:ana-sta}
\centering
\vspace{-0.4cm}
\begin{tabular}{lcccc}
\toprule
\textbf{Model} & \multicolumn{2}{c}{\textbf{Binary \%}} & \multicolumn{2}{c}{\textbf{MC \%}} \\
\cmidrule(lr){2-3} \cmidrule(lr){4-5}
              & 1 Fra. & 32 Fra. & 1 Fra. & 32 Fra. \\\midrule
LLaVA-Video-7B  & 70.5  & \textcolor{red}{$\uparrow$}80.9  & 60.1  & \textcolor{red}{$\uparrow$}82.3  \\ 
Aria-23B            & 68.5  & \textcolor{red}{$\uparrow$}77.5  & 67.1  & \textcolor{red}{$\uparrow$}67.7  \\ 
InternVL-8B     & 67.9  & \textcolor{red}{$\uparrow$}71.0  & 56.7  & \textcolor{red}{$\uparrow$}76.6  \\ 
Qwen2VL-7B      & 72.8  & \textcolor{red}{$\uparrow$}78.3  & 57.1  & \textcolor{red}{$\uparrow$}67.2  \\ 
Oryx-7B         & 68.7  & \textcolor{blue}{$\downarrow$}59.6  & 59.0  & \textcolor{blue}{$\downarrow$}46.6  \\ 
LLaVA-NV-32B & 68.2  & \textcolor{red}{$\uparrow$}78.4  & 59.3  & \textcolor{red}{$\uparrow$}76.1  \\ 
\bottomrule
\end{tabular}
\vspace{-0.3cm}
\end{table}

For stage aspect, we compare results between prompting a single action-centered frame and 32 frames. \acm{Understanding actions requires LVMs to integrate tokens across multiple frames. Table \ref{tb:ana-sta} shows that a single frame is insufficient for accurate judgment, highlighting MESH’s uniqueness in video understanding.} Furthermore, high-performing LVMs achieve significant gains with longer video prompts, while low-performing LVMs show limited improvement. This demonstrates that our stage hallucination question set effectively distinguishes LVMs based on their ability to capture chronological action information across the entire video.

\subsubsection{Hallucination level aligns with VQA}\label{section:ablation_vqa}
\begin{table}[h!]
\vspace{-0.3cm}
\caption{LVMs' on MESH vs. VQA tasks}
\label{tb:ana-vqa}
\centering
\vspace{-0.4cm}
\begin{tabular}{lcccc}
\toprule
\textbf{Model} & \multicolumn{2}{c}{\textbf{MESH \%}} & \multicolumn{2}{c}{\textbf{VQA \%}} \\
\cmidrule(lr){2-3} \cmidrule(lr){4-5}
              & Base & Adv & VMME & MLVU \\\midrule
InternVL2.5-78B  & \textbf{90.1}  & \textbf{73.4}  & \textbf{72.1}  & \textbf{75.7}  \\ 
LLaVA-Video-72B  & 89.5 & 70.1  & 70.6  & 74.4  \\ 
LLaVA-OV-72B     & 81.9  & 62.4  & 66.3  & 68.0  \\ 
Aria-23B      & 81.9  & 57.0  & 67.6  & 70.6  \\ 
LLaVA-NV-32B & 76.7  & 54.0  & 60.2  & 65.5  \\ 
\hline
GPT-4o & 83.8  & 58.7  & 71.9  & 64.6\\
\bottomrule
\end{tabular}
\vspace{-0.4cm}
\end{table}
We calculate base and advanced hallucination levels based on question difficulties (see Appendix \ref{ap:ep-diff}). Table \ref{tb:ana-vqa} shows that hallucination levels measured by MESH align with VQA benchmarks (V-MME \cite{vqa-1} and MLVU \cite{mlvu}), despite their distinct focus. 
\rewr{LVMs with advanced video element detection, from basic to fine details, are better at answering complex questions accurately. 
However, most LVMs struggle more with advanced hallucination tasks than with VQA tasks. 
This indicates that fine-grained hallucination questions remain underutilized in VQA. 
\acm{
We also show in Appendix \ref{ap:case-study} that when LVMs hallucinate on MESH questions, they often produce incorrect or ambiguous video descriptions.
}
Incorporating finer hallucination aspects could enhance high-level VQA task design.}
\section{Conclusion}
In this work, we introduce a bottom-up approach, inspired by human video cognition, to assess LVM hallucinations across setting, character, and stage perspectives (\miset).
We enhance evaluation with binary and multi-choice questions.
\acm{Using this framework, we created MESH, a comprehensive benchmark building upon TV series with complex human activities spanning basic to advanced difficulty levels.}
Extensive experiments on diverse LVMs configurations revealed that while these models excel at recognizing basic elements from a single frame, their susceptibility to hallucinations increases significantly when handling fine character details or associating multiple actions across frames. 
We also demonstrate MESH effectively differentiates LVMs capable of leveraging multi-frame tokens to detect fine features and actions, with stronger models performing better in VQA and captioning tasks. These findings confirm the effectiveness and stability of our evaluation pipeline and benchmark.

\bibliographystyle{ACM-Reference-Format}
\bibliography{main}

\appendix
\appendix
\onecolumn
\section{Human understanding videos}\label{ap:human-under}
Video, first captured and screened by the Lumière brothers in the late 19th century, evolved from silent image sequences to include sound, enhancing its multimodal nature \cite{re-v1-1}.
Video understanding, as a methodology combining visual, auditory, and temporal elements, has evolved significantly over time. Research has primarily focused on processing and interpreting these modalities by humans and machines \cite{lvlm-survey-1, lvlm-survey-2, re-v1-2}. Our benchmark specifically examines human understanding of the visual components of videos, excluding audio. It focuses on \textit{mise-en-scène}, which refers to all elements captured in videos \cite{human-under-1, human-under-2}. In video comprehension, \textit{mise-en-scène} is categorized into three domains: \textbf{setting}, \textbf{people and their appearance} (e.g., clothing, makeup), and \textbf{stage} (e.g., actions, talking).
\begin{itemize}
    \item \textbf{Setting} refers to the physical location where the video takes place, as well as the objects present within the scene. These settings provide crucial context for understanding the narrative and the relationships between people and their environment \cite{human-under-1, human-under-3}.
    \item \textbf{People and Their Appearance} involve the individuals featured in the video, with particular attention to their physical characteristics, gender, clothing, and makeup. These elements contribute to the portrayal of their identity \cite{human-under-2}.
    \item \textbf{Stage} encompasses the actions and movements of both the characters and objects within the scene, including conversations, which require the integration of temporal information to understand the sequence of events. The dynamic nature of these is critical in establishing the progression flow of video contents \cite{human-under-3}.
\end{itemize}
Human perception of video content typically follows a process in which the viewer first identifies the environment and objects in the opening frame \cite{human-under-1, human-under-2}. Attention is often directed toward the people featured in the scene, particularly their appearance and behaviors \cite{human-under-6}. As the video progresses, viewers become attuned to changes in subsequent frames, identifying who is de livering words and subtle shifts that introduce new information or obscure prior details. This process of dynamic attention and interpretation aligns with the principles of human video comprehension \cite{human-under-4, human-under-5}. 
The dataset developed for our study is designed to assess the ability of LVMs in avoiding hallucination results to replicate these aspects of human video understanding. By focusing on the elements of \textit{mise-en-scène}, we aim to evaluate how effectively LVMs can recognize and process the visual components that humans instinctively interpret when viewing videos.

\section{MESH Benchmark}\label{ap:bench}

\subsection{Metric}
\subsubsection{Why using multi-choice questions?}\label{ap:metric-mc}

\begin{figure}[h]
    \centering
    \includegraphics[width=0.9\columnwidth]{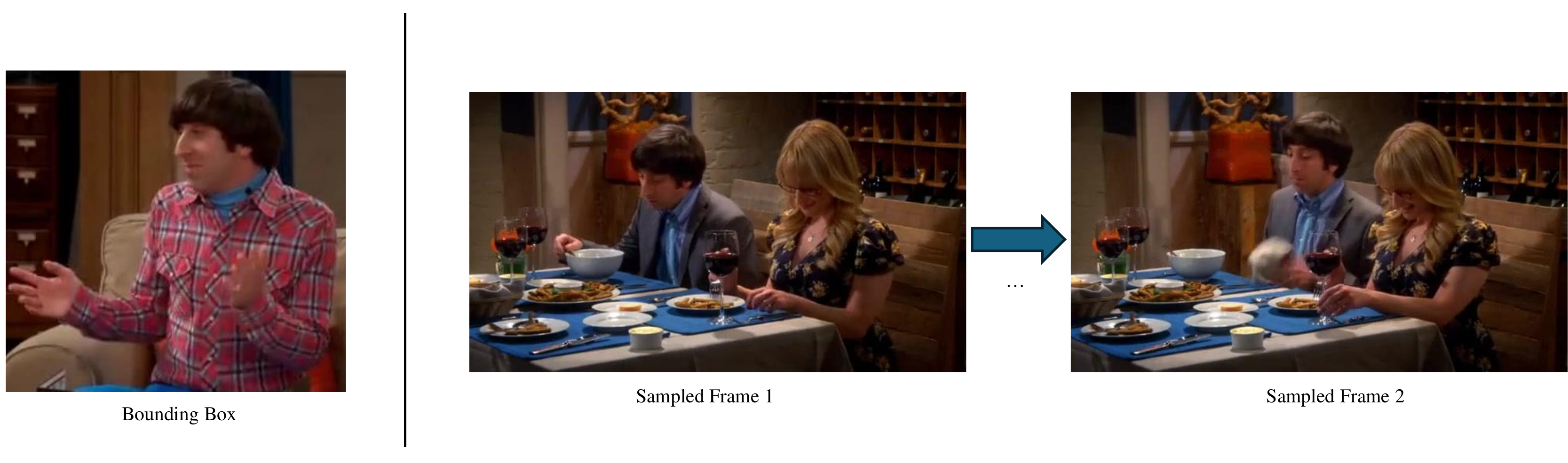} 
    \caption{Left: "A man wearing long-sleeved collared pink button-up shirt with pockets and not wearing glasses appears in the video". Right: "A man wearing a gray jacket picks up his napkin under the table.".}
    \label{fig:ambiguity}
    \vspace{-0.2cm}
    \Description{fig/ambiguity}
\end{figure}

\paragraph{Efficiency} Multi-choice questions offer enhanced efficiency in evaluation by aggregating more information per question compared to binary formats. While binary questions assess a single statement about the video, multi-choice questions can simultaneously examine a model's understanding across multiple facets of the video content through a single question. Considering the substantial computational cost associated with LVLM inference, and recognizing that current LVLM architectures and implementations \cite{lmdeploy, vllm, huggingface} often face limitations in handling the extensive context length required for processing long-form video content, minimizing the total number of questions becomes paramount. Therefore, the ability of multi-choice questions to consolidate assessment points significantly reduces the overall resource expenditure and time required for comprehensive evaluation.

\paragraph{Clearness} Binary questions are inherently susceptible to ambiguity, posing challenges for precise LVLM evaluation. One primary source of ambiguity stems from the interpretation of features, particularly when moving beyond basic object recognition. Empirical evidence suggests that object recognition benchmarks are approaching saturation for advanced LVLMs, necessitating the use of finer-grained features to discern subtle differences in model performance. However, describing these more nuanced features often introduces ambiguity. For example, consider the statement about the pink shirt (Figure \ref{fig:ambiguity} left). While the dominant color is demonstrably pink, one could argue that the presence of gray and black threads undermines the statement's precision, making its binary correctness debatable. Another source of ambiguity arises from the inherent nature of LVLMs and their reliance on frame sampling for video understanding. This sampling can lead to a lack of perceived motion continuity, creating ambiguity in action interpretation. In the napkin example (Figure \ref{fig:ambiguity} right), the statement implies a self-initiated action, yet the limited temporal resolution could equally support interpretations of the man receiving the napkin from another person. Such ambiguities render the accuracy of binary tasks susceptible to subjective interpretations, potentially obscuring the objective evaluation of the LVLM's abilities. In contrast, the inherently contrastive nature of multi-choice questions mitigates this issue. By requiring models to select the most appropriate statement from a set of options, rather than judging absolute correctness, multi-choice questions reduce the impact of subjective interpretations and focus the evaluation on comparative understanding.

\paragraph{Discrimination} In binary question formats, the potential range of accuracy scores for a model is limited, theoretically spanning from chance performance (50\%) to perfect accuracy (100\%). This restricted range limits the ability to effectively differentiate between models with similar performance. When models exhibit closely aligned performance on binary tasks, their scores become nearly indistinguishable within this narrow interval, and accuracy measurements become more susceptible to the influence of random noise and minor variations in question design. In contrast, for multi-choice questions with 'n' options, the theoretical accuracy range expands significantly, from chance performance (approximately 100\%/n) to perfect accuracy (100\%). This broader spectrum of possible accuracy scores provides a greater capacity to discern subtle performance differences between models and reduces the relative impact of random fluctuations, leading to a more robust and reliable evaluation of LVLM capabilities.

\subsubsection{Additional metric: JSD}\label{ap:metric-jsd}
In the following detailed results, we further employ Jensen-Shannon Divergence (JSD) \cite{jsd} to measure the option distribution between ground-truth and prediction. We have
\[
\text{JS}(P \| Q) = \frac{1}{2} \sum_{x} P(x) \ln \frac{P(x)}{M(x)} + \frac{1}{2} \sum_{x} Q(x) \ln \frac{Q(x)}{M(x)},
\]
where \( M(x) = \frac{1}{2}(P(x) + Q(x)) \). 
Here, \( P(x) \) represents the ground-truth distribution, and \( Q(x) \) represents the predicted distribution. For binary tasks, \( x \in \{\text{"yes"}, \text{"no"}\}\), while for multiple-choice tasks, \( x \in \{\text{A}, \text{B}, \text{C}, \text{D}\} \).We normalize the JSD for the binary task by dividing by 0.216 and for the MC task by dividing by 0.38.

\subsection{Setting hallucination}\label{ap:mesh-hallu-set}

\subsubsection{Uniqueness of Setting}\label{ap:mesh-hallu-set-uniq}
Unlike image datasets like MSCOCO \cite{mscoo}, where annotated objects are present in nearly all frames, video annotations often miss objects in many frames. Motion blur and the complexity of real-world scenarios further complicate annotating all objects within a single frame \cite{video-set-1, video-set-2}, making manual annotation both costly and time-intensive. As discussed in Section \ref{ap:human-under} and Appendix \ref{ap:human-under}, humans comprehend videos by first identifying environments through object sets. For instance, 
in Figure 3,
items like wineglasses, flowers, and plates provide context about the setting, which influences our overall understanding of subsequent frames. To address sparse object annotations and leverage environment labels, we associate unique object sets with their corresponding environments.

\subsubsection{Object Annotation}
The object names depicted in the video clips are extracted from the TVQA+ dataset utilizing Deepseek-v3 \cite{deepseekv3}. Spatial grounding information within the TVQA+ dataset is leveraged to procure object names within the video content that are lifeless and discernible, such as "bed," while excluding terms like "mass," "mother," and "melting." In order to enhance the efficacy of the LLM, a comparative approach is employed for the extraction process. The methodology details and an example are presented in Table \ref{tab:pruning_algorithm} and Algorithm \ref{alg:object_annotation}.

\begin{algorithm}[ht]
	\caption{Object Annotation Filter} 
        \label{alg:object_annotation}
	\begin{algorithmic}[1]
        \Statex \textbf{Input:} Video $v_i$'s object annotation set $S_{o} = \{o_1, o_2, ..., o_n\}$
        \State $ A_{o} \leftarrow [o_r, o_1, o_2, ..., o_n]$, where $o_r$ is the reference object.
        \State $ A_{o} \leftarrow $ LLM ("... Sort the list by visibility and materiality of each element in $ A_{o}$ ...")
        \State $I_r\leftarrow \text{index} (A_{o}, o_{r})$.
        \State $S_{o_{taget}} \leftarrow A_{o} [:I_r]$
        \Statex \textbf{Output:} Target object set $S_{o_{target}} = \{o_{1'}, o_{2'}, ..., o_{n'}\}$
	\end{algorithmic} 
\end{algorithm}

\begin{table}[ht]
\centering
\caption{Concrete Example of Object Annotation Filter.}
\label{tab:pruning_algorithm}
\begin{tabular}{lp{7cm}p{8cm}}
\midrule  
\textbf{Step} & \textbf{Procedure} & \textbf{Resultant State} \\
\midrule  
1 & Video $v_i$'s object annotation set & Original set: \{\text{Crowd}, \text{Mass}, \text{Football}\} \\
\midrule
2 & Introduce the reference object \textit{Pepper} to the list. & Augmented list: [\text{Crowd}, \text{Mass}, \text{Football}, \text{Pepper}] \\
\midrule  
3 & Invoke the LLM to rank objects by the visibility and materiality . & LLM-sorted list: [\text{Football}, \text{Pepper}, \text{Mass}, \text{Crowd}] \\
\midrule  
4 & Identify the index of \textit{Pepper} and prune all terms behind the index. & Post-pruning set: \{\text{Football}\} \\
\midrule 
\end{tabular}
\end{table}

\subsubsection{Selecting Setting}\label{ap:mesh-hallu-set-anno}

\begin{figure}[ht]
  \centering
  \begin{subfigure}{0.45\textwidth}
    \centering
    \includegraphics[width=\linewidth]{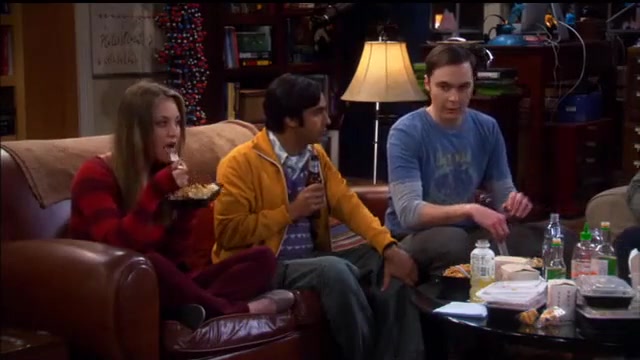} 
    \label{fig:left}
  \end{subfigure}
  \hfill
  \begin{subfigure}{0.45\textwidth}
    \centering
    \includegraphics[width=\linewidth]{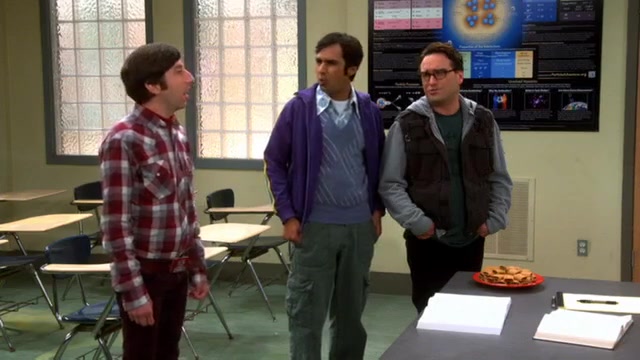} 
    \label{fig:right}
  \end{subfigure}
  \caption{Left: Representative frame from clip a depicting \textbf{A} domestic living room environment, Right: Representative frame from clip \textbf{B} illustrating a classroom setting.} 
  \label{fig:SettingExample}
\end{figure}

\begin{figure}
    \centering
    \includegraphics[width=\linewidth]{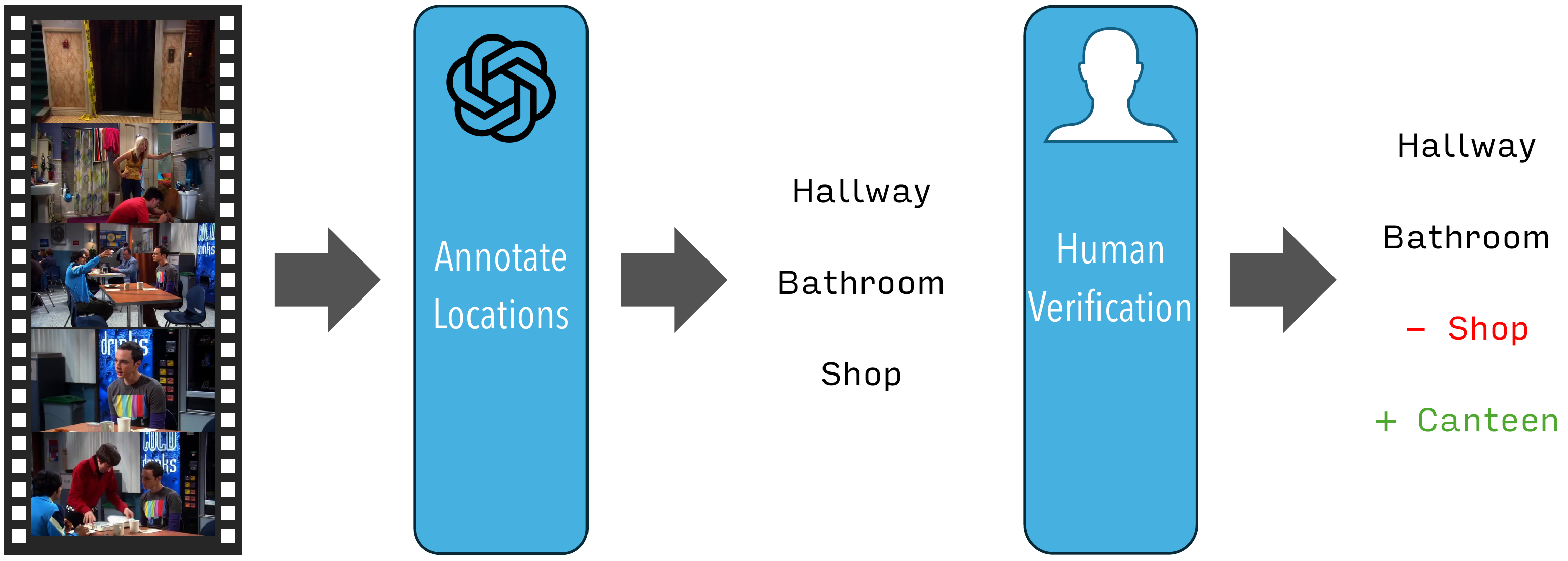}
    \caption{Pipeline for labeling the locations in the video.}
    \label{fig:locationAnnotationPipeline}
\end{figure}

The location of the scene within each video clip is extracted utilizing Int4 quantization version of LVM Qwen2-VL \cite{qwen2vl}. To account for potential variations in naming and hallucination output, human annotators unify names and correct the wrong content. 
\acm{This process is shown in Figure \ref{fig:locationAnnotationPipeline}.}
For each extracted object and its corresponding location, a location that appears in other video clips but not in the current one is selected. Subsequently, an open source LLM DeepSeek-v3 \cite{deepseekv3} is employed to identify five objects that are only present in the chosen location. These five objects are then used to formulate questions that inquire about the existence of objects not present in the scene. 
As Figure \ref{fig:SettingExample} shows, in the video clip \textbf{A}, the only visible scene is a living room containing a chair. By contrast, the classroom scene appears in clip \textbf{B} but not in clip \textbf{A}. To identify objects that are typically found in classrooms but rarely in living rooms, LLM is queried and it generates terms such as ``whiteboard'' and ``teacher’s podium,'' which align with the specified criteria. The details are presented in Algorithm \ref{alg:trap_object}.

\begin{algorithm}
	\caption{Trap Objects Selection} 
        \label{alg:trap_object}
	\begin{algorithmic}[1]
        \Statex \textbf{Input:} Video $v_i$'s location annotation set $S_{L} = \{L_1, L_2, ..., L_n\}$, Location set $\mathbb{L}$ that contains all possible locations, Number of iteration $n_t$.
	\For {$iteration=1,2,\ldots, n_t$} \Comment{Outer Loop}
            \State $L_{trap}\leftarrow \text{random}(\mathbb{L})$.
                \For {$L_j$ in $\{L_1, L_2, ..., L_n\}$}
                    \If {$L_j \approx L_{trap}$}
                        \State Continue the Outer Loop.
                    \EndIf
                \EndFor
            \State $o_{trap} \leftarrow$ LLM("...Choose objects (if exist) that is possible to appear in $L_{trap}$ but never appear in $L_1, L_2, ...,$ and $ L_n$ ...")
            
        \EndFor
        \Statex \textbf{Output:} Trap object set $S_{o_{trap}} =  \{o_{trap_1}, o_{trap_2}, ..., o_{trap_n}\}$
	\end{algorithmic} 
\end{algorithm}

\subsection{Character hallucination}\label{ap:mesh-hallu-char}

\begin{figure}
    \centering
    \includegraphics[width=\linewidth]{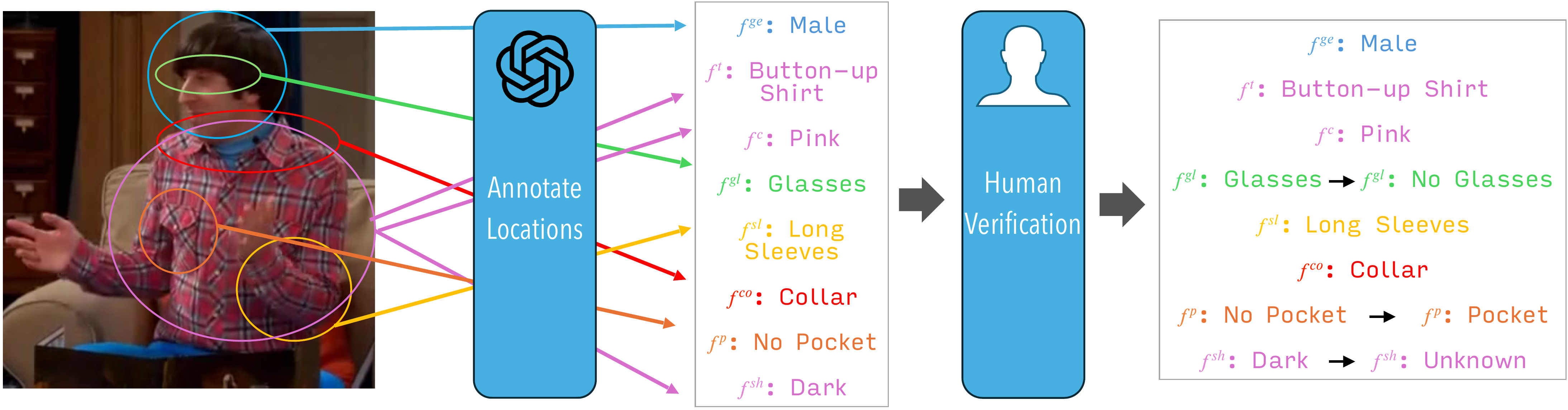}
    \caption{Pipeline for labeling the character features in the video.}
    \label{fig:personAnnotationPipeline}
\end{figure}

\begin{algorithm}
	\caption{Person Hallucination Questions Construction} 
        \label{alg:personAnnotationPipeline}
	\begin{algorithmic}[1]
        \Statex \textbf{Input:} $v_i$'s person set $S_{P} = \{P_1, P_2, ..., P_n\}$, Feature pool $[\mathbb{F}^{ge}, \mathbb{F}^{t}, \mathbb{F}^{c}, \mathbb{F}^{gl}, \mathbb{F}^{sl}, \mathbb{F}^{co}, \mathbb{F}^{p}, \mathbb{F}^{sh}]$,  Granularity mask $M = [m^{ge}, m^{t}, m^{c}, m^{gl}, m^{sl}, m^{co}, m^{p}, m^{sh}]$, Number of Iteration $n_t$.
        \For {$P_j$ in $ S_{P} = \{P_1, P_2, ..., P_n\}$}
            \State $P_j \leftarrow M \star P_j$
        \EndFor
        \For {$iteration=1,2,\ldots, n_t$} \Comment{Loop 1}
            \State $P_{trap} \leftarrow \text{random}(S_P)$.
            \For {$f^j_{trap}$ in $P_{trap} = [f^{ge}_{trap}, f^{t}_{trap}, f^{c}_{trap}, f^{gl}_{trap}, f^{sl}_{trap}, f^{co}_{trap}, f^{p}_{trap}, f^{sh}_{trap}]$}
                \If {$f^j_{trap}$ \text{is not masked}}
                \State $f^j_{trap} \leftarrow \text{random} (\mathbb{F}^j)$
                \EndIf
            \EndFor
            \For {$P_k$ in $ S_{P} = \{P_1, P_2, ..., P_n\}$} \Comment{Loop 2}
                \For {$f_k^j$ in $P_k = [f^{ge}_k, f^{t}_k, f^{c}_k, f^{gl}_k, f^{sl}_k, f^{co}_k, f^{p}_k, f^{sh}_k]$}
                    \If {$f_{trap}^j \not \approx f_k^j$}
                        \State Continue the Loop 2
                    \EndIf
                    \State Continue the Loop 1
                \EndFor
            \EndFor
        \EndFor
        \Statex \textbf{Output:} Target person set $S_{P_{target}} = \{P_1, P_2, ..., P_n\}$, Trap person set $S_{P_{trap}} = \{P_{trap_1}, P_{trap_2}, ..., P_{trap_{n'}}\}$
	\end{algorithmic} 
\end{algorithm}

\subsubsection{Uniqueness of Character}\label{ap:mesh-hallu-char-uniq}
As discussed in Appendix \ref{ap:human-under}, characters have been central to most videos since the medium’s inception, despite their lower average density compared to objects. Videos primarily convey information through human appearances and movements \cite{human-under-1, human-under-2}. In videos, characters naturally and dynamically appear and disappear (cut-ins and cut-offs) throughout videos.  Moreover, a single video often features multiple characters, not all of whom appear in any single frame. For instance, in 
Figure 4, 
four individuals are present, but no frame contains all of them simultaneously. Identifying the unique features of each character within a video forms the foundation for deeper comprehension.

\subsubsection{Generating character features}\label{ap:mesh-char}
In the process of feature extraction, spatial grounding information sourced from the TVQA+ dataset is utilized in conjunction with Deepseek-v3 \cite{deepseekv3} and GPT-4o \cite{gpt4-2}. Initially, individual name and their corresponding frame-level bounding boxes are derived from the spatial grounding information employing Deepseek-v3. Subsequently, by leveraging the extracted names and bounding box specifics, eight distinctive person features are generated by GPT-4o, containing the following attributes:
\begin{itemize}
  \item \textbf{Garment Type}: e.g., t-shirt, jacket, coat, button-up shirt, blouse, etc.
  \item \textbf{Garment Color}: e.g., red, blue, yellow, green, etc.
  \item \textbf{Gender}: Male or female.
  \item \textbf{Glasses}: Whether the person wears glasses or not.
  \item \textbf{Garment Sleeve}: e.g., long sleeves, short sleeves, no sleeves, and half sleeves.
  \item \textbf{Garment Collar}: Whether the upper garment has a collar or not.
  \item \textbf{Garment Pocket}: Whether the upper garment has pockets or not.
  \item \textbf{Garment Shade}: Whether the overall color of the upper garment is dark or light.
\end{itemize}
Following the feature extraction phase, a human verification stage is conducted on the dataset. These eight discriminative features are employed to denote an individual in the video, as exemplified by Sheldon's feature set in video clip \textbf{A} (Figure \ref{fig:SettingExample}):  
\{"Garment Type": "t-shirt", "Garment Color": "blue", "Gender": "male", "Glasses": "no glasses", "Garment Sleeve": "short sleeves", "Garment Collar": "no collar", "Garment Pocket": "no pocket", "Garment Shade": "unknown"\}. Each character feature goes through a human verification process as illustrated in Figure \ref{fig:personAnnotationPipeline}.

\subsubsection{Selecting Character}\label{ap:mesh-hallu-char-anno}
To construct the targets with different granularity in Binary Task and MC Task, undesired features are masked by Granularity Mask $M$, the desired features are formulated anonymous descriptors (e.g., ''A male wearing a blue t-shirt'' or ''A person in short-sleeved blue upper garment without glasses'') to indicate Sheldon in the video clip, replacing direct name references ''Sheldon''.

To construct high-quality traps, for each category of person-related features, human annotators manually cluster semantically similar attributes into taxonomically organized groups. For instance, within the ``Garment Type" classification, items such as button-up shirts and long-sleeved shirts are aggregated into a unified semantic category. During the generation of traps, for every character in the video clips, his/her features are replaced with alternative attributes from distinct categorical groups of the same feature type to form the trap character in the question. This mechanism is denoted by operation "$\not \approx$" in Algorithm \ref{alg:personAnnotationPipeline}. It generates traps that can be clearly distinguished from the original character, instead of simply "rephrasing" the features. 
The formal construction process is listed in Algorithm \ref{alg:personAnnotationPipeline}.

The detailed information of each subdataset with different granularity is listed below:

\begin{itemize}

\item \textbf{Coarse-Grained Subdataset:}  
This subdataset is constructed using the three most visually distinguishable features: \textbf{Garment Type, Garment Color,} and \textbf{Gender}, with undesired features like \textbf{Glasses} and \textbf{Garment Sleeve} being masked. These features serve as the primary identifiers for each individual within the subdataset. \textit{Example:} \textit{ \{"Garment Color": "green", "Garment Type": "t-shirt", "Gender": "female"\} } is transformed to "\textit{a woman wearing green t-shirt}".

\item \textbf{Medium-Grained Subdataset:}  
Building upon the Coarse-Grained Subdataset, this version unmasks two additional, less prominent features: \textbf{Glasses} and \textbf{Garment Sleeve.} \textit{Example:} \textit{ \{"Garment Color": "white", "Garment Type": "coat", "Glasses": "no glasses", "Garment Sleeve": "long sleeves", "Gender": "male"\} } is converted to "\textit{a man wearing long-sleeved white coat and not wearing glasses}".

\item \textbf{Mixed-Grained Subdataset:}  
In comparison to the Medium-Grained Subdataset, three additional, less noticeable features are unmasked: \textbf{Garment Collar, Garment Pocket,} and \textbf{Garment Shade,} resulting in a total of eight features. To assess the model's robustness against missing information, three out of the eight available features are randomly selected to be masked. \textit{Example:} \textit{ \{"Garment Color": "red", "Garment Shade": "unsure", "Garment Sleeve": "long sleeves, "Garment Pocket" : "pocket", "Garment Collar": "collar"\} } is formulated into "\textit{A person wearing long-sleeved collared red upper garment with pockets}".

\item \textbf{Fine-Grained Subdataset:}  
Unlike the Mixed-Grained Subdataset, where feature selection is random, all eight features are unmasked for this subdataset. To evaluate the model's sensitivity to fine-grained details, each trap is designed such that the \textbf{Garment Type} and \textbf{Garment Color} are always identical to those of a target in the same video. Consequently, the traps and targets differ only in the finer-grained features.\textit{Example:} \textit{\{"Garment Color": "white and brown", "Garment Shade": "light", "Glasses": "no glasses", "Garment Type": "jacket", "Garment Sleeve": "long sleeves", "Garment Pocket": "no pocket", "Garment Collar": "collar", "Gender": "male"\} } is equal to "\textit{A man wearing light long-sleeved collared red jacket with pockets and not wearing glasses}". 

\end{itemize}

\subsection{Stage hallucination}\label{ap:mesh-hallu-stage}

\subsubsection{Uniqueness of Stage}\label{ap:mesh-hallu-stage-uniq}
As discussed in \ref{ap:human-under}, videos rely on multi-frame sequences to convey chronological information, with actions across frames serving as key content. A video may include multiple characters performing similar actions, requiring humans to first identify the subject to understand the action. For example, in 
Figure 2, 
the actions \textit{“a man in a blue jacket walks into the room and closes the door”} and \textit{“a woman in a beige jacket leaves the hallway”} illustrate distinct interactions involving characters, actions, and objects—critical for video understanding. Dialogue is treated as a unique action set, as recognizing the speaker is essential for understanding video content, especially when combined with facial expressions and audio. For instance, in 
Figure 2, 
the video features four main characters, each delivering a passage in specific clips. Identifying the speaker requires linking their dialogue to their features during these moments.

\subsubsection{Filtering dialogue details}\label{ap:stage-filter}
To reduce time and cost, we utilize existing question-answer pairs from the TVQA+ dataset, denoted as $QA = \{q, [c_1, c_2, c_3, c_4, c_5], a\}$, to construct the event hallucination benchmark. However, two challenges arise in reusing this data. First, the characters in $QA$ are typically referred to by name (e.g., “Sheldon”, “Leonard”, “Penny”), which is unsuitable for a general-purpose video understanding benchmark. Using the person feature annotations introduced in the previous section, we replace names with coarse-grained descriptions (e.g., “a man wearing a blue T-shirt”). Second, some $QA$ pairs rely on subtitle-based information. To ensure the remaining $QA$ require only visual understanding, we design Algorithm 4, which filters out questions containing n-grams that appear solely in the subtitle sequence $S_{Sub} = [Sub_1, Sub_2,…, Sub_n]$ but not in the object set $S_o$. The algorithm outputs $Stm$, a description of a single event in the video. If the corresponding flag $B = \text{False}$, the $Stm$ is retained and used as the target event for generating event hallucination questions. Detailed process is listed in Algorithm \ref{alg:subtitle_filter}.
\begin{algorithm}
	\caption{Subtitle Filter} 
    \label{alg:subtitle_filter}
	\begin{algorithmic}[1]
        \Statex \textbf{Input:} TVQA+ Question answer pair $QA = (q, [c_1, c_2, c_3, c_4, c_5], a)$, Subtitles list $S_{Sub} = [Sub_1, Sub_2,..., Sub_n]$, Object set $S_o = \{o_1, o_2, ..., o_m\}$.
        \State $Stm \leftarrow$ LLM ("... Convert the question answer pair $\{q, c_a\}$ into several declarative sentences ..." )
        \State $Stm_{norm} \leftarrow$ LLM("... Extract all the norms that appear in $Stm$ ...")
        \State $Sub_{norm} \leftarrow $ LLM ("... Extract all the norms that appear in $S_{Sub}$ ...")
        \State $B \leftarrow$ False
        \For {$norm_i$ in $Stm_{norm} = \{norm_1, norm_2, ..., norm_\alpha\}$}
            \If {$norm_i \in Sub_{norm}$ and $norm_i \not \in S_o$ }
                \State $B \leftarrow$ True
                \State break
            \EndIf
        \EndFor
        \Statex \textbf{Output:} $Stm$ that describe part of the video, Whether the statement involved content from subtitle $B$.
	\end{algorithmic} 
\end{algorithm}

\subsubsection{Annotated dialogues}\label{ap:mesh-dia}
Conversation extraction is performed by analyzing TVQA+ subtitle talking information to identify conversational exchanges. Initially, the speaker's identity and temporal metadata (start/end timestamps) are extracted from the raw subtitle material. Subsequently, speaker names are systematically substituted with their corresponding feature descriptors (e.g., ``woman in a red blouse", ``man with glasses") using a predefined feature database derived from the person feature dataset. This process ensures anonymization while preserving contextual relevance through attribute-based identification. As illustrated in Figure 5a, this method produces anonymous output, such as ``A woman in a red dress without glasses is talking”, showing how the conversation is extracted from the video clip.

\subsubsection{Annotated actions}\label{ap:mesh-act}
The retain $Stm$ from Algorithm \ref{alg:subtitle_filter} are then parsed to extract actions in the syntactic template “subject-action”, where subject represents the agent and "action" represent the action performed by the agent (e.g., drinking white wine, sitting at the table eating dinner, or running into the door). Names within these extracted actions are substituted with their corresponding feature descriptors (e.g., garment type, gender) derived from the pre-extracted person feature database. The target is assembled by the feature descriptors and corresponding action. For the traps, to generate different categories of traps, a pool of subject and action is collected and different selecting algorithm is applied in Algorithm \ref{alg:action_question}, which is discussed in detailed in the next section \ref{ap:mesh-hallu-stage-anno}.

\subsubsection{Selecting Stage}\label{ap:mesh-hallu-stage-anno}

\begin{algorithm}
	\caption{Action Hallucination Questions Construction} 
    \label{alg:action_question}
	\begin{algorithmic}[1]
        \Statex \textbf{Input:} Video-Statement dictionary $\mathbb{D} = \{v_1: [Stm^{v_1}_1, Stm^{v_1}_2, ...], v_2: [...], ...\}$, Random sample algorithm $Random_c$, Number of Iteration $n_t$.
        \State $\mathbb{S}_{A} \leftarrow \{\}$.
        \State $\mathbb{S}_{P} \leftarrow \{\}$.
        \For {$Stm_j^{v_i}$ in $\mathbb{D}$}
            \State $(P, A) \leftarrow$ LLM("... divide the statement $Stm_j^{v_i}$ into two parts: subjective and action ...").
            \State $\mathbb{S}_A \leftarrow \mathbb{S}_A \cup \{A\}$.
            \State $\mathbb{S}_P \leftarrow \mathbb{S}_P \cup \{P\}$.
        \EndFor
        \State $\mathbb{D}_{trap} \leftarrow \{\} $.
        \For {$v_i$ in $\mathbb{D}$} 
            \State $\mathbb{D}_{trap} \leftarrow \mathbb{D}_{trap} \cup \{v_i:\{\}\} $
             \For {$iteration=1,2,\ldots, n_t$} \Comment{Loop 1}
                \State $A_{trap} \leftarrow Random_c (\mathbb{S}_A)$.
                \State $P_{trap} \leftarrow Random_c(\mathbb{S}_P)$
                \State $Stm_{trap} \leftarrow P_{trap} + A_{trap}$
                \For {$Stm_j^{v_i}$ in $\mathbb{D}[v_i]$}
                    \If {$Stm_{trap} \approx Stm_j^{v_i}$}
                        \State break Loop 1.
                    \EndIf
                \EndFor
                \State $\mathbb{D}_{trap}[v_i] \leftarrow \mathbb{D}_{trap}[v_i] \cup \{Stm_{trap}\}$
             \EndFor
        \EndFor
        \Statex \textbf{Output:} Trap dictionary $\mathbb{D}_{trap}$.
	\end{algorithmic} 
\end{algorithm}

For the selection stage in the dialogue dataset, two distinct motion hallucination categories are formally defined by replacing the original speaker with different types of individuals. The details of these categories are explained below, with the video clip from Figure 5a serving as an example:

\begin{itemize}
  \item \textbf{Character out of Video (CO):}  
    Generated by substituting the speaker with a non-existent person in the video.
  \textit{Example:} From \textit{``A woman in a red dress without glasses is talking”} to \textit{``A man in a red jacket without glasses is talking”}, where \textit{``A man in a red jacket without glasses”} has never appeared in the entire video.

  \item \textbf{Character in Video (CI):}  
  Generated by replacing the speaker with a person present in the video clip who has never spoken throughout the video.
  \textit{Example:} From \textit{``A woman in a red dress without glasses is talking} to \textit{``A man in a black suit with glasses is talking"}, where \textit{``A man in a black suit with glasses”} exists in the video but has never spoken.
\end{itemize}

For the selecting stage in the non-diaglogue dataset, by synthesizing agents and their corresponding actions from heterogeneous sources, five distinct motion hallucination categories are formally defined. Details of the five types are explained below, with the video clip from Figure 5b serving as an example:

\begin{itemize}
  \item \textbf{Action out of Video (AO):}  
  Constructed by pairing individuals present in the video with actions derived from external video contexts.  
  \textit{Example:} From \textit{``A man in a blue jacket without glasses is walking into the room”} to \textit{``A man in a blue jacket without glasses is cooking some food”}, where \textit{``cooking some food”} is an action from other video clips.

  \item \textbf{Character out of Video (CO):}  
  Generated by integrating extraneous agents (not present in the video) with actions native to the original scene.  
  \textit{Example:} From \textit{``A man in a blue jacket without glasses is walking into the room”} to \textit{``A woman in a blue jacket without glasses is walking into the room”}, where \textit{``A woman in a blue jacket without glasses"} has never appeared in the video.

  \item \textbf{Similar Action (SA):}  
  Created by semantically perturbing in-scene actions to produce plausible yet absent variants.  
  \textit{Example:} From \textit{``A man in a blue jacket without glasses is walking into the room”} to \textit{``A man in a blue jacket without glasses is jumping outside the room}, where \textit{``jumping outside the room”} is a semantically related yet non-occurring action.

  \item \textbf{Mixed in Video (MI):}  
  Formulated by interchanging roles between existing agents and their associated actions.  
  \textit{Example:} Swapping descriptors such as \textit{``A man in a blue jacket without glasses is walking into the room''} and \textit{``A man in a gray t-shirt is sitting''} to generate \textit{``A man in a blue jacket without glasses is sitting''} and \textit{``A man in a blue jacket without glasses is sitting''}, where the two actions before the change are actions that exist in the video clip.
\end{itemize}

\subsubsection{Quality Guarantee}
Finally, to guarantee traps can be clearly distinguished from the target instead of simply "rephrasing" the description, \textit{text-embedding-3-small}\footnote{https://platform.openai.com/docs/guides/embeddings} is utilized to vectorized the targets and the trap. If the similarity between targets and the trap, which is calculated by Cosine Distance, is greater than $0.65$, then the trap will be abandoned. This mechanism is represented by operation "$\approx$" in Algorithm \ref{alg:action_question}. The formal construction process is listed in Algorithm \ref{alg:action_question}.

\subsection{Spatial-Temporal Hallucination}\label{ap:tem-spa}

\acmo{After aligning characters with their movements, humans naturally interpret multiple actions by noticing and using their temporal order. Extending our hallucination benchmark to a higher level, we therefore evaluate the model’s understanding of temporal order. Using the existing question–answer pairs and time-slot annotations from the TVQA+ dataset, we construct a small dataset specifically for evaluating temporal hallucination. As with stage annotation pipeline in \ref{ap:stage-filter}, QA pairs that rely only on subtitle-based information are filtered out, and the remaining pairs are transformed into \textit{Stm}. Consequently, for each video, we obtain a list of actions along with their corresponding start and end timestamp tuples.}

\[[(Stm_1, ts_1^s, ts_1^e),\cdots, (Stm_i, ts_i^s, ts_i^e),\cdots, (Stm_n, ts_n^s, ts_n^e)]\]

\acmo{We select suitable videos and their corresponding action lists using the following criterion:}

\[\exists i,j,k \in \{1,\cdots,n\} \text{ and } i\not = j \not = k \quad \text{such that} \quad ts_i^e < ts_j^i \text{ and } ts_j^e < ts_k^s \]

\begin{figure}[h]
    \centering
    \includegraphics[width=0.9\columnwidth]{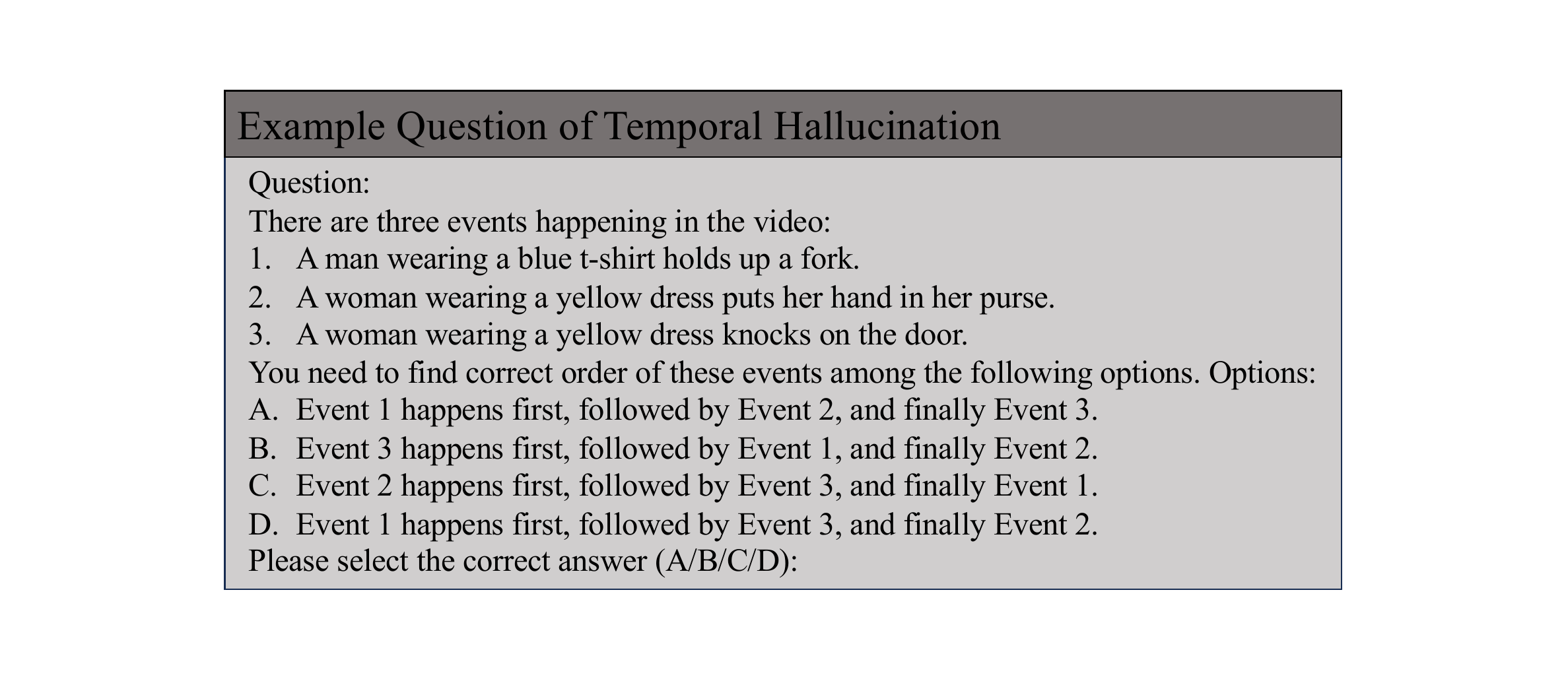} 
    \caption{An example of spatial-temporal hallucination question. The 1, 2, and 3 in the question can be permuted or substituted with different event description.}
    \label{fig:TemporalHallucinationExample}
    \vspace{-0.2cm}
\end{figure}

\acmo{This ensures that there are at least three distinct actions occurring in a strict temporal order. The selected actions are then randomly permuted to create a spatial-temporal hallucination question. An example is shown in Figure \ref{fig:TemporalHallucinationExample}.
In the experiments, we uniformly sample 64 frames from each video and provide them to the LVM along with the constructed question. The results are reported in Table \ref{TemporalHallucinationExp}.}

\begin{table}[H]
  \caption{Temporal hallucination experiment results on a small set of models.}
  \label{TemporalHallucinationExp}
  \centering
  \begin{tabular}{lllllll}
\toprule
\textbf{Model} & LLaVA-Video-7B & LLaVA-OV-7B & Qwen2-VL-7B & Qwen2-VL-2B & InternVL2.5-8B & InternVL2.5-2B \\ 
    \midrule
    \textbf{Performance} &0.515 & 0.431 & 0.389 & 0.198 & 0.393 & 0.290 \\
    \bottomrule
  \end{tabular}
\end{table}


\acmo{The results are broadly consistent with before. This suggests that higher-level interpretations rely on the correctness of lower-level hallucinations, supporting our bottom-up framework design. Notably, the \textit{LLaVA-Video} model, finetuned on video data, outperforms both the \textit{LLaVA-OV} model and other models. Furthermore, the complexity of temporal relations, long input sequences, and rich action contexts make temporal hallucination particularly challenging for LVMs.}

\subsection{Generalization of MESH Pipeline}\label{ap:transfer-method}
\begin{figure}[h!]
    \centering
    \begin{subfigure}[b]{0.32\textwidth}
        \includegraphics[width=\textwidth]{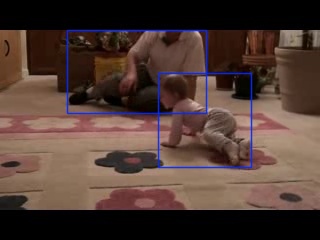}
    \end{subfigure}
    \hfill
    \begin{subfigure}[b]{0.32\textwidth}
        \includegraphics[width=\textwidth]{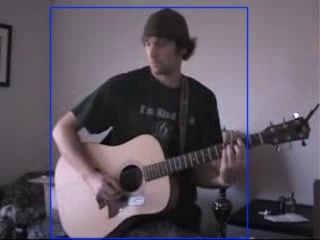}
    \end{subfigure}
    \hfill
    \begin{subfigure}[b]{0.32\textwidth}
        \includegraphics[width=\textwidth]{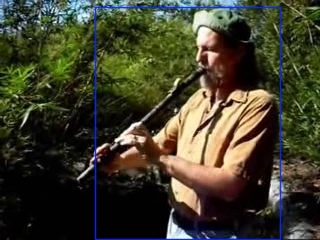}
    \end{subfigure}
    \caption{Using YOLO to generate bounding boxes for the characters in UCF101 dataset}
    \label{fig:bounding_box_yolo}
\end{figure}

\acm{
The setting–character–stage framework is widely used to analyze complex videos like films and TV series, which offer richer information flow than everyday footage. We chose TVQA+ for its dense annotations and rich visual cues—e.g., camera movement and character actions (see 
Figure 2
). \citet{visual} shows this approach generalizes to other video types. We demonstrate how our categorization and annotation pipeline can be applied across datasets.
}
\begin{itemize}
    \item \textbf{Setting: environments and objects} Location labels (e.g., “living room,” “kitchen”) are accurately generated using GPT-4o. While object annotations in TVQA+ are sparse, they can be supplemented using human labelers or vision models. Like humans, these models may miss some objects but can reliably infer environment contexts.
    
    \item \textbf{Characters: features and identity} Most video datasets lack fine-grained human annotations, which we address with a \textbf{bounding-box-first} pipeline. Using YOLO \cite{yolo} (see Figure \ref{fig:bounding_box_yolo}) to generate bounding boxes, we annotate physical features verified by human labelers. Such verification process is also seen in other datasets \cite{VELOCITI, TVBench}. This supports precise character recognition and enables generation of hallucination and stage questions based on diverse character–action pairs.
    
    \item \textbf{Stage: leveraging action annotations}  Many datasets already include useful action labels \cite{action_recog_1, action_recog_2}, which we pair with character features to generate stage-related questions. Dialogue attribution is straightforward via subtitles or audio and typically requires no manual annotation.
\end{itemize}

To further verify the generalization ability of MESH category and pipeline, we apply them to the UCF101\cite{ucf101} dataset. As mentioned above, for the setting hallucination, we prompt GPT-4o to generate the locations and objects appear in the video. Then, human annotator will examine the correctness of generated labels. Finally, the location and object labels are feed into the same pipeline described in \ref{ap:mesh-hallu-set} to generate targets and traps. The experiments result on UCF101 setting hallucination is presented in Table \ref{UCF101SettingHalluExp}.

\begin{table}[H]
  \caption{Setting hallucination on UCF101}
  \label{UCF101SettingHalluExp}
  \centering
  \begin{tabular}{lllllll}
\toprule
\textbf{Model} & LLaVA-Video-7B & LLaVA-OV-7B & Qwen2-VL-7B & Qwen2-VL-2B & InternVL2.5-8B & InternVL2.5-2B \\ 
    \midrule
    \textbf{Binary Task} &0.945  & 0.930 & 0.956 & 0.937 & 0.894 & 0.868 \\
    \textbf{Multi-Choice Task} &0.978   & 0.919 & 0.905 & 0.992 & 0.926 & 0.905 \\
    \bottomrule
  \end{tabular}
\end{table}

For the character hallucination, after generating bounding box for the people that appear in the video, the bounding box and video are feed into MESH character hallucination pipeline \ref{ap:mesh-hallu-char} to generate questions. The result is presented in Table \ref{UCF101CharacterHalluExp}.

\begin{table}[H]
\caption{Character hallucination on UCF101}
\label{UCF101CharacterHalluExp}
\centering
    \begin{tabular}{llrrrrrr}
    \toprule
    \textbf{Model} & \textbf{Grain} & \multicolumn{3}{c}{\textbf{Binary Task}} & \multicolumn{3}{c}{\textbf{Multi-Choice Task}} \\
    \cmidrule(lr){3-5} \cmidrule(lr){6-8}
              &        & Short & Medium & Long  & Short & Medium & Long \\
    \midrule
    
    \multirow{1}{*}{LLaVA-Video-7B} & Coarse &  0.964 & 0.956 & 0.964 &   0.964 & 0.964 & 0.964 \\
    &Medium & 0.894 & 0.894 & 0.912 &  0.824 & 0.842 & 0.859  \\
    & Mixed &  0.842 & 0.780 & 0.807 &   0.771 & 0.807 & 0.807 \\
    &Fine &  0.771 & 0.789 & 0.780 &   0.754 & 0.701 & 0.719 \\
    \midrule
    
    \multirow{1}{*}{LLaVA-OV-7B} & Coarse &  0.964 & 0.938 & 0.912 &  0.964 & 0.912 & 0.929 \\
    &Medium & 0.859 & 0.850 & 0.833 & 0.771 & 0.789 & 0.824  \\
    & Mixed &  0.789 & 0.780 & 0.780 &   0.736 & 0.807 & 0.789 \\
    &Fine &  0.719 & 0.719 & 0.710 &   0.596 & 0.596 & 0.614 \\
    
    \bottomrule
    \end{tabular}
\end{table}



\acmo{Preliminary results reveal both similarities and differences between the TVQA+ and UCF101 datasets. First, UCF101 is simpler, with short ($\sim$10s) videos and fixed locations. Second, in both benchmarks, character hallucination follows the same trend: finer granularity leads to lower accuracy. Third, TVQA+ shows a negative correlation between the number of input frames and accuracy, whereas UCF101 does not, likely due to its simplicity—more frames do not add useful information. Finally, unlike TVQA+, each UCF101 video contains only one character and one action, making it insufficient for stage hallucination evaluation.  }

\acmo{
To address this, we construct the UCF101-COMBINE dataset by concatenating 4–5 UCF101 videos into one, better matching the complexity of TVQA+. We then apply the setting and character hallucination pipelines, with results reported in Table~\ref{UCF101-COMBINESettingHalluExp} and Table~\ref{UCF101-COMBINECharacterHalluExp}.}

\begin{table}[H]
  \caption{Setting hallucination on UCF101-COMBINE}
  \label{UCF101-COMBINESettingHalluExp}
  \centering
  \begin{tabular}{lllllll}
\toprule
\textbf{Model} & LLaVA-Video-7B & LLaVA-OV-7B & Qwen2-VL-7B & Qwen2-VL-2B & InternVL2.5-8B & InternVL2.5-2B \\ 
    \midrule
    \textbf{Binary Task} &0.926  & 0.899 & 0.864 & 0.732 & 0.768 & 0.705 \\
    \textbf{Multi-Choice Task} &0.969   & 0.911 & 0.854 & 0.872 & 0.896  & 0.778 \\
    \bottomrule
  \end{tabular}
\end{table}

\begin{table}[H]
\caption{Character hallucination on UCF101-COMBINE}
\label{UCF101-COMBINECharacterHalluExp}
\centering
    \begin{tabular}{llrrrrrr}
    \toprule
    \textbf{Model} & \textbf{Grain} & \multicolumn{3}{c}{\textbf{Binary Task}} & \multicolumn{3}{c}{\textbf{Multi-Choice Task}} \\
    \cmidrule(lr){3-5} \cmidrule(lr){6-8}
              &        & Short & Medium & Long  & Short & Medium & Long \\
    \midrule
    
    \multirow{1}{*}{LLaVA-Video-7B} & Coarse &  0.928 & 0.892 & 0.852 &   0.941 & 0.868 & 0.838 \\
    &Medium & 0.884 & 0.827 & 0.776 &  0.827 & 0.775 & 0.737  \\
    & Mixed &  0.850 & 0.817 & 0.769 &   0.788 & 0.713& 0.620 \\
    &Fine &  0.855 & 0.807 & 0.724 &   0.721 & 0.635 & 0.581 \\
    \midrule
    
    \multirow{1}{*}{LLaVA-OV-7B} & Coarse &  0.912 & 0.860 & 0.816 &  0.920 & 0.866 & 0.769 \\
    &Medium & 0.845 & 0.785 & 0.724 & 0.730 & 0.642 & 0.558  \\
    & Mixed &  0.838 & 0.743 & 0.675 &   0.709 & 0.601 & 0.517 \\
    &Fine &  0.822 & 0.728 & 0.633 &   0.532 & 0.480 & 0.407 \\
    \midrule
    
    \multirow{1}{*}{Qwen2VL-7B} & Coarse &  0.880 & 0.816 & 0.789 &  0.875 & 0.782 & 0.689 \\
    &Medium & 0.794 & 0.757 & 0.667 & 0.655 & 0.575 & 0.504  \\
    & Mixed &  0.797 & 0.721 & 0.695 &   0.633 & 0.562 & 0.461 \\
    &Fine &  0.812 & 0.711 & 0.614 &   0.495 & 0.435 & 0.381 \\
    \midrule
    
    \multirow{1}{*}{Qwen2VL-2B} & Coarse &  0.846 & 0.775 & 0.700 &  0.859 & 0.773 & 0.674 \\
    &Medium & 0.699 & 0.587 & 0.552 & 0.642 & 0.592 & 0.510  \\
    & Mixed &  0.732 & 0.643 & 0.571 &   0.536 & 0.448 & 0.385 \\
    &Fine &  0.735 & 0.609 & 0.549 &   0.525& 0.467 & 0.411 \\
    \midrule
    
    \multirow{1}{*}{InternVL2.5-8B} & Coarse &  0.858 & 0.789 & 0.743 &  0.884 & 0.855 & 0.802 \\
    &Medium & 0.764& 0.741 & 0.681  & 0.762  & 0.728  &0.636 \\
    & Mixed &   0.717  &0.695  &0.621  & 0.679&0.604 &0.536 \\
    &Fine &  0.721& 0.667 &0.604 & 0.621& 0.565& 0.472  \\
    \midrule
    
    \multirow{1}{*}{InternVL2.5-2B} & Coarse &  0.917 &0.869 &0.806 &  0.935& 0.887& 0.825 \\
    &Medium & 0.859& 0.774&0.681  & 0.795&0.728& 0.659 \\
    & Mixed & 0.764 & 0.689& 0.606& 0.668& 0.590& 0.461 \\
    &Fine &  0.766& 0.684& 0.551 & 0.579 &0.525 & 0.467  \\
    
    \bottomrule
    \end{tabular}
\end{table}


\acmo{Results on the UCF101-COMBINE dataset are more consistent with those on TVQA+. Since each combined video includes multiple characters and actions, we can apply the stage hallucination pipeline (Section~\ref{ap:mesh-hallu-stage}). The results are given in Table~\ref{UCF101-COMBINEStageHalluExp}.}

\begin{table}[H]
  \caption{Stage hallucination on UCF101-COMBINE}
  \label{UCF101-COMBINEStageHalluExp}
  \centering
  \begin{tabular}{lllllll}
\toprule
\textbf{Category} & LLaVA-Video-7B & LLaVA-OV-7B & Qwen2-VL-7B & Qwen2-VL-2B & InternVL2.5-8B & InternVL2.5-2B \\ 
    \midrule
    \textbf{AOV} &0.955  & 0.917 & 0.860 & 0.799 & 0.890 & 0.856 \\
    \textbf{COV} &0.766   & 0.671 & 0.637 & 0.471  & 0.636  &  0.622 \\
    \textbf{MIV} &0.606   & 0.474 & 0.476 & 0.295 & 0.543  & 0.451 \\
    \bottomrule
  \end{tabular}
\end{table}

\acmo{Overall, applying the MESH categories and pipeline to the UCF101-COMBINE dataset replicates the key phenomena observed on TVQA+, demonstrating both the generalization of the MESH pipeline and the universality of our conclusions. At the same time, minor differences emerge. For instance, while the original UCF101 shows no correlation between video length and accuracy, in UCF101-COMBINE the accuracy on the COV stage hallucination category is consistently lower than on AOV. A likely reason is that actions are more salient, leading models to rely on them while ignoring character features. These subtle differences highlight that applying the MESH pipeline to new datasets can yield novel insights into LVMs' understanding of different videos types.}

\section{Experiments}\label{ap:ep}
\subsection{General details}\label{ap:ep-gd}
\subsubsection{Difficulty level}\label{ap:ep-diff}
For \textbf{setting}, all questions are classified as \textbf{basic}.  
For \textbf{characters}, questions on \textbf{coarse} and \textbf{medium}-level features are \textbf{basic}, while \textbf{mixed} and \textbf{fine}-level features are \textbf{advanced}.  
For \textbf{stage}, all \textbf{binary} questions are \textbf{basic}.  
For \textbf{action MC questions}, \textbf{AO} and \textbf{CO} are \textbf{basic}, while \textbf{SA} and \textbf{MI} are \textbf{advanced}.  
For \textbf{dialogue MC questions}, \textbf{CO} is \textbf{basic}, and \textbf{CI} is \textbf{advanced}.
\subsubsection{Dataset Statistics}
\begin{figure}[h]
    \centering
    \includegraphics[width=0.9\columnwidth]{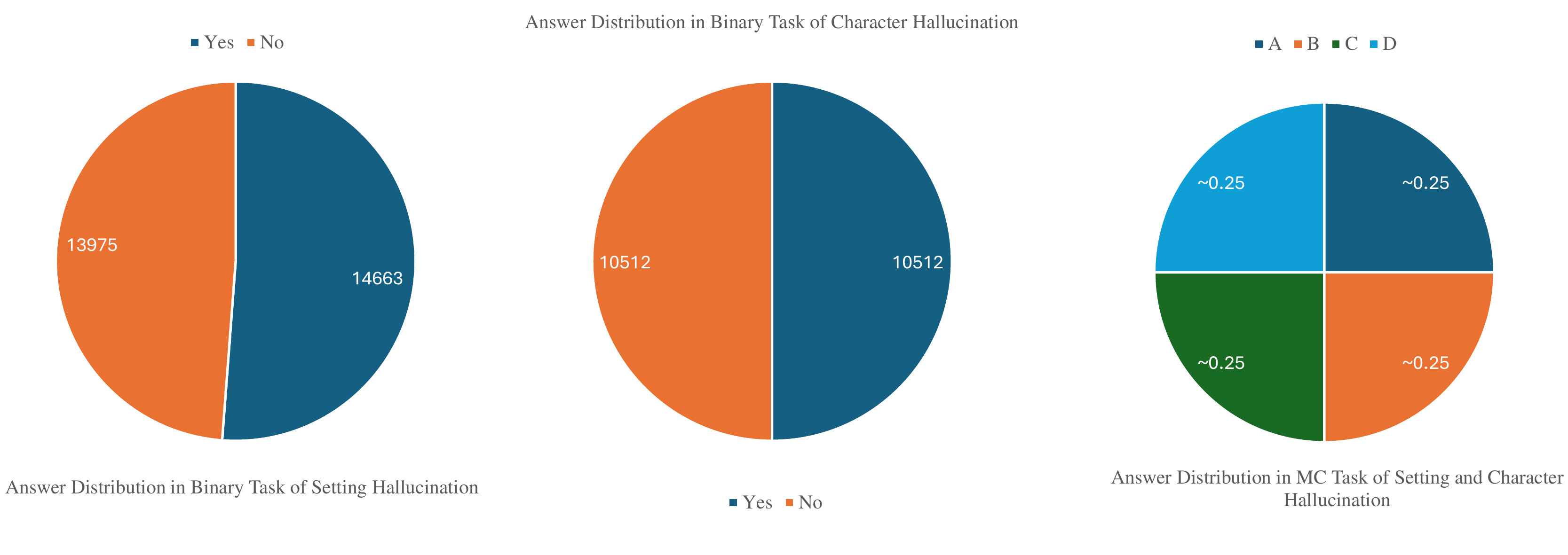} 
    \caption{Answer Distribution for binary task and multi-choice task in Setting and Character Hallucination.}
    \label{fig:AnswerDistribution1}
    \vspace{-0.2cm}
\end{figure}

The datasets employed in this study—the \textbf{Setting Hallucination Dataset} and the \textbf{Character Hallucination Dataset}—are structured into binary classification and multiple-choice task formats. Across these datasets and their associated subdatasets (detailed in Section \ref{ap:mesh-char}), the binary classification tasks maintain a balanced class distribution. In contrast, the multiple-choice tasks comprise questions with four answer options, each displaying an approximately uniform empirical distribution across all evaluated subdatasets, as illustrated in Figure \ref{fig:AnswerDistribution1}.  

\begin{figure}[h]
    \centering
    \includegraphics[width=0.9\columnwidth]{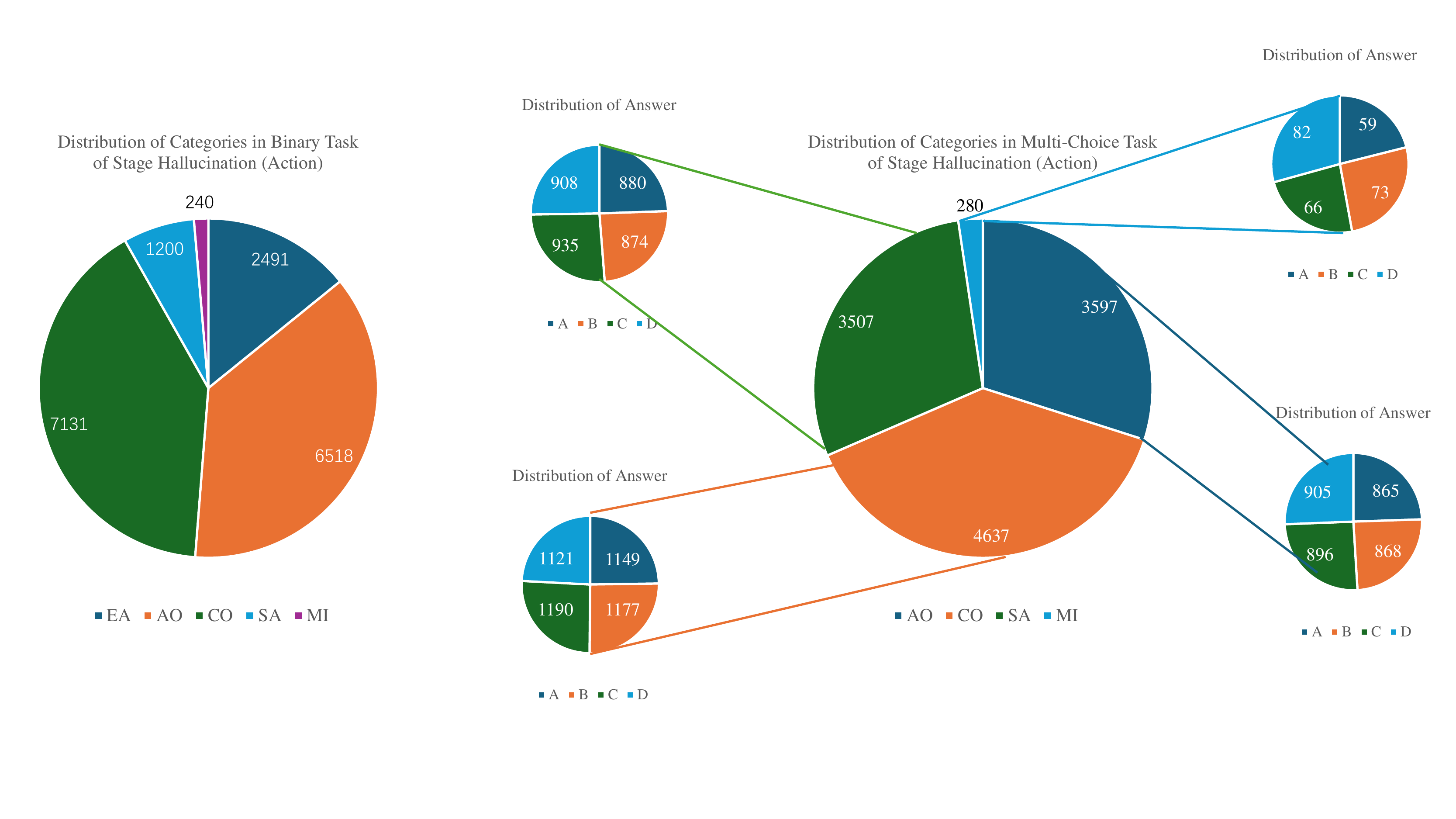} 
    \caption{Category and answer distribution for binary task and multi-choice task in Stage Hallucination (Action).}
    \label{fig:AnswerDistribution2}
    \vspace{-0.2cm}
\end{figure}

\begin{figure}[h]
    \centering
    \includegraphics[width=0.9\columnwidth]{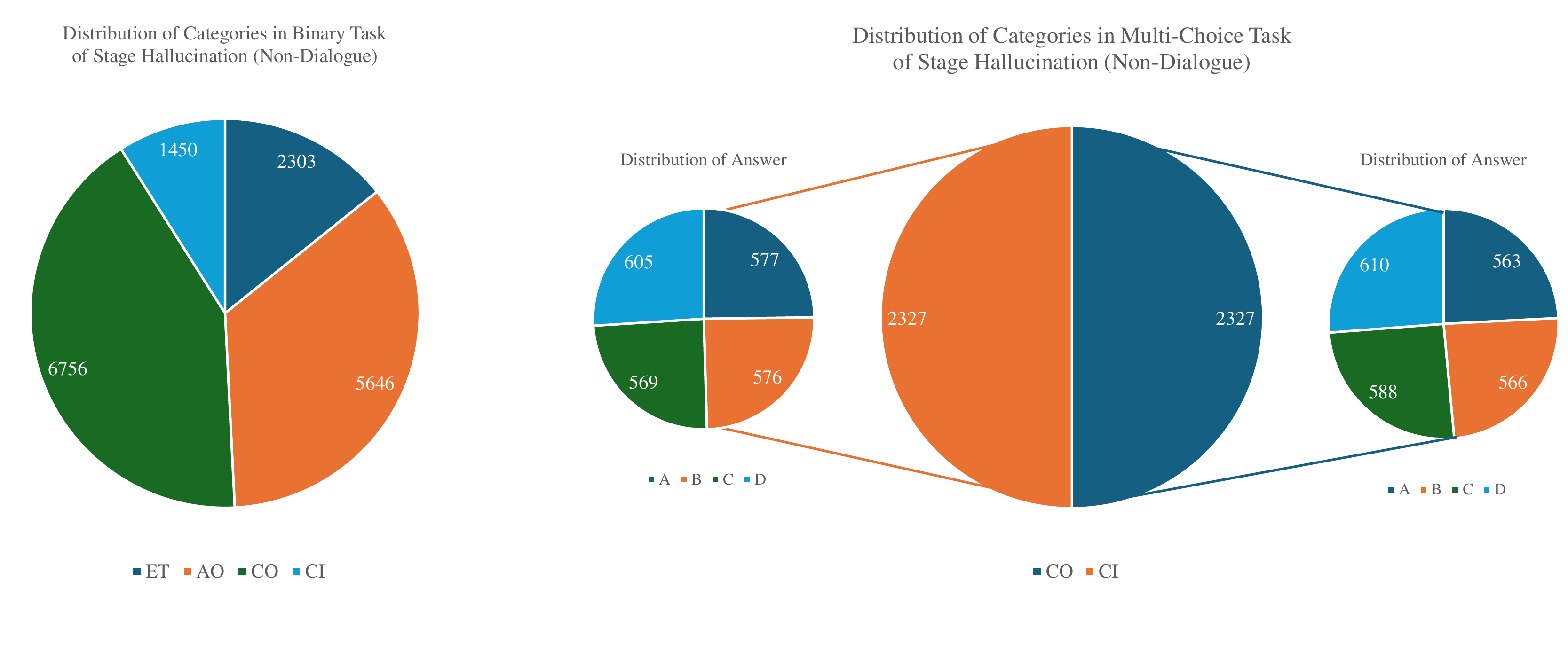} 
    \caption{Category and answer distribution for binary task and multi-choice task in Stage Hallucination (Dialogue).}
    \label{fig:AnswerDistribution3}
    \vspace{-0.2cm}
\end{figure}

The \textbf{Stage Hallucination Dataset} is similarly organized into binary classification and multiple-choice tasks but is further divided into two types and four distinct categories based on its generation methods (see Section \ref{ap:mesh-hallu-stage-anno} for details). Unlike the Setting and Character Hallucination datasets, the Stage Hallucination Dataset exhibits distinct distribution patterns across its categories and answer options. These differences are visually represented in Figures \ref{fig:AnswerDistribution2} (action) and \ref{fig:AnswerDistribution3} (dialogue), highlighting its unique structural and characteristics. 

\paragraph{Stage Hallucination Dataset (Action)}
\begin{itemize}
    \item \textbf{Binary Classification Task:} (Figure \ref{fig:AnswerDistribution2} left) The dataset includes 17,577 questions (2,488 labeled "yes" and 15,089 labeled "no"). The "no" responses are categorized into four groups: "AO" (6,518), "CO" (7,131), "SA" (1,200), and "MI" (240).
    
    \item \textbf{Multi-Choice Task:} (Figure \ref{fig:AnswerDistribution2} right) This task consists of a total of 12,021 questions. Those questions are divided into four partitions: "AO"(3,597), "CO" (4,637), "SA" (3,507), "MI"(280).
\end{itemize}

\paragraph{Stage Hallucination Dataset (Dialogue)}
\begin{itemize}
    \item \textbf{Binary Classification Task:}(Figure \ref{fig:AnswerDistribution3} left) This subset contains 16,155 questions (2,303 labeled "yes" and 13,852 labeled "no"). The "no" answers are classified into three categories: "AO" (5,646), "CO" (6,756), and "CI" (1,450).
    
    \item \textbf{Multi-Choice Task:}(Figure \ref{fig:AnswerDistribution3} right) There are a total of 4,654 questions in this category. The category distributions are as follows: "CO" (2,327), "CI" (2,327). 
\end{itemize}

\subsubsection{Evaluated Video-Language Models}

This section introduces the video-language models (LVLMs) evaluated in this study, providing a rationale for their selection and a concise overview of their key architectural characteristics.  The models selected represent a range of design choices and technological advancements within the field, allowing for a comprehensive analysis of performance across diverse benchmarks. We highlight variations in Large Language Model (LLM) backbones and vision encoders, as well as innovative techniques employed for video understanding and efficient processing.

\begin{itemize}

    \item \textbf{Video-LLaVA} \cite{tune-5}: This model leverages the Vicuna \cite{vicuna} LLM and the languageBind \cite{bind-1} vision encoder.  LanguageBind is pre-trained on a diverse set of modalities, including video, audio, infrared, and depth data, enabling enhanced video understanding capabilities. The selection of languageBind as a vision encoder aims to evaluate the influence of diverse pre-training modalities on video-language understanding.

    \item \textbf{LLaVA-NeXT-Video} \cite{llavanextvideo}: This model exhibits a scale-dependent architectural design.  The 7B parameter version integrates Vicuna \cite{vicuna} with the CLIP \cite{clip} vision encoder, while the 32B parameter version employs Qwen1.5 \cite{qwen-1} as the LLM and SigLIP as the vision encoder. The inclusion of LLaVA-NeXT-Video allows for investigation of the interplay between LLM scale, vision encoder choice, and resulting performance.

    \item \textbf{LLaVA-Video} \cite{llavavideo}:  This LVLM utilizes the Qwen2 \cite{qwen2vl} LLM, known for its strong performance on language tasks, and the SigLIP vision encoder, which has demonstrated competitive results in various vision-related benchmarks. Comparing with previous model can provide more in-depth analysis on different LLMs and vision encoders.

    \item \textbf{LLaVA-OneVision} \cite{llavaov}: Leveraging the Qwen2 LLM and SigLIP vision encoder, this model is designed to excel in understanding single images, multiple images, and video data. Its selection allows for a performance comparison between video models and general purpose models.

    \item \textbf{Aria} \cite{aria}: This model uses SigLIP as the vision encoder and a Mixture of Experts (MoE) architecture LLM with a total of 23B parameters.  During token prediction, only 3.9B parameters are activated. This design choice allows for experimentation with parameter efficient LLMs.

    \item \textbf{LLaMA-VID} \cite{llamavid}: Employing the CLIP vision encoder and Vicuna as the LLM, LLaMA-VID incorporates a novel multi-modal adapter. This adapter reduces the number of image tokens to two per image, enabling efficient inference on long video sequences. Its inclusion tests the impact of adapter on tokens and inference.

    \item \textbf{Oryx} \cite{oryx}: Utilizing the Qwen2 LLM and SigLIP as the vision encoder, Oryx introduces a dynamic downsampling layer tailored for video processing. This layer comprises a 4x4 pooling layer and a regional cross-attention layer.  The inclusion of Oryx assesses the impact of different video processing techniques.

    \item \textbf{Qwen2VL} \cite{qwen2vl}: This model uses Qwen2 as its LLM and features a specialized vision encoder capable of encoding video frames into a variable number of tokens, adaptable to different specifications and resolutions.

    \item \textbf{VideoAgent} \cite{videoagent}: This model adopts an agent-based approach, employing an LLM as the agent and utilizing diverse vision foundation models as tools. Function calling via LangChain \footnote{https://www.langchain.com/} is used to leverage these tools for completing video understanding tasks.  This choice allows for evaluation of the agent-based approach.

    \item \textbf{VILA} \cite{vila}: Integrating the LLaMA3 \cite{llama-1} as LLM, VILA incorporates image-text interleaved training.  This approach is reported to yield significant performance gains in image, multi-image, and video understanding tasks.

    \item \textbf{InternVL2.5} \cite{InternVL2.5}:  This model uses InternLM \cite{internlm2} as its LLM and InternViT \cite{internViT} as its vision encoder, where the latter is jointly trained with the LLM using various objective functions. Its training data underwent careful filtering and augmentation.

    \item \textbf{VideoXL} \cite{videoxl}:  VideoXL leverages SigLIP as the vision encoder and Qwen2 as the LLM. It is enhanced by a visual token compression module that groups similar tokens into visual summary tokens. This compression mechanism facilitates the handling of thousands of video frames during training and inference.

    \item \textbf{VideoLLaMA2} \cite{videollama2}: This model employs SigLIP and Qwen2, and implements a novel Spatial-Temporal Convolution Connector which consists of RegStage and 3D convolution layers to help spatial and temporal information aggregation for the process of vision feature extraction in videos.

    \item \textbf{Claude 3.5-Sonnet} \cite{claude3.5sonnet}: A proprietary video-language model developed by Anthropic. Its inclusion allows comparison of opensource model with one of the popular closesource models.

    \item \textbf{GPT-4o} \cite{gpt4-2}: A proprietary video-language model developed by OpenAI. Comparing with other models enables in-depth research.

    \item \textbf{Gemini 1.5-Pro} \cite{gemini1.5}: A proprietary video-language model developed by Google. This allows assessment by one of the most popular model in the field.

\end{itemize}

\subsubsection{Technique Details}

\paragraph{Deployment of Larger LVLMs with 8×A800 GPUs}

For LVLMs exceeding 70 billion parameters—such as LLaVA-Video-72B, InternVL2.5-78B, Qwen2VL-72B, and LLaVA-OneVision-72B—we employ a high-performance computing node equipped with 8 NVIDIA A800 GPUs. To optimize inference speed, different deployment frameworks are utilized based on official recommendations and support. Specifically, LLaVA-OneVision-72B and Qwen2VL-72B, which provide official support for vLLM~\cite{vllm}, are deployed and evaluated using this framework. InternVL2.5-78B is deployed using the lmdeploy \cite{lmdeploy} framework, as recommended by its developers. For LLaVA-Video-72B, since it has not yet been integrated into other frameworks, we utilize the Hugging Face Transformers library \cite{huggingface} for deployment and evaluation.

\paragraph{Deployment of Smaller LVLMs with RTX 3090/4090 GPUs}

For LVLMs with fewer than 40 billion parameters—such as VILA1.5-8B, Aria-23B, and LLaVA-NeXT-Video-32B—the Hugging Face Transformers library is employed for model deployment and experimentation. To minimize computational costs, these models are deployed on consumer-grade GPUs, specifically the NVIDIA RTX 3090 and RTX 4090.

\subsection{Setting hallucination}\label{ap:ep-set}
We present the detailed experiment settings and results in Table \ref{tb:ap-set-full}. 
Experiments on closed-source general models are given in Table \ref{ap-set-close}. 
Compared to open-source LVMs, these models show no significant overall performance improvement. Furthermore, on the MC task, all models underperform compared to the state-of-the-art LVM. This suggests that while popular multi-modality API-based models excel at object-scene distinction, they struggle in scenarios with increased confounding objects.
\begin{table}[H]
  \caption{Close-source models on Setting for Binary/MC tasks.}
  \label{ap-set-close}
  \centering
  \begin{tabular}{llllll}
\toprule
\textbf{Model} & \multicolumn{2}{c}{\textbf{Binary (\%)}} & \multicolumn{3}{c}{\textbf{MC (\%)}} \\
\cmidrule(lr){2-3} \cmidrule(lr){4-6}
              & Acc & Pos & Acc & OB & COB \\\midrule
    Best LVM & 90.3 & 49.1 & \textbf{94.9} & 0.71 & 0.52 \\
    \midrule
    GPT-4o     & \textbf{92.6} & 50.4 & \textbf{86.6} & 2.65 & 2.30  \\
    Gemini1.5-Pro & 90.6 & 49.8 & 83.8 & 3.38 & 2.29  \\
    Claude 3.5-Sonnet & 90.0 & 44.5 & 85.8 & 2.16 & 2.37 \\
    \bottomrule
  \end{tabular}
\end{table}
\subsection{Character hallucination}\label{ap:ep-char}
\begin{figure}[H]
    \centering
    \includegraphics[width=\columnwidth]{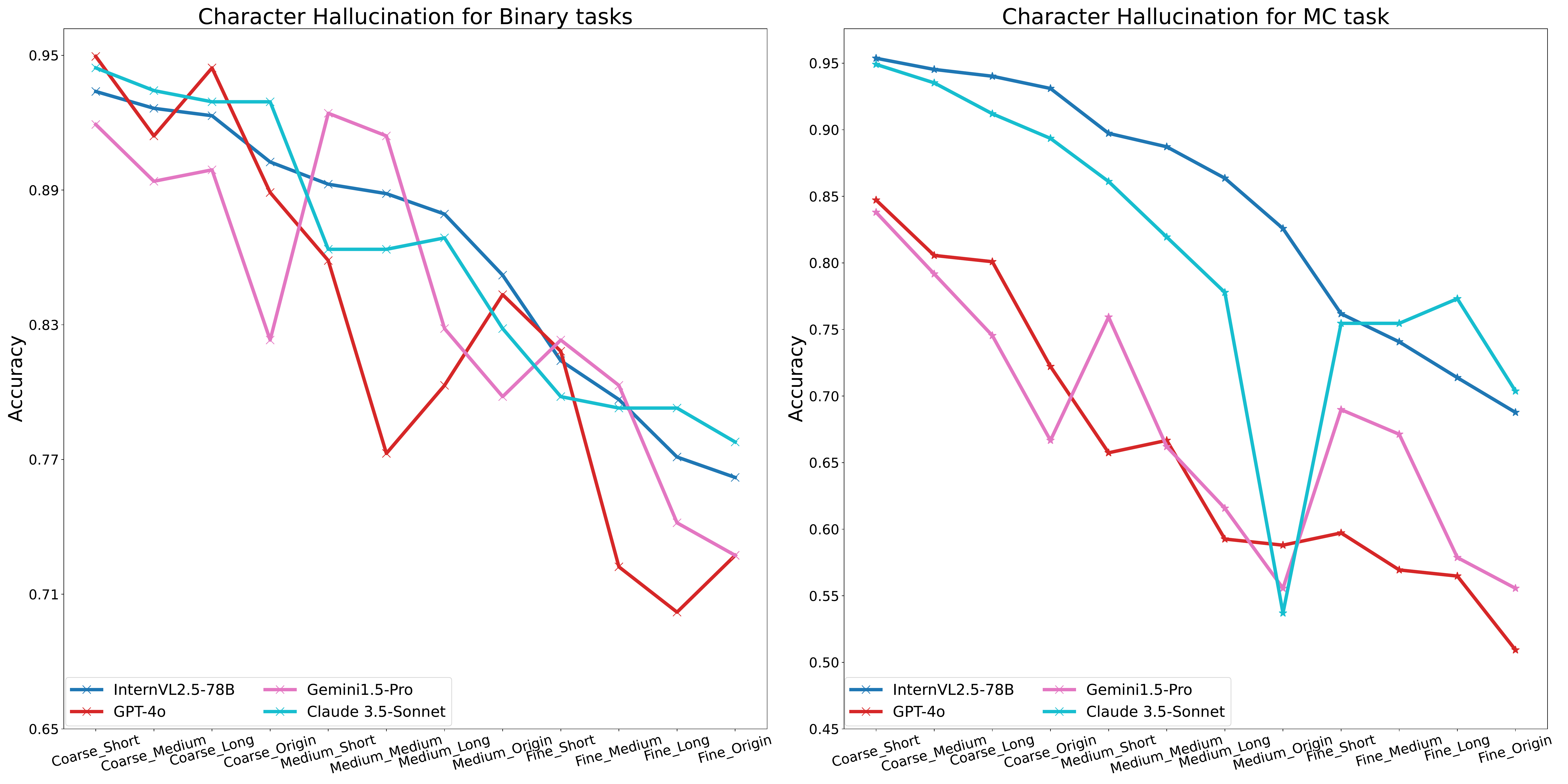} 
    \caption{Close-source models on Character for Binary/MC tasks.}
    \label{fig:char_results_compact-close}
\end{figure}

We present the detailed experiment settings and results in Table \ref{tb:ap-char-full-acc} and \ref{tb:ap-char-full-JSD}. 
For closed-source general models, Figure \ref{fig:char_results_compact-close} presents the comparison. These models generally follow the same decreasing trend as LVMs, from coarse to fine features and short to long video lengths. On binary tasks, their performance is comparable to the best LVM, but on the MC task, most underperform relative to InternVL2.5-78B, consistent with the results in Setting. A notable difference is their greater instability when varying feature granularity and video length, highlighting the challenges of using general models for video inputs.


\subsection{Stage hallucination}\label{ap:ep-sta}
We present the detailed experiment settings and results for action part in Table \ref{tb:ap-simpleQ_action} and \ref{tb:ap-mc_action}. 
We present the detailed experiment settings and results for dialogue part in Table \ref{tb:ap-simpleQ_dia} and \ref{tb:ap-mc_dia}.
The comparison in Tables \ref{tb:act_close} and \ref{tb:talk_close} shows that closed-source general models perform significantly worse than state-of-the-art LVMs. This is because specialized LVMs are fine-tuned on video datasets for video understanding. These results confirm our assumption that stage questions rely heavily on chronological information, requiring LVMs to leverage multi-frame data effectively—a capability general models lack.
\begin{table}[H]
\label{tb:talk_close_combined-close}
\centering
\begin{minipage}[t]{0.48\textwidth}
\centering
\caption{Close-source models on Stage: Action}
\label{tb:act_close}
\begin{tabular}{lcccc}
\toprule
\textbf{Model} & \multicolumn{2}{c}{\textbf{Binary \%}} & \multicolumn{2}{c}{\textbf{MC \%}} \\
\cmidrule(lr){2-3} \cmidrule(lr){4-5}
              & Acc & JSD & Acc & JSD \\\midrule
Best LVM & \textbf{85.3} & 0.05 & \textbf{85.5} & 0.05 \\
\midrule
GPT-4o & 81.4 & 0.15 & 67.3 & 0.66\\
Gemini1.5-Pro   & 73.3 & 0.03 & 60.6& 0.22\\
Claude 3.5-Sonnet    & 73.0 & 0.03& 75.1& 0.52\\
\bottomrule
\end{tabular}
\end{minipage}%
\hfill
\begin{minipage}[t]{0.48\textwidth}
\centering
\caption{Close-source models on Stage: Dialogue}
\label{tb:talk_close}
\begin{tabular}{lcccc}
\toprule
\textbf{Model} & \multicolumn{2}{c}{\textbf{Binary \%}} & \multicolumn{2}{c}{\textbf{MC \%}} \\
\cmidrule(lr){2-3} \cmidrule(lr){4-5}
              & Acc & JSD & Acc & JSD \\\midrule
    Best LVM & \textbf{80.6} & 4.40 & \textbf{83.8} & 0.09 \\
    \midrule
    GPT-4o & 79.3 & 0.14 & 73.4 & 0.45\\
    Gemini1.5-Pro       & \textbf{82.3} & 0.77& 77.9 & 0.33\\
    Claude 3.5-Sonnet      & 78.1 & 0.02& 77.8& 0.34\\
    \bottomrule
  \end{tabular}
\end{minipage}
\end{table}


\section{Additional Tables}\label{ap:tbs}

\begin{table}[h]
  \caption{Detailed performances on Setting for Binary/MC tasks.}
  \footnotesize
  \label{tb:ap-set-full}
  \centering
  \begin{tabular}{llllllllllll}
    \toprule
    \textbf{Model} & \textbf{NSF} & \textbf{LLM} & \multicolumn{3}{c}{\textbf{Binary Task}} & \multicolumn{4}{c}{\textbf{Multi-Choice Task}} \\
    \cmidrule(lr){4-6} \cmidrule(lr){7-10}
              &       &      & \textbf{Acc.} & \textbf{Pos.} & \textbf{JSD.} & \textbf{Acc.} & \textbf{OB.} & \textbf{COB.} & \textbf{JSD.} \\
    \midrule
        GPT-4o & 32 & GPT-4o    & \textbf{.9261} & .5041 & .000228 & \textbf{.8659} & .02650 & .02303 & .001246  \\
    Gemini1.5-Pro &32 & Gemini1.5-Pro& .9064 & .4975 & .000390 & .8379 & .03376 & .02292 & .002049  \\
    Claude 3.5-Sonnet &32 & Claude 3.5-Sonnet& .8998 & .4450 & .003243 & .8577 & .02156 & .02373 & .000344 \\\midrule
    LLaVA-Video-7B & 32 & Qwen2 & \textbf{.9064} & .4948 & .000148 & .9211 & .00377 & .00152 & .000085 \\
    LLaVA-Video-72B & 32 & Qwen2 & .9032 & .4905 & .000297 & .9225 & .00757 & .00645 & .000037 \\
    Aria-23B & 32 & Aria & .9003 & .4335 & .003091 & .8564 & .01950 & .00796 & .001334 \\
    VILA1.5-8B & 32 & Llama3 & .8724 & .4809 & .000488 & .7616 & .01577 & .00566 & .000869 \\
    InternVL2.5-78B & 32 & Qwen2.5 & .8692 & .4011 & .006598 & \textbf{.9486} & .00710 & .00524 & .000095 \\
    Qwen2VL-7B & 32 & Qwen2 & .8585 & .3971 & .006669 & .7319 & .02753 & .01995 & .002237 \\
    LLaVA-OneVision-72B & 32 & Qwen2 & .8384 & .5141 & .000000 & .8296 & .03133 & .01609 & .002189 \\
    LLaVA-OneVision-7B & 32 & Qwen2 & .8325 & .5427 & .000474 & .8413 & .04991 & .02178 & .006868 \\
    LLaVA-OneVision-7B-chat & 32 & Qwen2 & .8303 & .5559 & .000977 & .8408 & .04799 & .02126 & .006395 \\
    VideoLLaMA2.1-7B & 32 & Qwen2 & .8258 & .3614 & .011579 & .8493 & .02505 & .01206 & .001684 \\
    LongVILA-8B & 32 & Llama3 & .8070 & .6590 & .011117 & .6633 & .13745 & .08129 & .056180 \\
    LLaMA-VID-13B & 32 & Vicuna & .8010 & .6557 & .010416 & .5416 & .19723 & .09765 & .089951 \\
    InternVL2.5-4B & 32 & Qwen2.5 & .7941 & .4835 & .000408 & .8194 & .02835 & .01002 & .002259 \\
    Qwen2VL-72B & 32 & Qwen2 & .7841 & .3140 & .020958 & .8634 & .01520 & .00897 & .000840 \\
    InternVL2.5-8B & 32 & InternLM2.5 & .7837 & .4618 & .001251 & .8236 & .00782 & .00441 & .000317 \\
    InternVL2.5-2B & 32 & InternLM2.5 & .7803 & .4134 & .004898 & .7577 & .04353 & .01693 & .005349 \\
    VideoXL-7B & 32 & Qwen2 & .7709 & .5016 & .000072 & .7295 & .00927 & .00666 & .000249 \\
    Qwen2VL-2B & 32 & Qwen2 & .7694 & .2969 & .024246 & .8124 & .04746 & .03029 & .007235 \\
    LLaVA-NeXT-Video-7B & 24 & Vicuna & .7608 & .2827 & .028414 & .8190 & .05766 & .02434 & .007810 \\
    VideoAgent & ALL & Agent & .7236 & .5097 & .000019 & .6669 & .03081 & .02114 & .002765 \\
    InternVL2.5-1B & 32 & Qwen2.5 & .6815 & .5662 & .001477 & .5414 & .02313 & .01759 & .001265 \\
    Oryx1.5-7B & 32 & Qwen2.5 & .6935 & .2627 & .033192 & .6621 & .03468 & .01809 & .003739 \\
    Oryx1.5-32B & 32 & Qwen2.5 & .6896 & .2328 & .043309 & .6278 & .05836 & .02704 & .011931 \\
    Oryx-7B & 32 & Qwen2 & .6836 & .2242 & .045420 & .6256 & .03264 & .01731 & .002877 \\
    LLaVA-NeXT-Video-32B & 32 & Qwen1.5 & .6298 & .8634 & .074071 & .8137 & .02781 & .01349 & .001802 \\
    LLaMA-VID-7B & 32 & Vicuna & .5504 & .9260 & .114913 & .4847 & .20300 & .10089 & .095066 \\
    Video-LLaVA-7B & 32 & Vicuna & .5211 & .9888 & .184722 & .3306 & .35279 & .27039 & .299236 \\
    LLaVA-OneVision-0.5B & 32 & Qwen2 & .5121 & .9998 & .208337 & .6490 & .13945 & .06964 & .050139 \\
    LLaMA-VID-Long-Video & 32 & Vicuna & .4725 & .3077 & .021879 & .2346 & .36924 & .37310 & .339778 \\
    \bottomrule
  \end{tabular}
  \vspace{0.2cm}
\begin{minipage}{\columnwidth}
\scriptsize
  \centering
NSF for \textit{number of sampling frames}. LLM for \textit{language backbones}. OB and COB denote object-related metrics in multi-choice evaluation.
\end{minipage}
\end{table}
{\small
\begin{longtable}{llrrrrrrrr}
\caption{Accuracy performances on Character Hallucination}
\label{tb:ap-char-full-acc}
\\
\toprule
\textbf{Model} & \textbf{Grain} & \multicolumn{4}{c}{\textbf{Binary Task}} & \multicolumn{4}{c}{\textbf{Multi-Choice Task}} \\
\cmidrule(lr){3-6} \cmidrule(lr){7-10}
          &        & Short & Medium & Long & Origin & Short & Medium & Long & Origin\\
\midrule

\endfirsthead

\toprule
\textbf{Model} & \textbf{Grain} & \multicolumn{4}{c}{\textbf{Binary Task}} & \multicolumn{4}{c}{\textbf{Multi-Choice Task}} \\
\cmidrule(lr){3-6} \cmidrule(lr){7-10}
          &        & Short & Medium & Long & Origin & Short & Medium & Long & Origin\\
\midrule

\endhead
\midrule
\multicolumn{10}{r}{{Continued on next page}} \\
\midrule
\endfoot

\bottomrule
\endlastfoot
\multirow{1}{*}{GPT-4o} & Coarse &  .9495 & .9141 & .9444 & .8889 &  .8472 & .8056 & .8009 & .7222\\
&Medium & .8586 & .7727 & .8030 & .8434 & .6574 & .6667 & .5926 & .5880\\
& Mixed &  .7879 & .7576 & .7273 & .7525 &  .6204 & .5741 & .5417 & .5139\\
&Fine &  .8182 & .7222 & .7020 & .7273 &  .5972 & .5694 & .5648 & .5093\\
\midrule
\multirow{1}{*}{Gemini1.5-Pro} & Coarse &  .9192 & .8939 & .8990 & .8232 &  .8380 & .7917 & .7454 & .6667\\
&Medium & .9242 & .9141 & .8283 & .7980 & .7593 & .6620 & .6157 & .5556\\
&Mixed &  .8030 & .7929 & .7525 & .7424 &  .6481 & .5972 & .5163 & .5278\\
&Fine &  .8232 & .8030 & .7418 & .7273 &  .6898 & .6713 & .5787 & .5556\\\midrule

\multirow{1}{*}{Claude 3.5-Sonnet} &      Coarse &  .9444 & .9343 & .9293 & .9293 &  .9491 & .9352 & .9120 & .8935\\
&Medium & .8636 & .8636 & .8687 & .8283 & .8611 & .8194 & .7778 & .5370\\
&Mixed &  .7980 & .8384 & .8081 & .7828 &  .7361 & .7222 & .7037 & .6806\\
&Fine &  .7980 & .7929 & .7929 & .7778 &  .7546 & .7546 & .7731 & .7037\\
\midrule
\multirow{1}{*}{LLaVA-Video-7B} & Coarse & 0.9299 & 0.9217 & 0.9102 & 0.8872 & 0.9491 & 0.9389 & 0.9268 & 0.905 \\
 & Medium & 0.8139 & 0.7997 & 0.7792 & 0.732 & 0.7972 & 0.7724 & 0.744 & 0.6779 \\
 & Mixed & 0.7637 & 0.7483 & 0.7304 & 0.6882 & 0.6638 & 0.6419 & 0.6148 & 0.5551 \\
 & Fine & 0.7451 & 0.7251 & 0.6998 & 0.6511 & 0.5802 & 0.56 & 0.529 & 0.4657 \\
\midrule \multirow{1}{*}{Video-LLaVA-7B} & Coarse & 0.5318 & 0.5302 & 0.5269 & 0.5234 & 0.3369 & 0.3204 & 0.3032 & 0.2921 \\
 & Medium & 0.5002 & 0.5 & 0.5006 & 0.5 & 0.2599 & 0.2563 & 0.2539 & 0.2521 \\
 & Mixed & 0.5016 & 0.5018 & 0.5013 & 0.5005 & 0.2549 & 0.2544 & 0.2535 & 0.2519 \\
 & Fine & 0.5 & 0.5 & 0.5001 & 0.5 & 0.2533 & 0.2531 & 0.2519 & 0.2514 \\
\midrule \multirow{1}{*}{LongVILA-8B} & Coarse & 0.853 & 0.8501 & 0.8398 & 0.8055 & 0.8575 & 0.8374 & 0.8026 & 0.7547 \\
 & Medium & 0.7622 & 0.7422 & 0.7071 & 0.6768 & 0.6065 & 0.5771 & 0.5271 & 0.5007 \\
 & Mixed & 0.6736 & 0.6489 & 0.6236 & 0.5998 & 0.4683 & 0.4417 & 0.4123 & 0.3945 \\
 & Fine & 0.6898 & 0.6678 & 0.63 & 0.5764 & 0.4203 & 0.3956 & 0.3685 & 0.3332 \\
\midrule \multirow{1}{*}{LLaMA-VID-7B} & Coarse & 0.549 & 0.5387 & 0.5295 & 0.525 & 0.3751 & 0.3422 & 0.3151 & 0.2932 \\
 & Medium & 0.5074 & 0.5038 & 0.5024 & 0.5003 & 0.2608 & 0.2534 & 0.2503 & 0.2482 \\
 & Mixed & 0.5075 & 0.5054 & 0.5005 & 0.5001 & 0.2606 & 0.2549 & 0.2527 & 0.2502 \\
 & Fine & 0.502 & 0.5002 & 0.4993 & 0.4994 & 0.2693 & 0.26 & 0.2548 & 0.252 \\
\midrule \multirow{1}{*}{InternVL2.5-78B} & Coarse & 0.9339 & 0.9264 & 0.9231 & 0.9025 & 0.9537 & 0.9453 & 0.9402 & 0.931 \\
 & Medium & 0.8926 & 0.8884 & 0.8793 & 0.8521 & 0.8973 & 0.8872 & 0.8636 & 0.8258 \\
 & Mixed & 0.7603 & 0.7322 & 0.7289 & 0.7165 & 0.7618 & 0.7407 & 0.7138 & 0.6877 \\
 & Fine & 0.814 & 0.7967 & 0.7711 & 0.762 & 0.7811 & 0.7618 & 0.7483 & 0.6995 \\
\midrule \multirow{1}{*}{VideoXL-7B} & Coarse & 0.7957 & 0.7833 & 0.7574 & 0.7307 & 0.8142 & 0.7908 & 0.7629 & 0.7051 \\
 & Medium & 0.645 & 0.6193 & 0.5974 & 0.5692 & 0.5739 & 0.5336 & 0.4934 & 0.453 \\
 & Mixed & 0.621 & 0.6025 & 0.5789 & 0.5574 & 0.4549 & 0.4256 & 0.4117 & 0.3726 \\
 & Fine & 0.6142 & 0.5898 & 0.5623 & 0.5287 & 0.4205 & 0.3958 & 0.3673 & 0.3472 \\
\midrule \multirow{1}{*}{Qwen2-VL-72B} & Coarse & 0.8835 & 0.8926 & 0.8942 & 0.8868 & 0.9167 & 0.9108 & 0.8923 & 0.8956 \\
 & Medium & 0.7901 & 0.7785 & 0.7521 & 0.7157 & 0.7879 & 0.7483 & 0.702 & 0.6498 \\
 & Mixed & 0.7017 & 0.6893 & 0.6702 & 0.643 & 0.6423 & 0.6187 & 0.5909 & 0.5463 \\
 & Fine & 0.7198 & 0.6818 & 0.6545 & 0.6273 & 0.6456 & 0.6145 & 0.5943 & 0.5589 \\
\midrule \multirow{1}{*}{Oryx-7B} & Coarse & 0.7402 & 0.7136 & 0.6882 & 0.6635 & 0.6948 & 0.6566 & 0.6106 & 0.5644 \\
 & Medium & 0.6841 & 0.6614 & 0.6392 & 0.6185 & 0.5907 & 0.5559 & 0.5182 & 0.4808 \\
 & Mixed & 0.6246 & 0.6099 & 0.5946 & 0.5823 & 0.4858 & 0.46 & 0.428 & 0.4073 \\
 & Fine & 0.6261 & 0.6074 & 0.591 & 0.5729 & 0.43 & 0.4094 & 0.3883 & 0.3666 \\
\midrule \multirow{1}{*}{LLaVA-Video-72B} & Coarse & 0.9132 & 0.9058 & 0.8826 & 0.8653 & 0.9503 & 0.9428 & 0.9276 & 0.9074 \\
 & Medium & 0.9 & 0.8868 & 0.8727 & 0.8512 & 0.8662 & 0.8519 & 0.8291 & 0.798 \\
 & Mixed & 0.7884 & 0.7893 & 0.7645 & 0.7587 & 0.7189 & 0.7003 & 0.6667 & 0.6229 \\
 & Fine & 0.8099 & 0.8041 & 0.7694 & 0.7463 & 0.6961 & 0.7003 & 0.6742 & 0.6136 \\
\midrule \multirow{1}{*}{Qwen2-VL-7B} & Coarse & 0.8926 & 0.8846 & 0.8733 & 0.8615 & 0.8667 & 0.7656 & 0.714 & 0.7225 \\
 & Medium & 0.7458 & 0.6734 & 0.631 & 0.6316 & 0.7082 & 0.5684 & 0.4832 & 0.5047 \\
 & Mixed & 0.7098 & 0.6504 & 0.6158 & 0.612 & 0.6117 & 0.5165 & 0.4432 & 0.4639 \\
 & Fine & 0.7068 & 0.6235 & 0.5873 & 0.5895 & 0.5254 & 0.409 & 0.3565 & 0.3655 \\
\midrule \multirow{1}{*}{VILA1.5-8B} & Coarse & 0.9062 & 0.9026 & 0.88 & 0.8662 & 0.9088 & 0.8909 & 0.8429 & 0.8099 \\
 & Medium & 0.665 & 0.6574 & 0.6268 & 0.6066 & 0.6344 & 0.5861 & 0.4827 & 0.4486 \\
 & Mixed & 0.6205 & 0.6188 & 0.5969 & 0.5743 & 0.5039 & 0.4587 & 0.393 & 0.3654 \\
 & Fine & 0.6097 & 0.5941 & 0.5625 & 0.542 & 0.4369 & 0.3918 & 0.3297 & 0.3082 \\
\midrule \multirow{1}{*}{Qwen2-VL-2B} & Coarse & 0.8965 & 0.8907 & 0.8733 & 0.844 & 0.8618 & 0.8346 & 0.8027 & 0.7753 \\
 & Medium & 0.6468 & 0.619 & 0.5969 & 0.5691 & 0.5512 & 0.514 & 0.4787 & 0.4564 \\
 & Mixed & 0.6437 & 0.6093 & 0.5908 & 0.5633 & 0.4428 & 0.4137 & 0.3914 & 0.374 \\
 & Fine & 0.6292 & 0.595 & 0.573 & 0.5493 & 0.3908 & 0.3613 & 0.3465 & 0.3344 \\
\midrule \multirow{1}{*}{Oryx1.5-7B} & Coarse & 0.7492 & 0.7222 & 0.6925 & 0.6674 & 0.705 & 0.6676 & 0.6239 & 0.584 \\
 & Medium & 0.6921 & 0.6663 & 0.6474 & 0.622 & 0.5724 & 0.5447 & 0.5068 & 0.4752 \\
 & Mixed & 0.6359 & 0.6223 & 0.6044 & 0.5972 & 0.472 & 0.4519 & 0.4266 & 0.4027 \\
 & Fine & 0.6417 & 0.6193 & 0.6004 & 0.5848 & 0.4438 & 0.4246 & 0.3969 & 0.3826 \\
\midrule \multirow{1}{*}{LLaMA-VID-13B} & Coarse & 0.7542 & 0.7471 & 0.6879 & 0.6212 & 0.5748 & 0.5167 & 0.4571 & 0.3938 \\
 & Medium & 0.5942 & 0.5742 & 0.5553 & 0.5498 & 0.3218 & 0.2897 & 0.2557 & 0.2304 \\
 & Mixed & 0.5765 & 0.5641 & 0.5427 & 0.531 & 0.2987 & 0.2741 & 0.2561 & 0.2419 \\
 & Fine & 0.5606 & 0.5411 & 0.526 & 0.5143 & 0.2725 & 0.2628 & 0.2518 & 0.2395 \\
\midrule \multirow{1}{*}{LLaVA-OneVision-72B} & Coarse & 0.8347 & 0.8017 & 0.786 & 0.7793 & 0.8838 & 0.8577 & 0.8258 & 0.8215 \\
 & Medium & 0.8107 & 0.8025 & 0.7727 & 0.7273 & 0.7778 & 0.7399 & 0.7222 & 0.6684 \\
 & Mixed & 0.7421 & 0.7132 & 0.6917 & 0.6603 & 0.6347 & 0.6035 & 0.5901 & 0.537 \\
 & Fine & 0.7876 & 0.757 & 0.7116 & 0.6678 & 0.6103 & 0.601 & 0.5741 & 0.5118 \\
\midrule \multirow{1}{*}{Oryx1.5-32B} & Coarse & 0.7493 & 0.7156 & 0.6939 & 0.6712 & 0.7029 & 0.6643 & 0.6226 & 0.5697 \\
 & Medium & 0.6911 & 0.6605 & 0.6461 & 0.6138 & 0.5967 & 0.5601 & 0.5325 & 0.4801 \\
 & Mixed & 0.6563 & 0.6405 & 0.6178 & 0.6037 & 0.4787 & 0.45 & 0.4295 & 0.4055 \\
 & Fine & 0.6445 & 0.6209 & 0.6015 & 0.588 & 0.4341 & 0.3989 & 0.3975 & 0.3831 \\
\midrule \multirow{1}{*}{LLaMA-VID-Long-Video} & Coarse & 0.5014 & 0.5115 & 0.5117 & 0.5017 & 0.2746 & 0.2804 & 0.2715 & 0.2629 \\
 & Medium & 0.5034 & 0.49 & 0.496 & 0.4894 & 0.2792 & 0.2844 & 0.2824 & 0.2637 \\
 & Mixed & 0.5055 & 0.4957 & 0.5 & 0.4932 & 0.281 & 0.283 & 0.2765 & 0.2674 \\
 & Fine & 0.5135 & 0.5055 & 0.508 & 0.5051 & 0.2892 & 0.2932 & 0.2859 & 0.2754 \\
\midrule \multirow{1}{*}{\rewr{LLaVA-NeXT-Video-7B}} & Coarse & 0.8133 & 0.7972 & - & 0.7293 & 0.7287 & 0.6973 & - & 0.5788 \\
 & Medium & 0.6353 & 0.6148 & - & 0.5717 & 0.482 & 0.4491 & - & 0.367 \\
 & Mixed & 0.6102 & 0.5951 & - & 0.5547 & 0.4005 & 0.3775 & - & 0.3148 \\
 & Fine & 0.6082 & 0.5871 & - & 0.5351 & 0.3634 & 0.3408 & - & 0.2941 \\
\midrule \multirow{1}{*}{LLaVA-OneVision-7B-chat} & Coarse & 0.916 & 0.9103 & 0.9021 & 0.8823 & 0.9035 & 0.8858 & 0.8793 & 0.8408 \\
 & Medium & 0.8165 & 0.7852 & 0.7614 & 0.711 & 0.7301 & 0.6944 & 0.6753 & 0.6023 \\
 & Mixed & 0.7375 & 0.7093 & 0.688 & 0.6547 & 0.618 & 0.5854 & 0.5641 & 0.5066 \\
 & Fine & 0.7173 & 0.6835 & 0.6582 & 0.6205 & 0.5084 & 0.495 & 0.4745 & 0.4158 \\
\midrule \multirow{1}{*}{LLaVA-OneVision-7B} & Coarse & 0.9148 & 0.9084 & 0.902 & 0.8802 & 0.9015 & 0.8816 & 0.8758 & 0.8365 \\
 & Medium & 0.817 & 0.7867 & 0.7624 & 0.7107 & 0.7276 & 0.6904 & 0.6742 & 0.6011 \\
 & Mixed & 0.7377 & 0.7104 & 0.6883 & 0.6543 & 0.6164 & 0.5815 & 0.5638 & 0.5051 \\
 & Fine & 0.7183 & 0.6833 & 0.6615 & 0.6204 & 0.5077 & 0.4928 & 0.4738 & 0.4135 \\
\midrule \multirow{1}{*}{InternVL2.5-8B} & Coarse & 0.8361 & 0.8424 & 0.8389 & 0.8201 & 0.9042 & 0.9079 & 0.9006 & 0.8834 \\
 & Medium & 0.7086 & 0.7156 & 0.7064 & 0.6774 & 0.7628 & 0.7617 & 0.7519 & 0.7165 \\
 & Mixed & 0.6635 & 0.6559 & 0.6522 & 0.623 & 0.6097 & 0.5995 & 0.5844 & 0.545 \\
 & Fine & 0.6589 & 0.6564 & 0.6356 & 0.6049 & 0.5461 & 0.5339 & 0.5181 & 0.4808 \\
\midrule \multirow{1}{*}{VideoLLaMA2.1-7B} & Coarse & 0.6888 & 0.7307 & 0.7339 & 0.6967 & 0.8368 & 0.8215 & 0.8011 & 0.7347 \\
 & Medium & 0.5982 & 0.5911 & 0.5839 & 0.5655 & 0.5322 & 0.4913 & 0.4599 & 0.4216 \\
 & Mixed & 0.5702 & 0.5649 & 0.5573 & 0.5408 & 0.4629 & 0.4301 & 0.4079 & 0.3768 \\
 & Fine & 0.5557 & 0.5488 & 0.5357 & 0.5222 & 0.3602 & 0.3443 & 0.3281 & 0.3082 \\
\midrule \multirow{1}{*}{LLaVA-NeXT-Video-32B} & Coarse & 0.8238 & 0.8052 & 0.7735 & 0.7438 & 0.9147 & 0.8961 & 0.8758 & 0.8552 \\
 & Medium & 0.665 & 0.6492 & 0.6335 & 0.5917 & 0.5855 & 0.5686 & 0.5353 & 0.4798 \\
 & Mixed & 0.6105 & 0.6015 & 0.5818 & 0.5802 & 0.4787 & 0.4545 & 0.43 & 0.3712 \\
 & Fine & 0.57 & 0.5663 & 0.5453 & 0.5471 & 0.4231 & 0.4017 & 0.3749 & 0.3401 \\
\midrule \multirow{1}{*}{Aria} & Coarse & 0.9081 & 0.9148 & 0.9101 & 0.8991 & 0.8335 & 0.8119 & 0.7953 & 0.7736 \\
 & Medium & 0.7431 & 0.7386 & 0.7083 & 0.6934 & 0.7434 & 0.7116 & 0.679 & 0.655 \\
 & Mixed & 0.7015 & 0.6897 & 0.6636 & 0.6445 & 0.5803 & 0.5187 & 0.4985 & 0.4818 \\
 & Fine & 0.6802 & 0.6695 & 0.6346 & 0.615 & 0.5537 & 0.5082 & 0.4789 & 0.447 \\
\end{longtable}
}

{\small
\begin{longtable}{llrrrrrrrr}
\caption{JS Divergence performances on Character Hallucination}
\label{tb:ap-char-full-JSD}\\
\toprule
\textbf{Model} & \textbf{Grain} & \multicolumn{4}{c}{\textbf{Binary Task}} & \multicolumn{4}{c}{\textbf{Multi-Choice Task}} \\
\cmidrule(lr){3-6} \cmidrule(lr){7-10}
          &        & Short & Medium & Long & Origin & Short & Medium & Long & Origin\\
\midrule

\endfirsthead

\toprule
\textbf{Model} & \textbf{Grain} & \multicolumn{4}{c}{\textbf{Binary Task}} & \multicolumn{4}{c}{\textbf{Multi-Choice Task}} \\
\cmidrule(lr){3-6} \cmidrule(lr){7-10}
          &        & Short & Medium & Long & Origin & Short & Medium & Long & Origin\\
\midrule

\endhead
\midrule
\multicolumn{10}{r}{{Continued on next page}} \\
\midrule
\endfoot

\bottomrule
\endlastfoot
\multirow{1}{*}{Aria} & Coarse & 0.0014 & 0.0007 & 0.0008 & 0.0006 & 0.0006 & 0.001 & 0.0011 & 0.001 \\
 & Medium & 0.0229 & 0.0197 & 0.0261 & 0.0292 & 0.003 & 0.0026 & 0.0042 & 0.0066 \\
 & Mixed & 0.0263 & 0.0232 & 0.0274 & 0.0297 & 0.0009 & 0.0019 & 0.0023 & 0.0028 \\
 & Fine & 0.0275 & 0.0207 & 0.0263 & 0.0314 & 0.0023 & 0.0044 & 0.0037 & 0.006 \\
\midrule \multirow{1}{*}{InternVL2.5-78B} & Coarse & 0.0009 & 0.0006 & 0.0007 & 0.001 & 0.0001 & 0.0001 & 0.0002 & 0.0003 \\
 & Medium & 0.0017 & 0.0028 & 0.0036 & 0.0041 & 0.0003 & 0.0003 & 0.0006 & 0.0009 \\
 & Mixed & 0.0144 & 0.0175 & 0.0185 & 0.0191 & 0.0011 & 0.0013 & 0.0033 & 0.0033 \\
 & Fine & 0.0035 & 0.0045 & 0.0051 & 0.0045 & 0.0044 & 0.0046 & 0.0047 & 0.0064 \\
\midrule \multirow{1}{*}{InternVL2.5-8B} & Coarse & 0.0007 & 0.0009 & 0.001 & 0.0014 & 0.0005 & 0.0003 & 0.0002 & 0.0003 \\
 & Medium & 0.0001 & 0.0002 & 0.0004 & 0.001 & 0.0032 & 0.0026 & 0.0025 & 0.0029 \\
 & Mixed & 0.0003 & 0.0002 & 0.0 & 0.0001 & 0.0047 & 0.004 & 0.0036 & 0.0039 \\
 & Fine & 0.0001 & 0.0003 & 0.001 & 0.0027 & 0.0115 & 0.0131 & 0.0119 & 0.0146 \\
\midrule \multirow{1}{*}{LLaMA-VID-13B} & Coarse & 0.0005 & 0.0023 & 0.0333 & 0.0743 & 0.0166 & 0.0331 & 0.0605 & 0.0902 \\
 & Medium & 0.0094 & 0.0027 & 0.0008 & 0.0211 & 0.006 & 0.0157 & 0.0259 & 0.0344 \\
 & Mixed & 0.002 & 0.0 & 0.0053 & 0.0315 & 0.0011 & 0.009 & 0.0224 & 0.0331 \\
 & Fine & 0.0003 & 0.0024 & 0.0139 & 0.0439 & 0.0252 & 0.0469 & 0.0641 & 0.0809 \\
\midrule \multirow{1}{*}{LLaMA-VID-7B} & Coarse & 0.117 & 0.1279 & 0.1376 & 0.1483 & 0.193 & 0.2366 & 0.2819 & 0.3386 \\
 & Medium & 0.164 & 0.1843 & 0.1939 & 0.2059 & 0.2474 & 0.3 & 0.3912 & 0.4563 \\
 & Mixed & 0.0894 & 0.1073 & 0.1249 & 0.1479 & 0.3001 & 0.3386 & 0.3969 & 0.4408 \\
 & Fine & 0.1798 & 0.1918 & 0.2059 & 0.2103 & 0.3434 & 0.4142 & 0.487 & 0.5172 \\
\midrule \multirow{1}{*}{LLaMA-VID-Long-Video} & Coarse & 0.0414 & 0.0347 & 0.0181 & 0.0291 & 0.284 & 0.2742 & 0.2848 & 0.3193 \\
 & Medium & 0.035 & 0.0176 & 0.0088 & 0.0089 & 0.266 & 0.2454 & 0.245 & 0.2621 \\
 & Mixed & 0.0193 & 0.0148 & 0.0082 & 0.0108 & 0.2418 & 0.2303 & 0.2275 & 0.247 \\
 & Fine & 0.0283 & 0.025 & 0.0211 & 0.0155 & 0.2741 & 0.2586 & 0.265 & 0.2642 \\
\midrule \multirow{1}{*}{LLaVA-NeXT-Video-32B} & Coarse & 0.0133 & 0.0177 & 0.0241 & 0.0309 & 0.0003 & 0.0002 & 0.0006 & 0.0015 \\
 & Medium & 0.0002 & 0.0004 & 0.0009 & 0.0042 & 0.0032 & 0.0044 & 0.0041 & 0.0067 \\
 & Mixed & 0.0005 & 0.0039 & 0.0054 & 0.0119 & 0.0022 & 0.0044 & 0.0061 & 0.01 \\
 & Fine & 0.0001 & 0.0001 & 0.0002 & 0.0003 & 0.0006 & 0.0027 & 0.0039 & 0.0102 \\
\midrule \multirow{1}{*}{LLaVA-Video-7B} & Coarse & 0.0004 & 0.0008 & 0.0015 & 0.0031 & 0.0001 & 0.0001 & 0.0001 & 0.0001 \\
 & Medium & 0.0009 & 0.0005 & 0.0003 & 0.0001 & 0.001 & 0.0013 & 0.0012 & 0.0014 \\
 & Mixed & 0.0008 & 0.0002 & 0.0 & 0.0001 & 0.0015 & 0.0015 & 0.0011 & 0.0007 \\
 & Fine & 0.0012 & 0.0006 & 0.0002 & 0.0 & 0.006 & 0.0045 & 0.0031 & 0.0022 \\
 \midrule \multirow{1}{*}{LLaVA-NeXT-Video-7B} & Coarse & 0.0082 & 0.0093 & - & 0.024 & 0.0173 & 0.0234 & - & 0.0554 \\
 & Medium & 0.0357 & 0.0366 & - & 0.0417 & 0.0108 & 0.0146 & - & 0.0266 \\
 & Mixed & 0.0394 & 0.0373 & - & 0.0462 & 0.018 & 0.0232 & - & 0.0357 \\
 & Fine & 0.0218 & 0.0215 & - & 0.0238 & 0.0318 & 0.038 & - & 0.0435 \\
\midrule \multirow{1}{*}{LLaVA-OneVision-0.5B} & Coarse & 0.2158 & 0.2158 & 0.2158 & 0.2158 & 0.0119 & 0.013 & 0.0132 & 0.0101 \\
 & Medium & 0.2158 & 0.2158 & 0.2158 & 0.2157 & 0.0154 & 0.0229 & 0.0227 & 0.0139 \\
 & Mixed & 0.2158 & 0.2158 & 0.2158 & 0.2155 & 0.0218 & 0.0325 & 0.0299 & 0.0202 \\
 & Fine & 0.2158 & 0.2158 & 0.2158 & 0.2157 & 0.0201 & 0.0374 & 0.047 & 0.0444 \\
\midrule \multirow{1}{*}{LLaVA-OneVision-72B} & Coarse & 0.0119 & 0.0166 & 0.0194 & 0.0208 & 0.0008 & 0.0008 & 0.0013 & 0.0011 \\
 & Medium & 0.0143 & 0.0149 & 0.0205 & 0.0313 & 0.0008 & 0.0016 & 0.0019 & 0.0016 \\
 & Mixed & 0.0099 & 0.0079 & 0.0146 & 0.0201 & 0.0007 & 0.0006 & 0.0012 & 0.0012 \\
 & Fine & 0.003 & 0.0032 & 0.0081 & 0.0124 & 0.0009 & 0.0015 & 0.0042 & 0.0041 \\
\midrule \multirow{1}{*}{LLaVA-OneVision-7B} & Coarse & 0.0005 & 0.0005 & 0.0007 & 0.0006 & 0.0012 & 0.0024 & 0.0029 & 0.0041 \\
 & Medium & 0.0006 & 0.0014 & 0.0017 & 0.0042 & 0.0035 & 0.006 & 0.0051 & 0.0079 \\
 & Mixed & 0.0023 & 0.005 & 0.0055 & 0.0096 & 0.0079 & 0.0112 & 0.0098 & 0.0134 \\
 & Fine & 0.0024 & 0.0054 & 0.0069 & 0.0126 & 0.0117 & 0.0102 & 0.0063 & 0.007 \\
\midrule \multirow{1}{*}{LLaVA-OneVision-7B-chat} & Coarse & 0.0004 & 0.0005 & 0.0008 & 0.0008 & 0.0011 & 0.0021 & 0.0025 & 0.0035 \\
 & Medium & 0.0006 & 0.0014 & 0.0016 & 0.0039 & 0.0034 & 0.0052 & 0.0044 & 0.0065 \\
 & Mixed & 0.0022 & 0.005 & 0.0052 & 0.0089 & 0.0078 & 0.01 & 0.0082 & 0.011 \\
 & Fine & 0.0025 & 0.0058 & 0.007 & 0.0125 & 0.0103 & 0.0078 & 0.0045 & 0.005 \\
\midrule \multirow{1}{*}{LLaVA-Video-72B} & Coarse & 0.0006 & 0.0018 & 0.0036 & 0.0042 & 0.0001 & 0.0 & 0.0001 & 0.0002 \\
 & Medium & 0.0 & 0.0003 & 0.0015 & 0.0013 & 0.0003 & 0.0006 & 0.0003 & 0.0008 \\
 & Mixed & 0.0001 & 0.0 & 0.0001 & 0.0002 & 0.0006 & 0.001 & 0.0003 & 0.001 \\
 & Fine & 0.0002 & 0.0 & 0.0001 & 0.0002 & 0.0028 & 0.0011 & 0.0039 & 0.0054 \\
\midrule \multirow{1}{*}{LongVILA-8B} & Coarse & 0.0046 & 0.0052 & 0.0071 & 0.0139 & 0.0093 & 0.0121 & 0.0188 & 0.0301 \\
 & Medium & 0.0017 & 0.0011 & 0.0007 & 0.0021 & 0.0534 & 0.0545 & 0.0641 & 0.0684 \\
 & Mixed & 0.0146 & 0.0126 & 0.0112 & 0.0175 & 0.0697 & 0.0688 & 0.0783 & 0.0822 \\
 & Fine & 0.0086 & 0.0068 & 0.0052 & 0.0122 & 0.0984 & 0.0866 & 0.0871 & 0.0832 \\
\midrule \multirow{1}{*}{Oryx-7B} & Coarse & 0.0227 & 0.0301 & 0.0378 & 0.0469 & 0.0015 & 0.0016 & 0.0018 & 0.0014 \\
 & Medium & 0.0383 & 0.0468 & 0.0553 & 0.0643 & 0.0038 & 0.0035 & 0.0041 & 0.004 \\
 & Mixed & 0.0521 & 0.0588 & 0.0663 & 0.0749 & 0.0066 & 0.0064 & 0.0061 & 0.0059 \\
 & Fine & 0.0441 & 0.0524 & 0.0616 & 0.0705 & 0.0253 & 0.0282 & 0.032 & 0.0358 \\
\midrule \multirow{1}{*}{Oryx1.5-32B} & Coarse & 0.025 & 0.0322 & 0.0381 & 0.0482 & 0.006 & 0.0069 & 0.0105 & 0.0125 \\
 & Medium & 0.0276 & 0.035 & 0.0429 & 0.0478 & 0.0096 & 0.014 & 0.016 & 0.0198 \\
 & Mixed & 0.036 & 0.0417 & 0.0457 & 0.0518 & 0.0197 & 0.0229 & 0.0275 & 0.0297 \\
 & Fine & 0.0271 & 0.039 & 0.0443 & 0.0488 & 0.0136 & 0.0188 & 0.0201 & 0.0219 \\
\midrule \multirow{1}{*}{Oryx1.5-7B} & Coarse & 0.018 & 0.0231 & 0.0303 & 0.0356 & 0.002 & 0.0024 & 0.0035 & 0.0055 \\
 & Medium & 0.0293 & 0.0376 & 0.0442 & 0.0503 & 0.0071 & 0.0085 & 0.0109 & 0.0121 \\
 & Mixed & 0.0358 & 0.0436 & 0.0499 & 0.0547 & 0.0117 & 0.0133 & 0.0138 & 0.016 \\
 & Fine & 0.0336 & 0.0417 & 0.0491 & 0.0563 & 0.0063 & 0.009 & 0.0127 & 0.0156 \\
\midrule \multirow{1}{*}{Qwen2-VL-2B} & Coarse & 0.0011 & 0.0012 & 0.0028 & 0.0058 & 0.0041 & 0.0069 & 0.0111 & 0.0136 \\
 & Medium & 0.0379 & 0.0504 & 0.0622 & 0.0846 & 0.0587 & 0.0752 & 0.0918 & 0.1021 \\
 & Mixed & 0.0245 & 0.0384 & 0.0516 & 0.0757 & 0.0705 & 0.0933 & 0.111 & 0.1121 \\
 & Fine & 0.0249 & 0.0371 & 0.0477 & 0.0694 & 0.0868 & 0.1257 & 0.1482 & 0.154 \\
\midrule \multirow{1}{*}{Qwen2-VL-72B} & Coarse & 0.003 & 0.0027 & 0.0017 & 0.0019 & 0.0011 & 0.0013 & 0.0024 & 0.002 \\
 & Medium & 0.015 & 0.0172 & 0.0221 & 0.0265 & 0.0067 & 0.0086 & 0.0145 & 0.0142 \\
 & Mixed & 0.0271 & 0.0304 & 0.0316 & 0.042 & 0.0257 & 0.0272 & 0.038 & 0.0465 \\
 & Fine & 0.0198 & 0.0267 & 0.0286 & 0.0364 & 0.019 & 0.0245 & 0.0265 & 0.0288 \\
\midrule \multirow{1}{*}{Qwen2-VL-7B} & Coarse & 0.0015 & 0.0009 & 0.0012 & 0.002 & 0.0001 & 0.0047 & 0.0053 & 0.0008 \\
 & Medium & 0.0154 & 0.0217 & 0.0253 & 0.0237 & 0.0027 & 0.0043 & 0.032 & 0.0052 \\
 & Mixed & 0.0109 & 0.0137 & 0.0171 & 0.0163 & 0.0059 & 0.0094 & 0.0593 & 0.0124 \\
 & Fine & 0.009 & 0.014 & 0.0164 & 0.0151 & 0.0103 & 0.0631 & 0.0749 & 0.0236 \\
\midrule \multirow{1}{*}{VILA1.5-8B} & Coarse & 0.0016 & 0.0009 & 0.0006 & 0.0006 & 0.0006 & 0.0007 & 0.0013 & 0.0015 \\
 & Medium & 0.0449 & 0.0371 & 0.0334 & 0.0349 & 0.0068 & 0.007 & 0.0097 & 0.0089 \\
 & Mixed & 0.0578 & 0.0449 & 0.0325 & 0.0335 & 0.007 & 0.0074 & 0.0097 & 0.0086 \\
 & Fine & 0.0609 & 0.0519 & 0.041 & 0.0387 & 0.0336 & 0.039 & 0.0531 & 0.049 \\
\midrule \multirow{1}{*}{Video-LLaVA-7B} & Coarse & 0.1556 & 0.1536 & 0.1462 & 0.1448 & 0.344 & 0.3589 & 0.3755 & 0.387 \\
 & Medium & 0.2115 & 0.2117 & 0.2117 & 0.2115 & 0.4697 & 0.4691 & 0.4781 & 0.4823 \\
 & Mixed & 0.2048 & 0.2018 & 0.2007 & 0.2011 & 0.482 & 0.4811 & 0.4843 & 0.4868 \\
 & Fine & 0.215 & 0.2153 & 0.215 & 0.2144 & 0.5126 & 0.512 & 0.5149 & 0.5137 \\
\midrule \multirow{1}{*}{VideoLLaMA2.1-7B} & Coarse & 0.0532 & 0.0365 & 0.0338 & 0.0456 & 0.0058 & 0.0039 & 0.0042 & 0.0048 \\
 & Medium & 0.0582 & 0.0544 & 0.0506 & 0.0597 & 0.0054 & 0.0058 & 0.0044 & 0.0074 \\
 & Mixed & 0.0745 & 0.0705 & 0.0659 & 0.0745 & 0.0094 & 0.0096 & 0.0096 & 0.011 \\
 & Fine & 0.0647 & 0.0662 & 0.0615 & 0.0657 & 0.0195 & 0.0104 & 0.0092 & 0.0121 \\
\midrule \multirow{1}{*}{VideoXL-7B} & Coarse & 0.0003 & 0.0003 & 0.0006 & 0.0007 & 0.0001 & 0.0002 & 0.0001 & 0.0001 \\
 & Medium & 0.0052 & 0.0073 & 0.0092 & 0.0099 & 0.0053 & 0.0052 & 0.0056 & 0.0062 \\
 & Mixed & 0.0038 & 0.0053 & 0.0081 & 0.0095 & 0.009 & 0.0097 & 0.0085 & 0.01 \\
 & Fine & 0.0037 & 0.0055 & 0.0089 & 0.01 & 0.0076 & 0.0085 & 0.0112 & 0.0151 \\
\end{longtable}
}

\begin{table}
  \caption{LVMs' on Stage: Non-dialogue for Binary tasks}
  \label{tb:ap-simpleQ_action}
  \centering
  \begin{tabular}{llllllll}
    \toprule
    LVM    & W Avg & JSD & EA &  COV & AOV  & SA & MIV  \\\midrule
    GPT-4o              & .81407 & .00033 & .72115 & .94059 & .93069 & .89000 & .86667 \\
    Gemini1.5-Pro   & .73256 & .00007 & .57843 & .95050 & .87129 & .85000 & .87500 \\
    Claude 3.5-Sonnet    & .72956 & .00059 & .53846 & .99010 & .91089 & .89000 & .89167 \\
    \midrule
    InternVL2.5-78B     & .85254 & .00005 & .77170 & .97843 & .95092 & .92500 & .87917 \\
    LLaVA-Video-72B     & .83019 & .00936 & .87942 & .87211 & .80675 & .79500 & .65000 \\
    LLaVA-Video-7B     & .80974 & .00803 & .85490 & .89104 & .85394 & .78000 & .53333 \\
    LLaVA-OneVision-7B-chat     & .78596 & .01446 & .87982 & .89398 & .77693 & .69750 & .40000 \\
    LLaVA-OneVision-7B     & .78495 & .01481 & .87982 & .89020 & .77677 & .69333 & .40000 \\
    LLaVA-NeXT-Video-32B     & .78447 & .01730 & .86174 & .74632 & .75997 & .72250 & .60000 \\
    Qwen2VL-7B     & .78279 & .00991 & .80868 & .89861 & .79564 & .77083 & .56250 \\
    LLaVA-OneVision-72B     & .77736 & .03623 & .94695 & .73721 & .63574 & .60819 & .45000 \\
    Aria-23B     & .77530 & .00026 & .67042 & .94629 & .91777 & .89000 & .76667 \\
    Qwen2VL-72B     & .74846 & .00415 & .55064 & .98108 & .96319 & .94083 & .90000 \\
    VILA1.5-8B     & .73806 & .02415 & .82195 & .82064 & .71019 & .69833 & .38750 \\
    InternVL2.5-8B   & .71039 & .05260 & .85002 & .65049 & .64297 & .61829 & .37131 \\
    Qwen2VL-2B     & .69379 & .00119 & .47548 & .97378 & .93541 & .93083 & .80833 \\
    VideoXL-7B     & .66517 & .00787 & .54976 & .80415 & .82631 & .79732 & .69456 \\
    VideoLLaMA2.1-7B     & .65353 & .00311 & .38505 & .96340 & .95627 & .95583 & .81250 \\
    LongVILA-8B     & .64279 & .12937 & .94534 & .55168 & .33016 & .32500 & .15417 \\
    LLaMA-VID-13B    & .62602 & .02439 & .59188 & .73056 & .71923 & .66165 & .52917 \\
    Oryx1.5-7B     & .59861 & .00414 & .26125 & .95597 & .93710 & .92583 & .92500 \\
    Oryx-7B     & .59579 & .00650 & .24397 & .97125 & .94584 & .93167 & .94167 \\
    Oryx1.5-32B     & .59453 & .00669 & .24598 & .97588 & .94477 & .93083 & .92083 \\
    LLaVA-NeXT-Video-7B     & .58980 & .01182 & .21584 & .97840 & .97330 & .97000 & .93333 \\
    LLaMA-VID-Long-Video-7B     & .52466 & .16089 & .74719 & .32913 & .33691 & .30083 & .24167 \\
    LLaMA-VID-7B     & .51225 & .39224 & .97588 & .03225 & .06722 & .04917 & .04583 \\
    Video-LLaVA-7B     & .50402 & .43679 & .99116 & .01641 & .02363 & .01500 & .01250 \\ 
    VideoAgent     & .58139 & .00056 & .33548 & .81202 & .84462 & .87477 & .77778 \\
    LLaVA-OneVision-0.5B     & .50056 & .46938 & .99799 & .00224 & .00445 & .00167 & .00417 \\
    \bottomrule
  \end{tabular}
\end{table}

\begin{longtable}{lllllll}
\caption{LVMs' on Stage: Non-dialogue for Multi-choice tasks} \label{tb:ap-mc_action} \\

\toprule
LVM & Metrics & Overall & COV & AOV & SA & MIV \\
\midrule
\endfirsthead

\toprule
LVM & Metrics & Overall & COV & AOV & SA & MIV \\
\midrule
\endhead

\midrule
\multicolumn{7}{r}{Continued on next page...} \\
\midrule
\endfoot

\bottomrule
\endlastfoot
    \multirow{4}{*}{GPT-4o}         & Accuracy    & .67338 & .63492  & .80620 & .80952 & .44286  \\
& OB     & .02344 & .04940  & .01586 & .02280  & .03945  \\
& COB    & .01663 & .05796  & .03189 & .02246 & .02908  \\
& JSD    & .00251 & .00262  & .00308 & .00060 & .01370  \\
\midrule
    \multirow{4}{*}{Gemini1.5-Pro}  & Accuracy    & .60624 & .63492  & .65116 & .67460 & .46429  \\
& OB     & .02104 & .04811  & .07134 & .04143  & .01956  \\
& COB    & .02612 & .05728  & .07576 & .04353 & .01276  \\
& JSD    & .00084 & .00477  & .00434 & .00353 & .00974  \\
\midrule
    \multirow{4}{*}{Claude 3.5-Sonnet}  & Accuracy    & .75106 & .80952  & .82171 & .73016 & .64286  \\
& OB     & .03370 & .05219  & .05409 & .04105  & .07053  \\
& COB    & .02812 & .04926  & .05399 & .03994 & .06111  \\
& JSD    & .00197 & .00166  & .00420 & .00478 & .00692  \\
\midrule
    \multirow{4}{*}{InternVL2.5-78B} & Accuracy    & .85534 & .90517  & .90575 & .86399 & .74643  \\
& OB     & .01309 & .02149  & .01321 & .01352  & .03847  \\
& COB    & .00989 & .01996  & .01138 & .01234 & .04037  \\
& JSD    & .00017 & .00037  & .00017 & .00007 & .00070  \\
\midrule
    \multirow{4}{*}{LLaVA-Video-72B}    & Accuracy    & .82966 & .87759  & .90575 & .84602 & .68929  \\
& OB     & .01253 & .01764  & .01202 & .02064  & .01451  \\
& COB    & .01049 & .01877  & .00995 & .01544 & .03305  \\
& JSD    & .00014 & .00009  & .00017 & .00042 & .00078  \\
\midrule
    \multirow{4}{*}{LLaVA-Video-7B}    & Accuracy    & .82264 & .90597  & .89769 & .83690 & .65000  \\
& OB     & .01107 & .00856  & .01483 & .01348  & .04315  \\
& COB    & .00716 & .00674  & .01092 & .00946 & .04891  \\
& JSD    & .00013 & .00002  & .00015 & .00036 & .00077  \\
\midrule
    \multirow{4}{*}{InternVL2.5-8B} & Accuracy    & .76609 & .86451  & .83795 & .77707 & .58484  \\
& OB     & .01530 & .01752  & .01912 & .01089  & .00469  \\
& COB    & .00924 & .01388  & .01276 & .00437 & .02411  \\
& JSD    & .00042 & .00031  & .00063 & .00041 & .00119  \\
\midrule
    \multirow{4}{*}{LLaVA-NeXT-Video-32B}    & Accuracy    & .76125 & .80259  & .86155 & .78443 & .59643  \\
& OB     & .01691 & .02196  & .01900 & .01792  & .07012  \\
& COB    & .01277 & .01811  & .01520 & .01490 & .06845  \\
& JSD    & .00029 & .00021  & .00037 & .00051 & .00456  \\
\midrule
    \multirow{4}{*}{LLaVA-OneVision-7B-chat}     & Accuracy    & .74606 & .84214  & .84042 & .79812 & .50357  \\
& OB     & .02528 & .02247  & .02685 & .02972  & .02661  \\
& COB    & .01474 & .02012  & .01434 & .01159 & .03452  \\
& JSD    & .00150 & .00083  & .00175 & .00220 & .00584  \\
\midrule
    \multirow{4}{*}{LLaVA-OneVision-7B}     & Accuracy    & .74177 & .83696  & .83570 & .79441 & .50000  \\
& OB     & .02940 & .02717  & .02996 & .03354  & .03369  \\
& COB    & .01701 & .02265  & .01534 & .01396 & .03388  \\
& JSD    & .00198 & .00129  & .00215 & .00273 & .00695  \\
\midrule
    \multirow{4}{*}{Qwen2VL-72B}     & Accuracy    & .73912 & .78190  & .79066 & .81608 & .56786  \\
& OB     & .04514 & .04403  & .04860 & .03183  & .10561  \\
& COB    & .02893 & .02652  & .03271 & .02212 & .10462  \\
& JSD    & .00467 & .00569  & .00536 & .00218 & .01697  \\
\midrule
    \multirow{4}{*}{LLaVA-OneVision-72B}     & Accuracy    & .73003 & .78362  & .77731 & .76989 & .58929  \\
& OB     & .02153 & .02809  & .01909 & .02765  & .02513  \\
& COB    & .01364 & .02409  & .01235 & .01505 & .03866  \\
& JSD    & .00055 & .00055  & .00048 & .00125 & .00141  \\
\midrule
    \multirow{4}{*}{VILA1.5-8b}     & Accuracy    & .69174 & .79685  & .77287 & .72940 & .46786  \\
& OB     & .02914 & .03352  & .03052 & .02456  & .03562  \\
& COB    & .01625 & .02140  & .01727 & .01065 & .02665  \\
& JSD    & .00180 & .00169  & .00198 & .00179 & .00425  \\
\midrule
    \multirow{4}{*}{Aria-23B}     & Accuracy    & .67730 & .68989  & .71031 & .78757 & .52143  \\
& OB     & .02706 & .03333  & .02448 & .02604  & .02461  \\
& COB    & .01794 & .02994  & .01665 & .00998 & .03407  \\
& JSD    & .00152 & .00198  & .00098 & .00174 & .00144  \\
\midrule
    \multirow{1}{*}{Qwen2VL-7B}     & Accuracy    & .67171 & .71857  & .72894 & .76076 & .47857  \\
& OB     & .06840 & .07078  & .07165 & .05800  & .15205  \\
& COB    & .04717 & .04231  & .05362 & .04366 & .13459  \\
& JSD    & .01047 & .01187  & .01151 & .00747 & .04513  \\
\midrule
    \multirow{4}{*}{Qwen2VL-2B}     & Accuracy    & .64973 & .65862  & .74368 & .76447 & .43214  \\
& OB     & .07865 & .09046  & .07532 & .06471  & .11172  \\
& COB    & .05239 & .06150  & .05166 & .04195 & .10970  \\
& JSD    & .01380 & .01866  & .01263 & .00935 & .02842  \\
\midrule
    \multirow{4}{*}{VideoLLaMA2.1-7B}     & Accuracy    & .63791 & .65236  & .75480 & .75877 & .38571  \\
& OB     & .05798 & .07124  & .04685 & .04163  & .19722  \\
& COB    & .02626 & .02505  & .02445 & .02510 & .16654  \\
& JSD    & .00588 & .01021  & .00357 & .00253 & .04725  \\
\midrule
    \multirow{4}{*}{LongVILA-8B}     & Accuracy    & .60895 & .65797  & .66083 & .64557 & .47143  \\
& OB     & .11411 & .12307  & .10548 & .11037  & .12748  \\
& COB    & .08391 & .09547  & .07543 & .07740 & .08819  \\
& JSD    & .03425 & .03846  & .02975 & .03262 & .05534  \\
\midrule
    \multirow{4}{*}{VideoXL-7B}     & Accuracy    & .59611 & .63060  & .69331 & .63964 & .42086  \\
& OB     & .04204 & .03565  & .04169 & .04851  & .08392  \\
& COB    & .02596 & .01550  & .02870 & .03487 & .08361  \\
& JSD    & .00335 & .00288  & .00303 & .00460 & .00752  \\
\midrule
    \multirow{4}{*}{LLaVA-NeXT-Video-7B}    & Accuracy    & .55734 & .51736  & .66741 & .64461 & .40000  \\
& OB     & .06321 & .08413  & .04757 & .05487  & .03360  \\
& COB    & .03847 & .06915  & .02555 & .02145 & .01786  \\
& JSD    & .00891 & .01429  & .00546 & .00721 & .00823  \\
\midrule
    \multirow{4}{*}{LLaVA-OneVision-0.5B}     & Accuracy    & .49069 & .48135  & .57492 & .54577 & .36071  \\
& OB     & .11133 & .12498  & .09776 & .10805  & .11359  \\
& COB    & .07676 & .09557  & .06184 & .07369 & .08699  \\
& JSD    & .02462 & .03015  & .01918 & .02433 & .02648  \\
\midrule
    \multirow{4}{*}{Oryx1.5-32B}     & Accuracy    & .47065 & .51758  & .49124 & .54520 & .32857  \\
& OB     & .05783 & .07413  & .06159 & .03452  & .05107  \\
& COB    & .03068 & .02995  & .03955 & .02555 & .02306  \\
& JSD    & .00655 & .01225  & .00709 & .00211 & .00473  \\
\midrule
    \multirow{4}{*}{Oryx-7B}     & Accuracy    & .46555 & .50205  & .50236 & .52923 & .32857  \\
& OB     & .03317 & .04875  & .02088 & .02834  & .06911  \\
& COB    & .01657 & .02025  & .00929 & .02504 & .06700  \\
& JSD    & .00230 & .00585  & .00124 & .00122 & .00784  \\
\midrule
    \multirow{4}{*}{Oryx1.5-7B}     & Accuracy    & .48457 & .55661  & .50570 & .52238 & .35357  \\
& OB     & .05427 & .06030  & .05390 & .04817  & .06522  \\
& COB    & .03274 & .02756  & .03673 & .03772 & .05731  \\
& JSD    & .00601 & .00826  & .00612 & .00432 & .00512  \\
\midrule
    \multirow{4}{*}{VideoAgent}     & Accuracy    & .40169 & .37671  & .38631 & .46018 & .38356  \\
& OB     & .04951 & .05674  & .05547 & .03348  & .07526  \\
& COB    & .05376 & .08826  & .05472 & .02451 & .03571  \\
& JSD    & .00409 & .00288  & .00500 & .00568 & .01923  \\
\midrule
    \multirow{1}{*}{LLaMA-VID-13B}     & Accuracy    & .37579 & .35728  & .45010 & .44223 & .25357  \\
& OB     & .13043 & .13447  & .11696 & .14326  & .10824  \\
& COB    & .08911 & .11056  & .07094 & .08812 & .09148  \\
& JSD    & .03114 & .03181  & .02676 & .03735 & .02803  \\
\midrule
    \multirow{4}{*}{LLaMA-VID-7B}     & Accuracy    & .34046 & .31324  & .40391 & .40097 & .24373  \\
& OB     & .27100 & .30786  & .24318 & .24774  & .30998  \\
& COB    & .19869 & .26168  & .16361 & .16729 & .26735  \\
& JSD    & .11762 & .15328  & .09503 & .09741 & .18199  \\
    \midrule
        \multirow{4}{*}{Video-LLaVA-7B}     & Accuracy    & .26384 & .26957  & .28135 & .28657 & .21786  \\
& OB     & .39500 & .40049  & .39634 & .38562  & .40438  \\
& COB    & .35356 & .37054  & .34716 & .33635 & .39555  \\
& JSD    & .28011 & .28973  & .28253 & .26252 & .33570  \\
    \midrule
    \multirow{4}{*}{LLaMA-VID-Long-Video-7B}     & Accuracy    & .24726 & .26634  & .25910 & .26718 & .19643  \\
& OB     & .31721 & .32747  & .30794 & .31516  & .29721  \\
& COB    & .30468 & .32098  & .29409 & .29500 & .29016  \\
& JSD    & .19453 & .20196  & .19326 & .18820 & .18904  \\
\end{longtable}

\begin{longtable}{lllllll}
\caption{LVMs' on Stage: Dialogue for Multi-choice tasks} 
\label{tb:ap-mc_dia} \\

\toprule
LVM & Metrics & Overall & COV & CIV \\
\midrule
\endfirsthead

\toprule
LVM & Metrics & Overall & COV & CIV \\
\midrule
\endhead

\midrule
\multicolumn{7}{r}{Continued on next page...} \\
\midrule
\endfoot

\bottomrule
\endlastfoot
    \multirow{4}{*}{GPT-4o}         & Accuracy    & .73359 & .81853  & .64865  \\
& OB     & .02294 & .03816  & .01647  \\
& COB    & .01870 & .03285  & .00729  \\
& JSD    & .00172 & .00198  & .00154  \\
    \midrule
        \multirow{4}{*}{Gemini1.5-Pro}      & Accuracy    & .77889 & .83051  & .72727  \\
& OB     & .02783 & .03241  & .02340  \\
& COB    & .02425 & .03166  & .01926  \\
& JSD    & .00126 & .00134  & .00165  \\
\midrule
    \multirow{4}{*}{Claude 3.5-Sonnet}  & Accuracy    & .77799 & .88417  & .67181  \\
& OB     & .02711 & .03425  & .02274  \\
& COB    & .02263 & .02919  & .02207  \\
& JSD    & .00129 & .00150  & .00223  \\
\midrule
\multirow{4}{*}{InternVL2.5-78B}     & Accuracy    & .83820 & .94886  & .72755  \\
& OB     & .01543 & .00988  & .02209  \\
& COB    & .01114 & .00711  & .01686  \\
& JSD    & .00034 & .00021  & .00088  \\
\midrule
    \multirow{4}{*}{LLaVA-Video-72B}      & Accuracy    & .82166 & .92351  & .71981  \\
& OB     & .00290 & .00756  & .00178  \\
& COB    & .00526 & .00544  & .00578  \\
& JSD    & .00014 & .00036  & .00005  \\
\midrule
    \multirow{4}{*}{LLaVA-Video-7B}      & Accuracy    & .81951 & .93124  & .70778  \\
& OB     & .01033 & .00685  & .01401  \\
& COB    & .00862 & .00643  & .01270  \\
& JSD    & .00012 & .00005  & .00031  \\
\midrule
    \multirow{4}{*}{Qwen2VL-72B}     & Accuracy    & .79387 & .91147  & .67627  \\
& OB     & .05425 & .02754  & .08116  \\
& COB    & .03737 & .01953  & .06146  \\
& JSD    & .00559 & .00102  & .01483  \\
\midrule
    \multirow{4}{*}{VILA1.5-8B}     & Accuracy    & .78814 & .89686  & .67942  \\
& OB     & .01139 & .00977  & .01502  \\
& COB    & .00895 & .00539  & .01394  \\
& JSD    & .00029 & .00033  & .00053  \\
\midrule
    \multirow{4}{*}{LLaVA-OneVision-72B}     & Accuracy    & .78513 & .88354  & .68672  \\
& OB     & .00672 & .00478  & .01380  \\
& COB    & .00722 & .00529  & .01158  \\
& JSD    & .00009 & .00021  & .00033  \\
\midrule
    \multirow{4}{*}{InternVL2.5-8B}     & Accuracy    & .78061 & .90102  & .66020  \\
& OB     & .01118 & .00854  & .01494  \\
& COB    & .00644 & .00319  & .01147  \\
& JSD    & .00044 & .00031  & .00078  \\
\midrule
    \multirow{4}{*}{LongVILA-8B}     & Accuracy    & .76472 & .85131  & .67813  \\
& OB     & .05053 & .04321  & .05795  \\
& COB    & .02672 & .02318  & .03182  \\
& JSD    & .00679 & .00544  & .00840  \\
\midrule
    \multirow{4}{*}{Aria-23B}    & Accuracy    & .75330 & .85039  & .65621  \\
& OB     & .01528 & .01867  & .02041  \\
& COB    & .00845 & .01039  & .01060  \\
& JSD    & .00081 & .00132  & .00110  \\
\midrule
    \multirow{4}{*}{LLaVA-NeXT-Video-32B}     & Accuracy    & .75118 & .85905  & .64332  \\
& OB     & .01603 & .02019  & .01439  \\
& COB    & .00621 & .00626  & .00764  \\
& JSD    & .00059 & .00117  & .00034  \\
\midrule
    \multirow{4}{*}{LLaVA-OneVision-7B-chat}     & Accuracy    & .75075 & .89729  & .60421  \\
& OB     & .01435 & .00568  & .02511  \\
& COB    & .00836 & .00269  & .01821  \\
& JSD    & .00059 & .00035  & .00130  \\
\midrule
    \multirow{4}{*}{LLaVA-OneVision-7B}     & Accuracy    & .74839 & .89514  & .60163  \\
& OB     & .01405 & .00533  & .02478  \\
& COB    & .00847 & .00340  & .01768  \\
& JSD    & .00058 & .00032  & .00128  \\
\midrule
\multirow{4}{*}{Qwen2VL-7B}     & Accuracy    & .74710 & .85131  & .64289  \\
& OB     & .05738 & .04220  & .07338  \\
& COB    & .03913 & .02779  & .05467  \\
& JSD    & .00655 & .00310  & .01188  \\
\midrule
        \multirow{4}{*}{Oryx1.5-32B}     & Accuracy    & .74602 & .85303  & .63902  \\
& OB     & .03544 & .02817  & .04350  \\
& COB    & .02362 & .01633  & .03381  \\
& JSD    & .00189 & .00097  & .00328  \\
\midrule
    \multirow{4}{*}{Oryx1.5-7B}     & Accuracy    & .74431 & .83455  & .65406  \\
& OB     & .02129 & .01773  & .02691  \\
& COB    & .01510 & .01329  & .01779  \\
& JSD    & .00062 & .00042  & .00116  \\
\midrule
    \multirow{4}{*}{Oryx-7B}     & Accuracy    & .74108 & .83025  & .65191  \\
& OB     & .02634 & .02869  & .02796  \\
& COB    & .01075 & .00799  & .01564  \\
& JSD    & .00139 & .00212  & .00123  \\
\midrule
    \multirow{4}{*}{VideoLLaMA2.1-7B}     & Accuracy    & .74044 & .83928  & .64160  \\
& OB     & .02815 & .01974  & .03885  \\
& COB    & .01748 & .01277  & .02496  \\
& JSD    & .00112 & .00037  & .00260  \\
\midrule
    \multirow{4}{*}{Qwen2VL-2B}     & Accuracy    & .72733 & .83627  & .61839  \\
& OB     & .06667 & .05269  & .08306  \\
& COB    & .04773 & .03588  & .06447  \\
& JSD    & .00996 & .00569  & .01687  \\
    \midrule
    \multirow{4}{*}{LLaVA-OneVision-0.5B}     & Accuracy    & .72174 & .82639  & .61710  \\
& OB     & .03995 & .03636  & .04382  \\
& COB    & .02286 & .01849  & .03009  \\
& JSD    & .00454 & .00401  & .00511  \\
\midrule
    \multirow{4}{*}{VideoXL-7B}      & Accuracy    & .71612 & .81298  & .61925  \\
& OB     & .01767 & .01337  & .02310  \\
& COB    & .01489 & .01064  & .02077  \\
& JSD    & .00034 & .00013  & .00075  \\
\midrule
    \multirow{4}{*}{LLaVA-NeXT-Video-7B}     & Accuracy    & .65813 & .73709  & .57917  \\
& OB     & .07936 & .08412  & .07471  \\
& COB    & .03629 & .03377  & .03997  \\
& JSD    & .01391 & .01591  & .01207  \\
\midrule
    \multirow{4}{*}{LLaMA-VID-13B}     & Accuracy    & .54598 & .63086  & .46111  \\
& OB     & .10499 & .09757  & .11302  \\
& COB    & .05561 & .04498  & .07061  \\
& JSD    & .02018 & .01838  & .02260  \\
\midrule
    \multirow{4}{*}{LLaMA-VID-7B}     & Accuracy    & .36506 & .39708  & .33305  \\
& OB     & .30118 & .29315  & .30920  \\
& COB    & .21621 & .19421  & .24258  \\
& JSD    & .15487 & .14768  & .16252  \\
\midrule
    \multirow{4}{*}{LLaMA-VID-Long-Video-7B}     & Accuracy    & .30705 & .30855  & .30554  \\
& OB     & .21638 & .21046  & .22231  \\
& COB    & .20458 & .19543  & .21388  \\
& JSD    & .11542 & .11025  & .12105  \\
\midrule
    \multirow{4}{*}{Video-LLaVA-7B}      & Accuracy    & .28578 & .29308  & .27847  \\
& OB     & .40171 & .39936  & .40407  \\
& COB    & .34912 & .33238  & .36678  \\
& JSD    & .29504 & .29185  & .29855  \\
\end{longtable}

\begin{table}[t]
  \caption{LVMs' on Stage: Dialogue for Binary tasks}
  \label{tb:ap-simpleQ_dia}
  \centering
  \begin{tabular}{lllllll}
    \toprule
    LVM     & Weighted Average & JSD & ET &  COV & AOV  &  CIV \\
    \midrule
GPT-4o              & .79326 & .00003 & .69672 & .98347 & .91736 & .76860 \\
Gemini1.5-Pro       & .82319 & .00166 & .78689 & .99174 & .87603 & .71074 \\
Claude 3.5-Sonnet      & .78094 & .00004 & .68033 & .00000 & .92562 & .71901 \\
\midrule
    LLaVA-Video-72B     & .80595 & .00951 & .88146 & .91864 & .85269 & .42000 \\
    LLaVA-Video-7B     & .80082 & .00683 & .83326 & .94538 & .86734 & .49241 \\
    VILA1.5-8B     & .79207 & .01546 & .87799 & .90231 & .80995 & .40621 \\
    InternVL2.5-78B     & .79024 & .00001 & .72167 & .98595 & .95397 & .63655 \\
    Qwen2VL-7B     & .78416 & .00837 & .79939 & .94538 & .80694 & .55448 \\
    LLaVA-OneVision-7B-chat     & .78519 & .01581 & .87712 & .91978 & .79384 & .36621 \\
    LLaVA-OneVision-7B     & .78369 & .01671 & .87842 & .91652 & .78551 & .36483 \\
    LLaVA-OneVision-72B     & .77140 & .02492 & .92271 & .80621 & .72025 & .33379 \\
    Qwen2VL-2B     & .76223 & .00095 & .68736 & .97232 & .91002 & .62897 \\
    LLaVA-NeXT-Video-32B     & .75449 & .02242 & .88971 & .73151 & .84632 & .28000 \\
    LLaVA-NeXT-Video-7B     & .74967 & .00006 & .61435 & .94146 & .95229 & .76121 \\
    Oryx1.5-32B     & .74696 & .00015 & .61224 & .97528 & .89391 & .77586 \\
    Oryx-7B     & .74578 & .00037 & .61876 & .96507 & .89125 & .76207 \\
    Oryx1.5-7B     & .73947 & .00070 & .61919 & .95249 & .89160 & .73517 \\
    Qwen2VL-72B     & .73510 & .00350 & .57794 & .98595 & .96530 & .72552 \\
    Aria-23B     & .73221 & .00001 & .60703 & .97661 & .94793 & .64759 \\
    LLaMA-VID-13B     & .71878 & .02283 & .78637 & .74452 & .88009 & .32897 \\
    VideoXL-7B     & .71121 & .01506 & .69952 & .79165 & .84194 & .53509 \\
    VideoLLaMA2.1-7B-16f     & .69792 & .00107 & .49023 & .97795 & .96440 & .77448 \\
    InternVL2.5-8B     & .69924 & .02956 & .74727 & .80786 & .67040 & .47540 \\
    LongVILA-8B     & .65889 & .12075 & .93096 & .61590 & .30252 & .24207 \\
    LLaMA-VID-7B     & .51728 & .40015 & .99739 & .01273 & .09529 & .00345 \\
    LLaMA-VID-Long-Video-7B     & .50609 & .24305 & .80808 & .19050 & .17906 & .24276 \\
    LLaVA-OneVision-0.5B     & .50061 & .47166 & .99957 & .00104 & .00390 & .00000 \\
    Video-LLaVA-7B     & .50400 & .45490 & 1. & .00148 & .02249 & .00000 \\
    \bottomrule
  \end{tabular}
\end{table}

\section{Additional Case Study}\label{ap:case-study}
\begin{figure}[H]
    \centering
    \includegraphics[width=0.9\columnwidth]{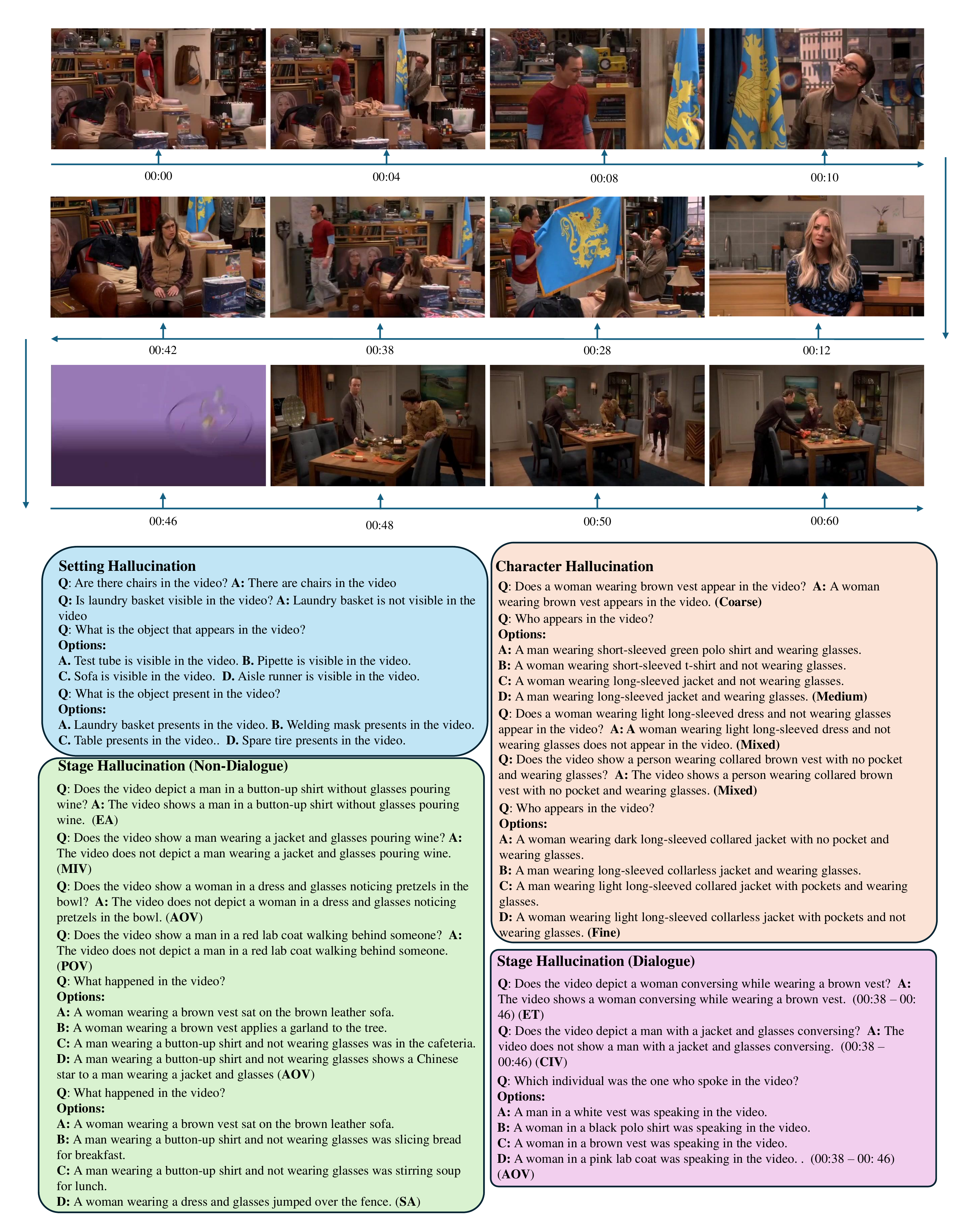} 
    \caption{An example of one video and part of its corresponding hallucination questions.}
    \label{fig:CaseStudy1}
\end{figure}

\begin{figure}[H]
    \centering
    \includegraphics[width=0.9\columnwidth]{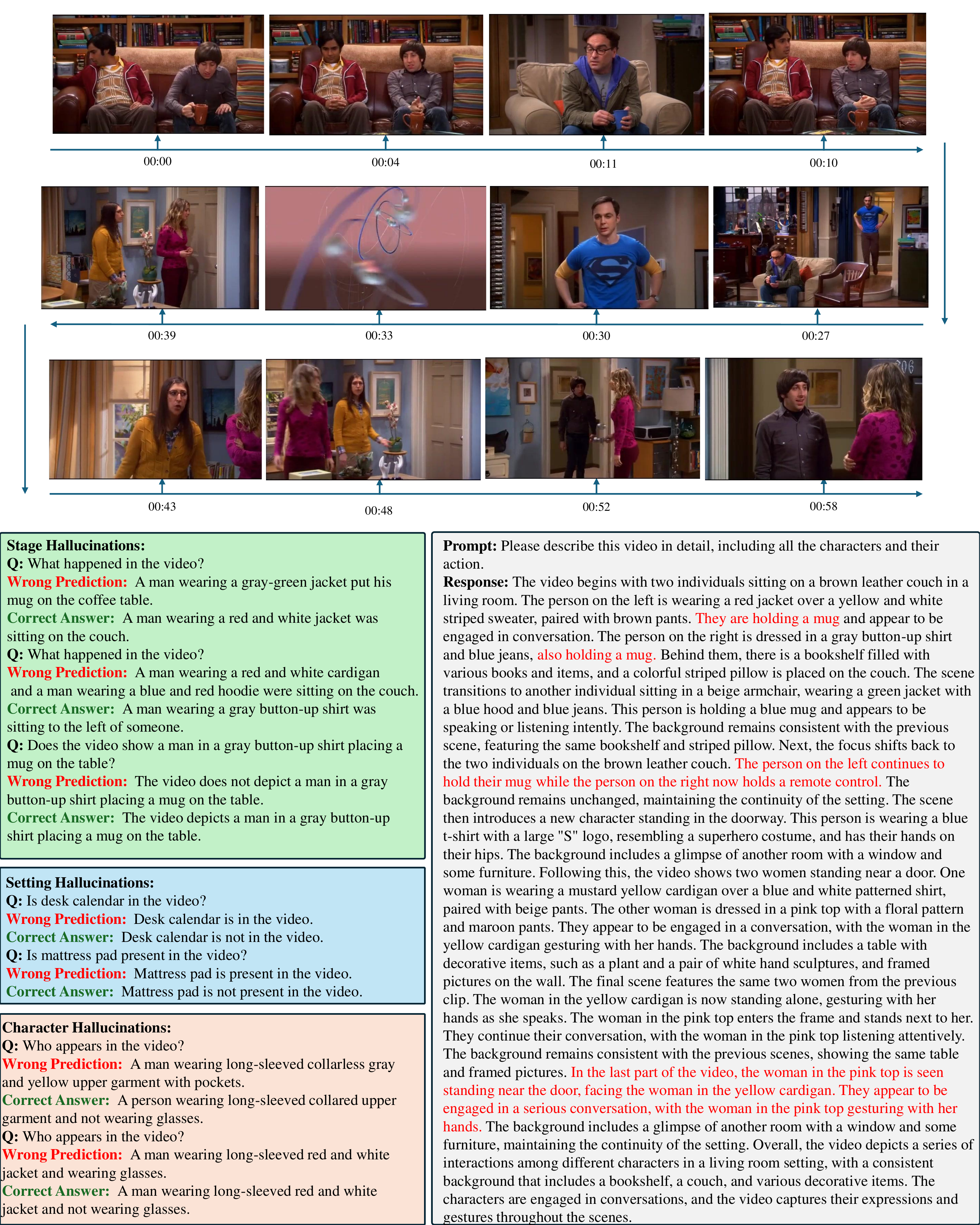} 
    \caption{Analysis of video question answering and description tasks reveals that LVMs exhibit a tendency to avoid generating descriptions for content that elicits uncertainty, as specifically detected through question answering; instances of hallucinated content are highlighted in red.}
    \label{fig:CaseStudy2}
\end{figure}

\end{document}